\documentclass[final]{statsoc}
\title[Interpretable Classification Models for Recidivism Prediction]{Interpretable Classification Models for Recidivism Prediction}
\author[Zeng, Ustun, and Rudin]{Jiaming Zeng$^{\dagger}$, Berk Ustun$^{\dagger}$, Cynthia Rudin}
\email{jiaming@mit.edu}
\providecommand{\daggerexp}[0]{\quad\small$^{\dagger}${These authors contributed equally to this work.}\normalsize}

\usepackage{amsfonts,bbm,bm,times,dsfont}
\usepackage{hyperref}
\usepackage{etex}
\usepackage{relsize}
\usepackage{adjustbox}

\usepackage{amsmath,amssymb}
\usepackage{mathtools}
\usepackage{eqnarray}
\usepackage{bm}

\usepackage{array}
\usepackage{tabularx}
\usepackage{dcolumn}
\usepackage{multirow}
\usepackage{hhline}
\usepackage{booktabs}

\newcommand{\predcell}[2]{\begin{tabular}{#1}\footnotesize{\textbf{#2}} \end{tabular}}
\newcommand{\cell}[2]{\setlength{\tabcolsep}{0pt}\begin{tabular}{#1}#2 \end{tabular}}

\newcommand{\bfcell}[2]{\setlength{\tabcolsep}{0pt}\textbf{\begin{tabular}{#1}#2\end{tabular}}}

\newcolumntype{d}{D{.}{.}{-1}}

\usepackage{comment}
\usepackage{verbatim}
\usepackage{fancyvrb}
\usepackage{colortbl}
\usepackage{placeins}

\usepackage{enumitem} 

\usepackage{graphicx}
\usepackage{epstopdf} 

\usepackage{tikz,pgfplots}
\usetikzlibrary{shapes,arrows}
\usetikzlibrary{matrix}
\pgfplotsset{compat=newest}
\pgfplotsset{plot coordinates/math parser=false}

\newlength\fheight
\newlength\fwidth
\newcommand{\tableds}[1]{\texttt{#1}}
\newcommand{\textds}[1]{\texttt{\footnotesize{#1}}}
\newcommand{\textfn}[1]{\textit{#1}}

\newcommand{\lzero}{\ell_0}
\newcommand{\lone}{\ell_1}

\newcommand{\indic}[1]{\mathbbm{1}\left[#1\right]}

\newcommand{\for}{\textnormal{ for }}
\newcommand{\sign}[1]{\textnormal{sign}\left(#1\right)}
\newcommand{\lambdab}{\bm{\lambda}}
\newcommand{\xb}{\bm{x}}

\newcommand{\iplus}{\mathcal{I}^{+}} 
\newcommand{\iminus}{\mathcal{I}^{-}}
\newcommand{\wplus}{W^{+}} 
\newcommand{\wminus}{W^{-}}
\newcommand{\nplus}{N^{+}}
\newcommand{\nminus}{N^{-}}

\newcommand{\Lset}{\mathcal{L}}

\newcommand{\R}{\mathbb{R}}

\newcommand{\Z}{\mathbb{Z}}
\newcommand{\B}{\{0,1\}}

\newcommand{\IntPen}{\Phi}
\newcommand{\loss}{z}

\newcommand{\st}{\textnormal{s.t.}}
\newcommand{\mprange}[3]{{#1}={#2}\textnormal{...}{#3}}
\newcommand{\mpdes}[1]{\textit{\scriptsize #1~~} }

\newcommand{\pkg}[1]{{\fontseries{b}\selectfont #1}} 

\begin{document} 
\maketitle 
\daggerexp

\begin{abstract}
We investigate a long-debated question, which is how to create predictive models of recidivism that are sufficiently accurate, transparent, and interpretable to use for decision-making. This question is complicated as these models are used to support different decisions, from sentencing, to determining release on probation, to allocating preventative social services. Each case might have an objective other than classification accuracy, such as a desired true positive rate (TPR) or false positive rate (FPR). Each (TPR, FPR) pair is a point on the receiver operator characteristic (ROC) curve. We use popular machine learning methods to create models along the full ROC curve on a wide range of recidivism prediction problems. We show that many methods (SVM, SGB, Ridge Regression) produce equally accurate models along the full ROC curve. However, methods that designed for interpretability (CART, C5.0) cannot be tuned to produce models that are accurate and/or interpretable. To handle this shortcoming, we use a recent method called Supersparse Linear Integer Models (SLIM)  to produce accurate, transparent, and interpretable scoring systems along the full ROC curve. These scoring systems can be used for decision-making for many different use cases, since they are just as accurate as the most powerful black-box machine learning models for many applications, but completely transparent, and highly interpretable. 
\end{abstract}


\thispagestyle{empty}

\section{Introduction}\label{Sec::Introduction}


Forecasting has been used for criminology applications since the 1920s \citep{borden1928factors,burgess1928factors} when various factors derived from age, race, prior offense history, employment, grades, and neighborhood background were used to estimate success of parole. Many things have changed since then, including the fact that we have developed machine learning methods that can produce accurate predictive models, and have collected large high-dimensional datasets on which to apply them.

Recidivism prediction is still extremely important. In the United States, for example, a minority of individuals commit the majority of the crimes \citep{wolfgang1987delinquency}: these are the ``power few" of  \citet{sherman2007power} on which we should focus our efforts. We want to ensure that public resources are directed effectively, be they correctional facilities or preventative social services. \citet{milgram2014ted} recently discussed the critical importance of accurately predicting if an individual who is released on bail poses a risk to public safety, pointing out that high-risk individuals are being released 50\% of the time while low-risk individuals are being released less often then they should be. Her observations are in line with longstanding work on clinical versus actuarial judgment, which shows that humans, on their own, are not as good at risk assessment as statistical models \citep{dawes1989clinical,grove1996comparative}. This is the reason that several U.S. states have mandated the use of predictive models for sentencing decisions \citep{pew2011risk,wroblewski2014letter}.

There has been some controversy as to whether sophisticated machine learning methods \citep[such as random forests, see e.g.,][]{breimanRF,berk2009forecasting,ritter2013predicting} are necessary to produce accurate predictive models of recidivism, or if traditional approaches such as logistic regression or linear discriminant analysis would suffice \citep[see e.g.,][]{tollenaar2013method,berk2013statistical,bushway2013there}. Random forests may produce accurate predictive models, but these models effectively operate as black-boxes,  which make it difficult to understand \textit{how} the input variables are producing a predicted outcome. If a simpler, more transparent, but equally accurate predictive model could be developed, it would be more usable and defensible for many decision-making applications. There is a precedent for using such models in criminology \citep{steinhart2006juvenile, andrade2009handbook}; \citet{pitfall} argues that a ``decent transparent model that is actually used will outperform a sophisticated system that predicts better but sits on a shelf." This discussion is captured nicely by \citet{bushway2013there}, who contrasts the works of \citet{berk2013statistical} and \citet{tollenaar2013method}. \citet{berk2013statistical} claim we need sophisticated machine learning methods due to their substantial benefits in accuracy, whereas \citet{tollenaar2013method} claim that ``modern statistical, data mining and machine learning models provides no real advantage over logistic regression and LDA," assuming that humans have done appropriate pre-processing. In this work, we argue that the answer to the question is far more subtle than a simple yes or no.

In particular, the answer depends on how the models will be used for decision-making. For each use case (e.g., sentencing, parole decisions, policy interventions), one might need a decision point at a different level of true positive rate (TPR) and false positive rate (FPR) \citep[see also][]{ritter2013predicting}. Each (TPR, FPR) pair is a point on the receiver operator characteristic (ROC) curve. To determine if one method is better than another, one must consider the appropriate point along the ROC curve for decision-making. As we show, for a wide range of recidivism prediction problems, many machine learning methods (support vector machines, random forests) produce equally accurate predictive models along the ROC curve. However, there are trade-offs between accuracy, transparency, and interpretability: methods that are designed to yield transparent models (CART, C5.0) cannot be tuned to produce as accurate models along the ROC curve, and do not always yield models that are interpretable. This is not to say that interpretable models for recidivism prediction do not exist. The fact that many machine learning methods produce models with similar levels of predictive accuracy indicates that there is a large class of approximately-equally-accurate predictive models (called the ``Rashomon" effect by \citealt{breiman-cultures}). In this case, there may exist interpretable models that also attain the same level of accuracy. Finding models that are accurate and interpretable, however, is computationally challenging. 

In this paper, we explore whether such accurate-yet-interpretable models exist and how to find them. To this end, we use a new machine learning method known as a Supersparse Linear Integer Model  \citep[SLIM;][]{ustun2015slim} to learn \textit{scoring systems} from data. Scoring systems that have used for many criminal justice applications because they let users make quick predictions by adding, subtracting and multiplying a few small numbers \citep[see e.g.,][]{hoffman1980salient,sentencing1987,pasc2012report4}. In contrast to existing tools, which have been built using heuristic approaches \citep[see e.g.,][]{gottfredson2005mathematics}, the models built by SLIM are fully optimized for accuracy and sparsity, and can handle additional constraints (e.g., bounds on the false positive rate, monotonicity properties for the coefficients). We use SLIM to produce a set of simple scoring systems at different decision points across the full ROC curve, and provide a comparison with other popular machine learning methods. Our findings show that the SLIM scoring systems are often just as accurate as the most powerful black-box machine learning models, but transparent and highly interpretable.

\subsection{Structure}
The remainder of this paper is structured as follows.
In Section \ref{Sec::RelatedWork}, we discuss related work. 
In Section \ref{Sec::Data}, we describe how we derived 6 recidivism prediction problems.
In Section \ref{Sec::SLIMDescription}, we provide a brief overview of SLIM and describe several new techniques that can reduce the computation required to produce scoring systems.
In Section \ref{Sec::Results}, we compare the accuracy and interpretability of models produced by the 9 machine learning methods on the 6 recidivism prediction problems.
We include additional results related to the accuracy and interpretability of models from different methods in the Appendix.
%
\subsection{Related Work}\label{Sec::RelatedWork}
%
%
Predictive models for recidivism have been in widespread use in different countries and different areas of the criminal justice system since the early 1920s \citep[see e.g.,][]{borden1928factors,burgess1928factors,tibbitts1931success}. The use of these tools has been spurred on by continued research into the superiority of actuarial judgment \citep{dawes1989clinical,grove1996comparative} as well as a desire to efficiently use limited public resources  \citep{clements1996offender,simon2005reversal,mccord1978thirty,mccord2003cures}. In the U.S., federal guidelines currently mandate the use of a predictive recidivism measure known as the Criminal History Category for sentencing \citep{sentencing1987}. Besides the U.S., countries that currently use risk assessment tools include Canada \citep{hanson2003notes}, the Netherlands \citep{tollenaar2013method}, and the U.K. \citep{howard2009ogrs}. Applications of these tools can be seen in evidence-based sentencing \citep{hoffman1994twenty}, corrections and prison administration \citep{belfrage2000prediction}, informing release on parole \citep{pew2011risk}, determining the level of supervision during parole \citep{barnes2012classifying,ritter2013predicting}, determining appropriate sanctions for parole violations \citep{turner2009development}, and targeted policy interventions \citep{lowenkamp2004understanding}.

Our paper focuses on binary classification models to predict general recidivism (i.e., recidivism of any type of crime) as well as crime-specific recidivism (i.e., recidivism for drug, general violence, domestic violence, sexual violence, and fatal violence offenses). Risk assessment tools for general recidivism include: the Salient Factor Score \citep{hoffman1980salient,hoffman1994twenty}, the Offender Group Reconviction Scale \citep{copas1998offender,maden2006assessing,howard2009ogrs}, the Statistical Information of Recidivism scale \citep{nafekh2002statistical}, and the Level of Service/Case Management Inventory \citep{andrews2000level}. Crime-specific applications include risk assessment tools for domestic violence \citep[see e.g., the Spousal Abuse Risk Assessment of][]{kropp2000spousal}, sexual violence \citep[see e.g., ][]{hanson2003notes,langton2007actuarial}, and general violence (see e.g., Historical Clinical and Risk Management tool of \citealt{webster1997hcr}, or the Structured Assessment of Violence Risk in Youth tool of \citealt{borum2006manual}).

The scoring systems that we present in this paper are designed to mimic the form of risk scores that are currently used throughout the criminal justice system -- that is, linear classification models that only require users to add, subtract and multiply a few small numbers to make a prediction \citep[][]{ustun2015slim}. These tools are unique in that they allow users make quick predictions by hand, without a computer, calculator, or nomogram (which is a visualization tool for more difficult calculations). Current examples of such tools include: the Salient Factor Score (SFS) \citep{hoffman1980salient}, the Criminal History Category (CHC) \citep{sentencing1987}, and the Offense Gravity Score (OGS) \citep{pasc2012report4}. Our approach aims to produce scoring systems that are fully optimized for accuracy and sparsity without any post-processing. In contrast, current tools are produced through heuristic approaches that primarily involve logistic regression with some ad-hoc post processing to ensure that the models are sparse and use integer coefficients \citep[see e.g., the methods described in][]{gottfredson2005mathematics}.

Our scoring systems differ from existing tools in that they directly output a predicted outcome (i.e., prisoner $i$ will recidivate) as opposed to an predicted probability of the outcome (i.e. the predicted probability that prisoner $i$ will recidivate is 90\%). The predicted probabilities from existing tools are typically converted into an outcome by imposing a threshold (i.e., classify a prisoner as ``high-risk" if the predicted probability of arrest $>70\%$). In practice, users arbitrarily pick several thresholds to translate predicted probabilities into an ordinal outcome (e.g., prisoner $i$ is ``low risk," if the predicted probability is $< 30\%$, ``medium risk" if the predicted probability is $<60\%$, and ``high risk" otherwise). These arbitrary threshholds make it difficult, if not impossible, to effectively assess the predictive accuracy of the tools \citep{hannah2013actuarial}. \citet{netter2007using}, for instance, mentions that ``the possibility of making a prediction error (false positive or false negative) using a risk tool is probable, but not easily determined." In contrast to existing tools, the scoring systems let users assess accuracy in a straightforward way (i.e., through the true positive rate and true negative rate). Further, our approach has the advantage that is can yield a scoring system that optimizes the class-based accuracy at a particular decision point (i.e., produce the model that maximizes the true positive rate, given a false-positive rate of at most $30\%$).

Our work is related to a stream of research that has aimed to leverage new methods for predictive modeling in criminology. In contrast to our work, much of the research to date has focused on improving predictive accuracy by training powerful black-box models such as random forests \citep{breimanRF} and stochastic gradient boosting \cite{friedman2002stochastic}. Random forests \citep{breimanRF}, in particular, have been used for several criminological applications, including: predicting homicide offender recidivism \citep{neuilly2011predicting}; predicting serious misconduct among incarcerated prisoners \citep{berk2006forecasting}; forecasting potential murders for criminals on probation or parole \citep{berk2009forecasting}; forecasting domestic violence and help inform court decisions at arraignment \citep{berk2014machine}. We note that not all studies in used black-box models: \citet{berk2005developing}, for instance, help the Los Angeles Sheriff's Department develop a simple and practical screener to forecast domestic violence using decision trees. More recently, \citep{goel2015precinct}, developed a simple scoring system to help the New York Police Department address stop and frisk by first running logistic regression, and then rounding the coefficients.

\section{Data and Prediction Problems}
\label{Sec::Data}
Each problem is a binary classification problem with $N=33,796$ prisoners and $P = 48$ input variables. The goal is to predict whether a prisoner will be arrested for a certain type of crime within 3 years of being released from prison. In what follows, we describe how we created each prediction problem.
%

\subsection{Database Details}
\label{Sec::DataDescription}

We derived the recidivism prediction problems in our paper from the ``Recidivism of Prisoners Released in 1994" database, assembled by the \citet{dataset}. It is the largest publicly available database on prisoner recidivism in the United States. The study tracked 38,624 prisoners for 3 years following their release from prison in 1994. These prisoners were randomly sampled from the population of all prisoners released from 15 U.S. states (Arizona, California, Delaware, Florida, Illinois, Maryland, Michigan, Minnesota, New Jersey, New York, North Carolina, Ohio, Oregon, Texas, and Virginia). The sampled population accounts for roughly two-thirds of all prisoners that were released from prison in the U.S. in 1994. Other studies that use this database include: \citet{bhati2007estimating, bhati2007estimating2,zhang2009indeterminate}. 

The database is composed of 38,624 rows and 6,427 columns, where each row represents a prisoner and each column represents a feature (i.e. a field of information for a given prisoner). The 6,427 columns consist of 91 fields that were recorded before or during release from prison in 1994 (e.g., date of birth, effective sentence length), and 64 fields that were repeatedly recorded for up to 99 different arrests in the 3 year follow-up period (e.g., if a prisoner was rearrested three times with 3 years, there would be three record cycles recorded). The information for each prisoner is sourced from record-of-arrest-and-prosecution (RAP) sheets kept by state law enforcement agencies and/or the FBI. A detailed descriptive analysis of the database was carried out by statisticians at the U.S. Bureau of Justice Statistics \citep{langan2002recidivism}. This study restricted its attention to 33,796 of the 38,624 prisoners to exclude extraordinary or unrepresentative release cases. 
To be selected for the analysis of \citet{langan2002recidivism}, a prisoner had to be alive during the 3 year follow-up period, and had to have been released from prison in 1994 for an original sentence that was at least 1 year or longer. Prisoners with certain release types -- release to custody/detainer/warrant, absent without leave, escape, transfer, administrative release, and release on appeal -- were excluded. To mirror the approach of \citet{langan2002recidivism}, we restricted our attention to the same subset of prisoners.

This dataset has some serious flaws which we point out below. To begin, many important factors that could be used to predict recidivism are missing, and many included factors are noisy enough to be excluded from our preliminary experiments. The information about education levels is extremely minimal; we do not even know whether each prisoner attended college, or completed high school. The information about courses in prison is only an indicator of whether the inmate took any education or vocation courses at all. Also, there is no family history for each prisoner (e.g., foster care) and no record of visitors while in prison (e.g., indicators of caring family members or friends). There is no information about reentry programs or employment history. While some of these factors exist, such as drug or alcohol treatment and in-prison vocational programs, the data is highly incomplete and therefore excluded from our analysis. For example, for drug treatment, less than 14\% of the prisoners had a valid entry. The rest were ``unknown." To include as many prisoners as possible, we chose to exclude factors with extremely sparse information.

\subsection{Deriving Input Variables}

We provide a summary of the $P = 48$ input variables derived from the database in Table \ref{Table::InputVariables}.  We encoded each input variable as a binary rule of the form $x_{ij} \in \{0,1\}$,  $j=1\ldots,P$, where $x_{ij} = 1$ if condition $j$ holds true about prisoner $i$. This allows a linear model to encode nonlinear functions of the original variables. We refer to input variables in the text using italicized font (e.g., \textfn{female}). All prediction problems in Table \ref{Table::PredictiveOutcomes} and all machine learning methods in Table \ref{Table::TrainingSetup} use these same input variables. 

The final set of input variables are representative of well-known risk factors for recidivism \citep{bushway2007inextricable, crow2008complexities} and have been used in risk assessment tools since 1928 \citep[see e.g.,][]{borden1928factors,us2005comparison,berk2006forecasting,baradaran2013race}.  They include:
1) information about prison release in 1994 (e.g., \textfn{time\_served}, \textfn{age\_at\_release}, \textfn{infraction\_in\_prison}); 
2) information from past arrests, sentencing, and convictions (e.g., \textfn{prior\_arrests$\geq$1}, \textfn{any\_prior\_jail\_time});%
\footnote{The \textfn{prior\_arrest} variable does not count the original crime for which they were released from prison in 1994; thus, about 12\% of the prisoners have \textfn{no\_prior\_arrests} =1 even though they were arrested at least once.}
3) history of substance abuse (e.g., \textfn{alcohol\_abuse})
4) gender (e.g., \textfn{female}).
%
These input variables are advantageous because: a) the information is easily accessible to law enforcement officials (all above information can be found in state RAP sheets); b) they do not include socioeconomic factors such as race, which would directly eliminate the potential to use these tools in applications such as sentencing.

We note that encoding the input variables as binary values presents many advantages. They produce models that are easier to understand (removing the wide range presented by continuous variables), and they avoid potential confusion stemming from coefficients of normalized inputs (for instance, after undoing the normalization for normalized coefficients, a small coefficient might be highly influential if it applies to a variable taking large values). Binarization is especially useful for SLIM as we can fit SLIM models by solving a slightly easier discrete optimization problem when the data only contains binary input variables (as discussed in Section \ref{Sec::BSLIMFormulation}). In Appendix \ref{Sec::BinaryComparison}, we explore the change in predictive accuracy if continuous variables are included and show that the changes in performance are minor for most methods. There are some exceptions; for example, CART and C5.0T experienced an improvement of $4.6\%$ for \textds{drug} and SVM RBF experienced a $7.7\%$ improvement for \textds{fatal\_violence}. Yet even for these methods, no clear improvement is seen across all problems.

\begin{table}
\caption{\label{Table::InputVariables}Overview of input variables for all prediction problems. Each variable is a binary rule of the form $x_{ij} \in \{0,1\}$. We list conditions required for $x_{ij} = 1$ under the Definition column.}
\begin{tabularx}{\textwidth}{@{}p{0mm}@{ }>{\small\itshape}l>{\small\centering}p{2cm}>{\small}X@{}}
\toprule
& {\textnormal{\textbf{Input Variable}}} & {\textbf{P}}{$(x_{ij}=1)$} & {\textbf{Definition}} \\
\toprule

& female & 0.06 & prisoner $i$ is female \\





& prior\_alcohol\_abuse & 0.20 & prisoner $i$ has history of alcohol abuse \\
& prior\_drug\_abuse & 0.16 & prisoner $i$ has history of drug abuse \\ 

\hline

& age\_at\_release$\leq$17 & 0.00 & prisoner $i$ was $\leq$17 years old at release in 1994 \\
& age\_at\_release\_18\_to\_24 & 0.19 & prisoner $i$ was 18-24 years old at release in 1994 \\
& age\_at\_release\_25\_to\_29 & 0.21 & prisoner $i$ was 25-29 years old at release in 1994 \\
& age\_at\_release\_30\_to\_39 & 0.38 & prisoner $i$ was 30-39  years old at release in 1994 \\
& age\_at\_release$\geq$40 & 0.21 & prisoner $i$ was $\geq$40 years old at release in 1994 \\

\hline

& released\_unconditional & 0.11 & prisoner $i$ released at expiration of sentence \\
& released\_conditional & 0.87 & prisoner $i$ released by parole or probation \\

\hline

& time\_served$\leq$6mo & 0.23 & prisoner $i$ served $\leq$6 months \\
& time\_served\_7\_to\_12mo & 0.20 & prisoner $i$ served 7--12 months \\
& time\_served\_13\_to\_24mo & 0.23 & prisoner $i$ served 13--24 months \\
& time\_served\_25\_to\_60mo & 0.25 & prisoner $i$ served 25--60 months \\

& time\_served$\geq$61mo & 0.10 & prisoner $i$ served $\geq$61 months \\

\hline

 & infraction\_in\_prison & 0.24 & prisoner $i$ has a record of misconduct in prison \\

\hline

& age\_1st\_arrest$\leq$17 & 0.14 & prisoner $i$ was $\leq$17 years old at 1st arrest \\

 & age\_1st\_arrest\_18\_to\_24 & 0.61 & prisoner $i$ was 18-24 years old at 1st arrest \\

 & age\_1st\_arrest\_25\_to\_29 & 0.10 & prisoner $i$ was 25-29 years old at 1st arrest \\

 & age\_1st\_arrest\_30\_to\_39 & 0.09 & prisoner $i$ was 30-39 years old at 1st arrest \\

 & age\_1st\_arrest$\geq$40 & 0.04 & prisoner $i$ was $\geq$40 years at 1st arrest \\

\hline

 & age\_1st\_confinement$\leq$17 & 0.03 & prisoner $i$ was $\leq$17 years old at 1st confinement \\

 & age\_1st\_confinement\_18\_to\_24 & 0.46 & prisoner $i$ was 18-24 years old at 1st confinement \\

 & age\_1st\_confinement\_25\_to\_29 & 0.18 & prisoner $i$ was 25-29 years old at 1st confinement \\

 & age\_1st\_confinement\_30\_to\_39 & 0.21 & prisoner $i$ was 30-39 years old at 1st confinement \\

 & age\_1st\_confinement$\geq$40 & 0.12 & prisoner $i$ was $\geq$40 years at 1st confinement\\

\hline

 & prior\_arrest\_for\_drug & 0.47 & prisoner $i$ was once arrested for drug offense\\

 & prior\_arrest\_for\_property & 0.67 & prisoner $i$ was once arrested for property offense\\

 & prior\_arrest\_for\_public\_order & 0.62 & prisoner $i$ was once arrested for public order offense\\

 & prior\_arrest\_for\_general\_violence & 0.52 & prisoner $i$  was once arrested for general violence\\

 & prior\_arrest\_for\_domestic\_violence & 0.04 & prisoner $i$ was once arrested for domestic violence\\

 & prior\_arrest\_for\_sexual\_violence & 0.03 & prisoner $i$ was once arrested for sexual violence\\

& prior\_arrest\_for\_fatal\_violence & 0.01 & prisoner $i$ was once arrested for fatal violence\\

\hline
 
& prior\_arrest\_for\_multiple\_types & 0.77 & prisoner $i$ was once arrested for multiple types of crime\\

 & prior\_arrest\_for\_felony & 0.84 & prisoner $i$ was once arrested for a felony\\ 

 & prior\_arrest\_for\_misdemeanor & 0.49 & prisoner  $i$ was once arrested for a misdemeanor\\ 

 & prior\_arrest\_for\_local\_ordinance & 0.01 & prisoner $i$ was once arrested for local ordinance\\ 

 & prior\_arrest\_with\_firearms\_involved & 0.09 & prisoner $i$ was once arrested or an incident involving firearms\\

 & prior\_arrest\_with\_child\_involved & 0.17 & prisoner $i$ was once arrested for an incident involving children \\ 

\hline

 & no\_prior\_arrests & 0.12 & prisoner $i$ has no prior arrests\\

 & prior\_arrests$\geq$1 & 0.88 & prisoner $i$ has at least 1 prior arrest\\

 & prior\_arrests$\geq$2 & 0.78 & prisoner $i$ has at least 2 prior arrests\\

 & prior\_arrests$\geq$5 & 0.60 & prisoner $i$ has at least 5 prior arrests\\

\hline

& multiple\_prior\_prison\_time & 0.43 & prisoner $i$ has been to prison multiple times\\
& any\_prior\_jail\_time & 0.47 & prisoner $i$ has been to jail at least once\\
& multiple\_prior\_jail\_time & 0.29 & prisoner $i$ has been to prison multiple times\\
& any\_prior\_probation\_or\_fine & 0.42 & prisoner $i$ has been on probation or paid a fine at least once \\
& multiple\_prior\_probation\_or\_fine & 0.22 & prisoner $i$ has been on probation or paid a fine multiple times\\
\bottomrule
\end{tabularx}
\end{table}

\subsection{Deriving Outcome Variables} 

We created a total of 6 recidivism prediction problems by encoding a binary outcome variable $y_i \in \{-1,+1\}$ such that  $y_i=+1$ if a prisoner is arrested for a particular type of crime within 3 years after being released from prison. For clarity, we refer to each prediction problem in the text using typewriter font (e.g., \textds{arrest}). We provide details on each recidivism prediction problems in Table \ref{Table::PredictiveOutcomes}. These include: an arrest for any crime (\textds{arrest}); an arrest for a drug-related offense (\textds{drug}); or an arrest for a certain type of violent offense (\textds{general\_violence}, \textds{domestic\_violence}, \textds{sexual\_violence}, \textds{fatal\_violence}).%

In the dataset, all crime types can be broken down into smaller subcategories (e.g., \textds{fatal\_violence} can be broken into 6 subcategories such as \textds{murder}, \textds{vehicular\_manslaughter}, etc.). We chose to use the broader crime categories for the sake of conciseness and clarity. Indeed, the study by \citet{langan2002recidivism} also split the crimes into the same major categories. We note that the outcomes of violent offenses are mutually exclusive, as different types of violence are treated differently within the U.S. legal system. In other words, $y_i=+1$ for \textds{general\_violence} does not necessarily imply $y_i=+1$ for \textds{domestic\_violence}, \textds{sexual\_violence}, \textds{fatal\_violence}).
\begin{table}
\caption{\label{Table::PredictiveOutcomes}Overview of recidivism prediction problems. The percentages $P(y_i=+1)$ do not add up to 100\% because a prisoner could be arrested for multiple types of crime at one time (e.g., both drug and public order offenses), and could also be arrested multiple times over the 3 year follow-up period.}
\begin{tabular}{m{0.2\textwidth}c m{0.6\textwidth}@{}m{0pt}}
\toprule 
\textbf{Prediction Problem} & \multicolumn{1}{r}{\textbf{P$(y_i=+1)$}} & \textbf{Outcome Variable}  &\\[2ex] 
\toprule

\tableds{arrest} & 59.0\% & $y_i=+1$ if prisoner $i$ is arrested for any offense within 3 years of release from prison &\\[0ex] 

\midrule

\tableds{drug} & 20.0\% & $y_i=+1$ if prisoner $i$ is arrested for drug-related offense (e.g., possession, trafficking) within 3 years of release from prison &\\[0ex] 

\midrule





\tableds{general\_violence} & 19.1\% & $y_i=+1$ if prisoner $i$ is arrested for a violent offense (e.g., robbery, aggravated assault) within 3 years of release from prison &\\[0ex] 

\midrule

\tableds{domestic\_violence} & 3.5\% & $y_i=+1$ if prisoner $i$ is arrested for domestic violence within 3 years of release from prison &\\[0ex] 

\midrule

\tableds{sexual\_violence} & 3.0\% & $y_i=+1$ if prisoner $i$ is arrested for sexual violence within 3 years of release from prison &\\[0ex] 

\midrule

\tableds{fatal\_violence} & 0.7\% & $y_i=+1$ if prisoner $i$ is arrested for murder or manslaughter within 3 years of release from prison &\\[0ex] 
\bottomrule
\end{tabular}
\end{table}

\FloatBarrier
\subsection{Relationships between Input and Output Variables}\label{Sec::SimpleInsights}
Table \ref{Table::ConditionalProbabilities} lists the conditional probabilities $P(y=1|x_j=1)$ between the outcome variable $y$ and each input variable $x_j$ for all prediction problems. Using this table, we can identify strong associations between the input and output for each prediction problem. These associations can help uncover insights into each problem and also help qualitatively validate predictive models in Section \ref{Sec::InterpretabilityResults}. 

Consider, for instance, the \textds{arrest} problem. Here, we can see that prisoners who are released from prison at a later age are less likely to be arrested (as the probability for arrest decreases monotonically as $\textfn{age\_at\_release}$ increases). This also appears to be the case for prisoners who were first confined (i.e., sent to prison or jail) at an older age (see e.g., \textfn{age\_of\_first\_confinement}). In addition, we can also see that prisoners with more prior arrests have a higher likelihood of being arrested (as the probability for arrest increases monotonically with $\textfn{prior\_arrest}$).

Similar insights can be made for crime-specific prediction problems. In \textds{drug}, for instance, we see that prisoners who were previously arrested for a drug-related offense are more likely to be rearrested for a drug-related offense (32\%) than those who were previously arrested for any other type of offense. Likewise, looking at $\textds{domestic\_violence}$, we see that the prisoners with the greatest probability of being arrested for a domestic violence crime are those with a history of domestic violence (13\%). 
\newcolumntype{t}{>{\tiny\raggedleft\arraybackslash}p{0.9cm}}
\newcommand{\dcell}[1]{\setlength{\tabcolsep}{0pt}\texttt{\begin{tabular}{r}#1 \end{tabular}}}
\begin{table}
\caption{\label{Table::ConditionalProbabilities}Table of conditional probabilities for all input variables (row) and prediction problems (columns). Each cell represents the conditional probability $P(y=+1|x=+1)$ where $x$ is the input variable that is specified in the row and $y$ is the outcome variable for the prediction problem specified in the column.}
\centering
\begin{tabular}{p{0mm}@{}>{\itshape}lrrrrrr}
\toprule
& \textnormal{\textbf{{Input Variable}}} & \multicolumn{6}{c}{\textbf{Prediction Problem}} \\
\toprule
& & \dcell{arrest} & \dcell{drug} & \dcell{general\\violence} & \dcell{domestic\\violence}  &  \dcell{sexual\\violence}  &  \dcell{fatal\\violence} \\ 
\cmidrule{3-8}

 & female & 0.54 & 0.21 & 0.11 & 0.02 & 0.01 & 0.0005\\
 & prior\_alcohol\_abuse & 0.58 & 0.18 & 0.20 & 0.04 & 0.03 & 0.01\\
 & prior\_drug\_abuse & 0.61 & 0.23 & 0.21 & 0.03 & 0.03 & 0.004\\
 
\hline

 & age\_at\_release$\leq$17 & 0.84 & 0.35 & 0.31 & 0.01 & 0.01 & 0.04\\
 & age\_at\_release\_18\_to\_24 & 0.71 & 0.24 & 0.25 & 0.04 & 0.03 & 0.01\\
 & age\_at\_release\_25\_to\_29 & 0.66 & 0.23 & 0.21 & 0.04 & 0.03 & 0.01\\
 & age\_at\_release\_30\_to\_39 & 0.59 & 0.20 & 0.17 & 0.04 & 0.03 & 0.01\\
 & age\_at\_release$\geq$40 & 0.41 & 0.12 & 0.09 & 0.02 & 0.03 & 0.003\\

\hline

 & released\_unconditional & 0.65 & 0.20 & 0.23 & 0.06 & 0.04 & 0.01 \\
 & released\_conditional & 0.58 & 0.20 & 0.17 & 0.03 & 0.03 & 0.01\\

\hline

 & time\_served$\leq$6mo & 0.67 & 0.27 & 0.19 & 0.04 & 0.03 & 0.01\\
 & time\_served\_7\_to\_12mo & 0.63 & 0.22 & 0.19 & 0.04 & 0.03 & 0.01\\
 & time\_served\_13\_to\_24mo & 0.59 & 0.20 & 0.17 & 0.04 & 0.03 & 0.01\\
 & time\_served\_25\_to\_60mo & 0.53 & 0.16 & 0.17 & 0.03 & 0.03 & 0.01\\
 & time\_served$\geq$61mo & 0.48 & 0.11 & 0.15 & 0.02 & 0.04 & 0.004\\

\hline

 & infraction\_in\_prison & 0.65 & 0.19 & 0.20 & 0.01 & 0.04 & 0.01\\

\hline 

 & age\_1st\_arrest$\leq$17 & 0.73 & 0.27 & 0.27 & 0.04 & 0.04 & 0.01\\
 & age\_1st\_arrest\_18\_to\_24 & 0.64 & 0.22 & 0.20 & 0.04 & 0.03 & 0.01\\
 & age\_1st\_arrest\_25\_to\_29 & 0.47 & 0.14 & 0.10 & 0.02 & 0.02 & 0.005\\
 & age\_1st\_arrest\_30\_to\_39 & 0.34 & 0.10 & 0.06 & 0.02 & 0.02 & 0.003\\
 & age\_1st\_arrest$\geq$40 & 0.21 & 0.05 & 0.03 & 0.01 & 0.02 & 0.002\\

\hline 

& age\_1st\_confinement$\leq$17 & 0.78 & 0.28 & 0.29 & 0.04 & 0.04 & 0.02\\
 & age\_1st\_confinement\_18\_to\_24 & 0.68 & 0.24 & 0.23 & 0.05 & 0.04 & 0.01\\
 & age\_1st\_confinement\_25\_to\_29 & 0.60 & 0.20 & 0.17 & 0.03 & 0.03 & 0.005\\
 & age\_1st\_confinement\_30\_to\_39 & 0.50 & 0.16 & 0.12 & 0.03 & 0.02 & 0.003\\
 & age\_1st\_confinement$\geq$40 & 0.34 & 0.09 & 0.07 & 0.01 & 0.02 & 0.002\\

\hline

& prior\_arrest\_for\_drug & 0.68 & 0.32 & 0.21 & 0.04 & 0.02 & 0.01\\
 & prior\_arrest\_for\_property & 0.67 & 0.24 & 0.22 & 0.04 & 0.03 & 0.01\\
 & prior\_arrest\_for\_public\_order & 0.65 & 0.24 & 0.22 & 0.04 & 0.03 & 0.01\\
 & prior\_arrest\_for\_general\_violence & 0.67 & 0.25 & 0.26 & 0.05 & 0.04 & 0.01\\
 & prior\_arrest\_for\_domestic\_violence & 0.66 & 0.21 & 0.27 & 0.13 & 0.04 & 0.01\\
 & prior\_arrest\_for\_sexual\_violence & 0.49 & 0.13 & 0.16 & 0.04 & 0.06 & 0.01\\
 & prior\_arrest\_for\_fatal\_violence & 0.54 & 0.19 & 0.21 & 0.04 & 0.03 & 0.01\\

\hline 

 & prior\_arrest\_for\_multiple\_crime\_types & 0.64 & 0.23 & 0.21 & 0.04 & 0.03 & 0.01\\
 & prior\_arrest\_for\_felony & 0.60 & 0.21 & 0.19 & 0.04 & 0.03 & 0.01\\
 & prior\_arrest\_for\_misdemeanor & 0.69 & 0.26 & 0.24 & 0.06 & 0.03 & 0.01\\
 & prior\_arrest\_for\_local\_ordinance & 0.91 & 0.29 & 0.43 & 0.15 & 0.05 & 0.02\\
 & prior\_arrest\_with\_firearms\_involved & 0.70 & 0.30& 0.27 & 0.06 & 0.03 & 0.01\\
 & prior\_arrest\_with\_child\_involved & 0.48 & 0.13 & 0.14 & 0.03 & 0.06 & 0.01\\

\hline 

 & no\_prior\_arrests & 0.32 & 0.07 & 0.08 & 0.02 & 0.02 & 0.003\\
 & prior\_arrest$\geq$1 & 0.63 & 0.22 & 0.19 & 0.04 & 0.03 & 0.01\\
 & prior\_arrest$\geq$2 & 0.66 & 0.23 & 0.20 & 0.04 & 0.03 & 0.01\\
 & prior\_arrest$\geq$5 & 0.70 & 0.25 & 0.22 & 0.04 & 0.03 & 0.01\\

\hline 

 & multiple\_prior\_prison\_time & 0.65 & 0.23 & 0.19 & 0.03 & 0.03 & 0.01\\
 & any\_prior\_jail\_time & 0.69 & 0.25 & 0.21 & 0.04 & 0.03 & 0.01\\
 & multiple\_prior\_jail\_time & 0.73 & 0.27 & 0.22 & 0.04 & 0.03 & 0.01\\
 & any\_prior\_probation\_or\_fine & 0.67 & 0.24 & 0.20 & 0.04 & 0.03 & 0.01\\
 & multiple\_prior\_probation\_or\_fine & 0.71 & 0.27 & 0.22 & 0.05 & 0.03 & 0.01\\
\bottomrule
\end{tabular}
\end{table}

\FloatBarrier
\section{Supersparse Linear Integer Models}
\label{Sec::SLIMDescription}

A \textit{Supersparse Linear Integer Model} (SLIM) is a new machine learning method for creating \textit{scoring systems} -- that is, binary classification models that only require users to add, subtract and multiply a few small numbers to make a prediction \citep[][]{ustun2015slim}. Scoring systems are widely used because they allow users to make quick predictions, without the use of a computer, and without extensive training in statistics. These models are also useful because their high degree of sparsity and integer coefficients let users easily gauge the influence of multiple input variables on the predicted outcome (see Section \ref{Sec::InterpretabilityResults} for an example). In what follows, we provide a brief overview of SLIM, and provide several new techniques to reduce the computation for problems with binary input variables. 

\subsection{Framework and Optimization Problem}
SLIM scoring systems are linear classification models of the form:
\[
\hat{y}_i = 
\begin{dcases} 
+ 1 &\mbox{ if } \sum_{j=1}^P{\lambda_j x_{ij}} > \lambda_0\\ 
- 1 &\mbox{ if } \sum_{j=1}^P{\lambda_j x_{ij}} \leq \lambda_0.\\ 
\end{dcases}
\]
Here, $\lambda_1,\ldots,\lambda_P$ represent the coefficients (i.e. the ``points" for input variables $j=1,\ldots,P$), and $\lambda_0$ represents an intercept (i.e. the ``threshold score" that has to be surpassed to predict $\hat{y}_i=+1$).  

The values of the coefficients are determined from data by solving a discrete optimization problem that has the following form:
%
\begin{align}\label{form1}
\begin{split}
\min_{\lambdab} & \qquad \frac{1}{N}\sum_{i=1}^N \indic{y_i\neq\hat{y_i}} + C_0 \sum_{j=1}^P{\indic{\lambda_j \neq 0}} + \epsilon  \sum_{j=1}^P{|\lambda_j|} \\
\st &  \qquad (\lambda_0,\lambda_1,...,\lambda_P) \in \Lset. 
\end{split}
\end{align}
Here, the objective directly minimizes the error rate $\frac{1}{N}\sum_{i=1}^N \indic{y_i\neq\hat{y_i}}$ and directly penalizes the number of non-zero terms $\sum_{j=1}^P{\indic{\lambda_j \neq 0}}$. The constraints restrict coefficients to a finite set such as $\Lset   =\{-10,\ldots,10\}^{P+1}$. Optionally, one could include additional operational constraints on the accuracy and interpretability of the desired scoring system. 

The objective includes a \textit{tiny} penalty on the absolute value of the coefficients to restrict coefficients to coprime values without affecting accuracy or sparsity. To illustrate the use of this penalty, consider a classifier such as $\hat{y}=\sign{x_1 + x_2}$. If SLIM only minimized the misclassification rate and the number of terms (the first two terms of the objective), then $\hat{y}=\sign{2 x_1 + 2 x_2}$ would have the same objective value as $\hat{y}=\sign{x_1 + x_2}$ because it makes the same predictions and has the same number of non-zero coefficients. Since coefficients are restricted to a discrete set, we use this \textit{tiny} penalty on the absolute value of these coefficients so that SLIM chooses the classifier with the smallest (coprime) coefficients, $\hat{y} = \sign{x_1+x_2}$. 

The $C_0$  parameter represents the maximum accuracy that SLIM is willing to sacrifice to remove a feature from the optimal scoring system. If, for instance, $C_0$ is set within the range $(1/N,2/N)$, we would sacrifice the accuracy of one observation to have a model with one fewer feature. Given $C_0$, we can set the $\ell_1$-penalty parameter $\epsilon$ to any value $$0 < \epsilon < \frac{\min(1/N, C_0) }{ \max_{\{\lambda_j\}_j\in\mathcal{L}}{\sum_{j=1}^P{|\lambda_j|}}}$$  so that it does not affect the accuracy or sparsity of the optimal classifier, but only induces the coefficients to be coprime for the features that are selected.

SLIM differs from traditional machine learning methods because it directly optimizes accuracy and sparsity without making approximations that other methods make for scalability (e.g., controlling for accuracy using convex surrogate loss functions). By avoiding these approximations, SLIM sacrifices the ability to fit a model in seconds or in a way that scales to extremely large datasets. In return, however, it gains the ability to fit models that are highly customizable, since one could directly encode a wide range of operational constraints into its integer programming formulation. 
In this paper, we primarily make use of a simple constraint to limit the number of non-zero coefficients, however, it is also natural to incorporate constraints on class-specific accuracy, structural sparsity, and prediction \citep[see][]{ustun2015slim}. 

In this paper we trained the following version of SLIM, which is different than (\ref{form1}) in that it includes class weights, and has specific constraints on the coefficients:
\begin{align}
\label{Eq::SLIMRecidivismFormulation}
\begin{split}
\min_{\lambdab} & \frac{\wplus}{N}\sum_{i\in\iplus}\indic{y_i\neq\hat{y_i}} + \frac{\wminus}{N}\sum_{i\in\iminus}\indic{y_i\neq\hat{y_i}} + C_0 \sum_{j=1}^P{\indic{\lambda_j \neq 0}} + \epsilon \sum_{j=1}^P{|\lambda_j|} \\ 
\st & \qquad \sum_{j=1}^P{\indic{\lambda_j \neq 0}} \leq 8 \\ 
    & \qquad \lambda_j \in \{-10,\ldots,10\} \for \mprange{j}{1}{P} \\ 
    & \qquad \lambda_0 \in \{-100,\ldots,100\}. 
\end{split}
\end{align}
In the formulation above, the constraints restrict each coefficient $\lambda_j$ to an integer between $-10$ and $10$, the threshold $\lambda_0$ to an integer between $-100$ and $100$, the number of non-zero to at most 8 \citep[i.e., within the range of cognitive entities humans could handle, as per][]{miller1956magical}. The parameters $\wplus$ and $\wminus$ are class-based weights that control the accuracy on positive and negative examples. We typically choose values of $\wplus$ and $\wminus$ such that $\wplus + \wminus = 2$, so that we recover an error-minimizing formulation by setting $\wplus = \wminus =1$. The $C_0$ parameter was set to a sufficiently small value so that SLIM would not sacrifice accuracy for sparsity: given $\wplus$ and $\wminus$, we can set $C_0$ to any value $$0 < C_0 < \min\{\wminus,\wplus\}/{(N\times P)}$$ to ensure this condition. The $\epsilon$ parameter was set to a sufficiently small value so that SLIM would produce a model with coprime coefficients without affecting accuracy or sparsity: given $\wplus$, $\wminus$ and $C_0$, we can set $\epsilon$ to any value $0 < \epsilon < C_0/\max{\sum_{j=1}^P{|\lambda_j|}}$ to ensure this condition. 

%
%
%

\subsection{General SLIM IP Formulation}\label{Sec::SLIMIPFormulation}
Training a SLIM scoring system requires solving an integer programming (IP) problem using a solver such as CPLEX, Gurobi, or CBC.  In general, we use the following IP formulation to recover the solution to the optimization problem \eqref{Eq::SLIMRecidivismFormulation}:
\begin{subequations}\label{IP:SLIM}
\begin{equationarray}{crcl>{\qquad}l>{\qquad}r}
\min_{\lambdab,\bf{\loss},\bf{\Phi},\bm{\alpha},\bm{\beta}} &\frac{1}{N}\sum_{i=1}^N \loss_i &  + &  \sum_{j=1}^{P} \IntPen_j  \notag \\
\st          & M_i \loss_i                  & \geq & \gamma -\sum_{j=0}^P y_i \lambda_j x_{i,j}       &\mprange{i}{1}{N} & \mpdes{error on $i$} \label{Con::SLIMLoss} \\
& \IntPen_j & = & C_0\alpha_j + \epsilon\beta_j &\mprange{j}{1}{P}& \mpdes{penalty for coef $j$} \label{Con::SLIMIntPenalty} \\
& -\Lambda_j\alpha_j  & \leq & \lambda_j \leq \Lambda_j\alpha_j  &\mprange{j}{1}{P} & \mpdes{$\lzero$-norm} \label{Con::SLIML0Norm} \\
& -\beta_j &  \leq & \lambda_j  \leq \beta_j &\mprange{j}{1}{P} & \mpdes{$\lone$-norm} \label{Con::SLIML1Norm} \\
& \lambda_j & \in &  \Z \cap [-\Lambda_j,\Lambda_j] &  \mprange{j}{0}{P} & \mpdes{coefficient set} \notag \\ 
& \loss_i & \in & \B &  \mprange{i}{1}{N} & \mpdes{loss variables} \notag  \\
& \IntPen_j  & \in & \R_+  & \mprange{j}{1}{P} & \mpdes{penalty variables} \notag \\
& \alpha_j  & \in & \B  & \mprange{j}{1}{P} & \mpdes{$\lzero$ variables} \notag \\
& \beta_j    & \in & \R_+ & \mprange{j}{1}{P}. & \mpdes{$\lone$ variables} \notag
\end{equationarray}
\end{subequations}
The constraints in \eqref{Con::SLIMLoss} compute the error rate by setting the \textit{loss variables} $\loss_i = \indic{y_i \lambdab^T\xb_i \leq 0}$ to $1$ if a linear classifier with coefficients $\lambdab$ misclassifies example $i$ (or is close to misclassifying it, depending on the margin $\gamma$). This is a \textit{Big-M constraint} for the error rate that depends on scalar parameters $\gamma$ and $M_i$ \cite[see e.g.,][]{rubin2009mixed}. The value of $M_i$ represents the maximum score when example $i$ is misclassified, and can be set as $M_i = \max_{\lambdab \in \Lset} (\gamma - y_i\lambdab^T\xb_i)$ which is easy to compute since $\Lset$ is finite. The value of $\gamma$ represents the margin, and the objective is penalized when points are either incorrectly classified, or within $\gamma$ of the decision boundary. How close a point is to the decision boundary (or whether it is misclassified) is determined by $y_i\lambdab^T\xb_i$. When the features are binary, and since the coefficients are integers, $\gamma$ can naturally be set to any value between 0 and 1. (In other cases, we can set $\gamma=0.5$ for instance, which makes an implicit assumption on the values of the features.) 
The constraints in \eqref{Con::SLIMIntPenalty} set the total penalty for each coefficient to $\IntPen_j = C_0 \alpha_j + \epsilon \beta_j$, where $\alpha_j := \indic{\lambda_j\neq 0}$ is defined by Big-M constraints in \eqref{Con::SLIML0Norm}, and $\beta_j := |\lambda_j|$ is defined by the constraints in \eqref{Con::SLIML1Norm}. We denote the largest absolute value of each coefficient as $\Lambda_j := \max_{\lambda_j\in\Lset_j} |\lambda_j|$.

Restricting coefficients to a finite set results in significant practical benefits for the SLIM IP formulation, especially in comparison to other IP formulations that minimize the 0--1-loss and/or penalize the $\ell_0$-norm. Without the restriction of $\lambda$ to a bounded set, we would not have a natural choice for the Big-M constant, which means the user chooses one that is very large, leading to a less efficient formulation \citep[see e.g.,][]{wolsey1998integer}. 
For SLIM, the Big-M constants used to compute the 0--1 loss in constraint \eqref{Con::SLIMLoss} is bounded as $M_i \leq \max_{\lambdab \in \Lset} (\gamma - y_i\lambdab^T\xb_i)$, and the Big-M constant used to compute the $\ell_0$-norm in constraints \eqref{Con::SLIML0Norm} is bounded as $\Lambda_j \leq \max_{\lambda_j\in\Lset_j} |\lambda_j|$. Bounding these constants lead to a tighter LP relaxation, which narrows the integrality gap, and improves the ability of commercial IP solvers to obtain a proof of optimality more quickly. 

\subsection{Improved SLIM IP Formulation}\label{Sec::BSLIMFormulation}
%
The following formulation provides a tighter relaxation of the IP which reduces computation. It relies on the fact that when the input variables are binary, we are likely to get repeated feature values among observations.
\begin{subequations}
\begin{equationarray}{crcl>{\qquad}lr}
\min_{\lambdab,\bf{\loss},\bf{\Phi},\bm{\alpha},\bm{\beta}} &\frac{\wplus}{N}\sum_{s \in \mathcal{S}} n_s \loss_s & + & \frac{\wminus}{N}\sum_{t \in \mathcal{T}} n_t \loss_t  + \sum_{j=1}^{P} \IntPen_j  \notag \\
\st   & M_s \loss_s                  & \geq & 1 -\sum_{j=0}^P \lambda_j x_{s,j}       &s \in \mathcal{S} & \mpdes{error on $s$} \label{Con::BSLIMLossPos} \\
& M_t \loss_t                  & \geq & \sum_{j=0}^P \lambda_j x_{t,j}       &t \in \mathcal{T} & \mpdes{error on $t$} \label{Con::BSLIMLossNeg} \\
& 1                 & = & \loss_s + \loss_t   &\forall s,t: \xb_s = \xb_t, y_s = -y_t & \mpdes{conflicting labels} \label{Con::BSLIMConflict} \\
& \IntPen_j & = & C_0\alpha_j + \epsilon\beta_j &\mprange{j}{1}{P}& \mpdes{penalty for coef $j$} \label{Con::BSLIMIntPenalty} \\
& -\Lambda_j\alpha_j  & \leq & \lambda_j \leq \Lambda_j\alpha_j  &\mprange{j}{1}{P} & \mpdes{$\lzero$-norm} \label{Con::BSLIML0Norm} \\
& -\beta_j &  \leq & \lambda_j  \leq \beta_j &\mprange{j}{1}{P} & \mpdes{$\lone$-norm} \label{Con::BSLIML1Norm} \\
& \lambda_j & \in & \Z \cap [-\Lambda_j,\Lambda_j]&  \mprange{j}{0}{P} & \mpdes{coefficient set} \notag \\ 
& \loss_s, \loss_t & \in & \B &  s\in\mathcal{S} ~ t\in\mathcal{T} & \mpdes{loss variables} \notag  \\
& \IntPen_j  & \in & \R_+  & \mprange{j}{1}{P} & \mpdes{penalty variables} \notag \\
& \alpha_j  & \in & \B  & \mprange{j}{1}{P} & \mpdes{$\lzero$ variables} \notag \\
& \beta_j    & \in & \R_+ & \mprange{j}{1}{P}. & \mpdes{$\lone$ variables} \notag
\end{equationarray}
\end{subequations}
The main difference between this formulation and the one in \eqref{IP:SLIM} is that we compute the error rate of the classifier using loss constraints that are expressed in terms of the number of \textit{distinct} points in the dataset. Here, the set $\mathcal{S}$ represents the set of distinct points with positive labels, and the set $\mathcal{T}$ represents the set of distinct points with negative examples. The parameters $n_s$ (and $n_t$) count the number of times a point of type $s$ (or $t$) are found in the original dataset so that $\sum_s{n_s} = \sum_{i=1}^N{\indic{y_i=+1}}$, $\sum_t{n_t} = \sum_{i=1}^N{\indic{y_i=-1}}$, and $N = \sum_s{n_s}  + \sum_t{n_t}$. 

The main computational benefits of this formulation are due to the fact that: (i) we can reduce the number of loss constraints by counting the number of repeated rows in the dataset; and (ii) we can directly encode a lower bound on the error rate by counting the number of points $s,t$ with identical feature but opposite labels (i.e., $\xb_s$ = $\xb_t$ but $y_s = -y_t$). Here (i) reduces the size of the problem that we pass to an IP solver, and (ii) produces a much stronger lower bound on the 0--1 loss (in comparison to the LP relaxation), which speeds up the progress of branch-and-bound type algorithms. Note that it would be possible to use this formulation on a dataset without binary input variables, though it would not necessarily be effective because it could be much less likely for a dataset to contain repeated rows in such a setting.

Another subtle benefit of this formulation is that the margin for the negative points is 0 while the margin for the positive points is 1. This means that for positive points, we have a correct prediction if and only if the score $\geq 1$. For negative points, we have a correct prediction if and only if the score $\leq 0$. This provides a slight computational advantage since the negative points do not need to have scores below -1 to be correctly classified, which reduces the size of the Big-M parameter and the coefficient set. For instance, say we would to produce a linear model that encode: ``predict rearrest unless $a_1$ or $a_2$ are true."  Using the previous formulation with the margin of $\gamma \in (0,1)$ on both positives and negatives, the optimal SLIM classifier would be: ``rearrest = sign$(1 - 2a_1 - 2a_2)$." In contrast, the margin of the current formulation is: ``rearrest = sign$(1 - a_1 - a_2)$", which uses smaller coefficients, and produces a slightly simpler model.

%
\subsection{Active Set Polishing}\label{Sec::SolutionPolishing}
On large datasets,  IP solvers may take long time to produce an optimal solution or provider users with a certificate of optimality. Here, we present a \textit{polishing} procedure that can be used to improve the quality of solutions locally. For a fixed set of features, this procedure optimizes the values of coefficients. 

The polishing procedure takes as input a feasible set of coefficients from the SLIM IP $\lambdab^{\text{feasible}}$, and returns a polished set of coefficients $\lambdab^{\text{polished}}$ by solving a a simpler IP formulation shown in \eqref{Eq::Polishing}. The polishing IP only optimizes the coefficients of features that belong to the \textit{active set} of $\lambdab^\text{feasible}$: that is, the set of features with nonzero coefficients $\mathcal{A} := \left\{j: \lambda^\text{feasible}_j \neq 0 \right\}$. The coefficients for features that do not belong to the active set are fixed to zero so that $\lambda_j = 0$ for $j \notin \mathcal{A}$. In this way, the optimization no longer involves feature selection, and the formulation becomes much easier to solve.
\begin{subequations}\label{Eq::Polishing}
\begin{equationarray}{crcl>{\qquad}lr}
\min_{\lambdab,\bf{\loss},\bf{\Phi},\bm{\alpha},\bm{\beta}} &\frac{\wplus}{N}\sum_{s \in \mathcal{S}} n_s \loss_s & + & \frac{\wminus}{N}\sum_{t \in \mathcal{T}} n_t \loss_t \\
\st   & M_s \loss_s                  & \geq & 1 -\sum_{j \in \mathcal{A}} \lambda_j x_{s,j}       &s \in \mathcal{S} & \mpdes{error on $s$} \label{Con::PSLIMLossPos} \\
& M_t \loss_t                  & \geq &\sum_{j \in \mathcal{A}} \lambda_j x_{t,j}       &t \in \mathcal{T} & \mpdes{error on $t$} \label{Con::PSLIMLossNeg} \\
& 1                 & = & \loss_s + \loss_t   &\forall s,t: \xb_s = \xb_t, y_s = -y_t & \mpdes{conflicting labels} \label{Con::PSLIMConflict} \\
& \lambda_j & \in &   \Z \cap [-\Lambda_j,\Lambda_j]&{j \in \mathcal{A}} & \mpdes{coefficient set} \notag \\ 
& \loss_s, \loss_t & \in & \B &  s\in\mathcal{S} ~ t\in\mathcal{T}. & \mpdes{loss variables} \notag  
\end{equationarray}
\end{subequations}
The polishing IP formulation is especially fast to solve to optimality for classification problems with binary input variables because this limits the number of loss constraints. Say for instance that we wish to polish a set of coefficients with only 5 nonzero variables, then there are at most $|\{-1,+1\}| \times |\{0,1\}^5| = 64$ possible unique data points, and thus the same number of possible loss constraints. 

In our experiments in Section \ref{Sec::Results}, we use the polishing procedure on all of the feasible solutions we find from the earlier formulation. In all cases, we can solve the polishing IP to optimality within a few seconds (i.e. a MIPGAP of 0.0\%). 

%

\section{Experimental Results}\label{Sec::Results}

In this section, we compare the accuracy and interpretability of recidivism prediction models from SLIM to models from 8 other popular classification methods. In Section \ref{Sec::Methodology}, we explain the experimental setup used for all the methods. In Section \ref{Sec::AccuracyComparison}, we compare the predictive accuracy of the methods with the AUC values and ROC curves. In Section \ref{Sec::tradeoffsAccuracyInterpretability} and \ref{Sec::InterpretabilityResults}, we evaluate the interpretability of the models. Finally, in Section \ref{Sec::ScoringSystems}, we present the scoring systems generated by SLIM. 

\subsection{Methodology}
\label{Sec::Methodology}

In what follows we discuss cost-sensitive classification for imbalanced problems, provide an overview of techniques.

\subsubsection{Evaluating Predictive Accuracy for Imbalanced Problems}
The majority of classification problems that we consider are \textit{imbalanced}, where the data contain a relatively small number of examples from one class and a relatively large number of examples from the other. 

Imbalanced problems necessitate changes in the way that we evaluate the performance of classification models. Consider, for instance, a heavily imbalanced problem such as \textds{fatal\_violence} where only P$(y_i=+1) = 0.7\%$ of individuals are arrested within 3 years of being released from prison. In this case, a method that maximizes overall classification accuracy is likely to produce a trivial model that predicts no one will be arrested for fatal offenses -- a result that is not surprising given that the trivial model is 99.3\% accurate on the overall population. Unfortunately, this model will never be able to identify individuals that will be arrested for a fatal offense, and therefore be 0\% accurate on the population of interest.

To provide a measure of classification model performance on imbalanced problems, we assess the accuracy of a model on the positive and negative classes separately. In our experiments, we report the class-based accuracy of each model using the \textit{true positive rate} (TPR), which reflects the accuracy on the positive class, and the \textit{false positive rate} (FPR), which reflects the error rate on negative class. For a given classification model, we compute these quantities as
\begin{align*}
TPR = \frac{1}{\nplus} \sum_{i\in\iplus} \indic{\hat{y}_i=+1} \;\;\textrm{ and }\;\; FPR = \frac{1}{\nminus} \sum_{i\in\iminus} \indic{\hat{y}_i=+1},
\end{align*}
where $\hat{y}_i$ denotes the predicted outcome for example $i$, $\nplus$ denotes the number of examples in the positive class $\iplus = \{i:y_i = +1\}$, and $\nminus$ denotes the number of examples from the negative class $\iminus = \{i:y_i = -1\}$. Ideally, a classification model should have high TPR and low FPR (i.e., TPR close to 1 and FPR = 0). 

Most classification methods can be adapted to yield a model that is more accurate on the positive class, but only if we are willing to sacrifice some accuracy on examples from the negative class, and vice-versa. To illustrate the trade-off of classification accuracy between positive and negative classes, we plot all models produced by a given method as points on a \textit{receiver operating characteristic} (ROC) curve, which plots the TPR on the vertical axis and the FPR on the horizontal axis. Having constructed an ROC curve, we then assess the \textit{overall} performance of each method by calculating the \textit{area under the ROC curve} (AUC).\footnote{We note that AUC is a summary statistic that is frequently misused in the context of classification problems. It is true that a method that with AUC = 1 always produces models that are more accurate than a method with AUC = 0. Other than this simple case, however, it is not possible to state that a method with high AUC always produces models that are more accurate than a method with low AUC.}
A detailed discussion of ROC analysis in recidivism prediction can be found in the work of \citet{maloof2003learning}.

\subsubsection{Fitting Models over the Full ROC Curve using a Cost-Sensitive Approach}

Different applications require predictive models at different points of the ROC curve. Models for sentencing, for example, need low FPR in order to avoid predicting that a low-risk individual will reoffend. Models for screening, however, need high TPR in order to capture as many high-risk individuals as possible.  In our experiments, we use a \textit{cost-sensitive approach} to produce classification models at different points of the ROC curve \citep[see e.g.,][]{berk2010balancing,berk2011asymmetric}. This approach involves controlling the accuracy on the positive and negative classes by tuning the misclassification costs for examples in each class. 
In what follows, we denote the misclassification cost on examples from the positive and negative classes as $\wplus$ and $\wminus$, respectively.  As we increase $\wplus$, the cost of making a mistake on a positive example increases, and we expect to obtain a model that classifies the positive examples more accurately (i.e. with higher TPR). 
We choose $\wplus$ and $\wminus$ so that $\wplus +\wminus = 2$. Thus, when $\wplus=2$, we obtain a trivial model that predicts $\hat{y}_i=+1$ and attains TPR = 1. When $\wplus=0$, we obtain a trivial model that predicts $\hat{y}_i=-1$ that attains FPR = 0.

%
%
%
\subsubsection{Choice of Classification Methods}
We compared SLIM scoring systems to models produced by eight popular classification methods, including those previously used for recidivism prediction (see Section \ref{Sec::RelatedWork}) or those that ranked among the ``top 10 algorithms in data mining" \citep{top10}. In choosing these methods, we restricted our attention to methods that have publicly-available software packages, and allow users to specify misclassification costs for positive and negative classes. Our final choice of methods includes:
\begin{itemize}
\item\textbf{C5.0 Trees and C5.0 Rules}: C5.0 is an updated version of the popular C4.5 algorithm \citep{quinlan2014c4,kuhn2013applied} that can create decision trees and rule sets.
\item\textbf{Classification and Regression Trees (CART)}: CART is a popular method to create decision trees through recursive partitioning of the input variables \citep{breiman1984classification}. 
\item\textbf{$L_1$ and $L_2$-Penalized Logistic Regression}: Variants of logistic regression that penalize the coefficients to prevent overfitting \citep{friedman2010glmnet}. $L_1$-penalized methods are typically used to create linear models that are sparse \citep{tibshirani1996regression,hesterberg2008least}. The $L_2$ regularized methods are called ``ridge" and are not generally sparse.
\item\textbf{Random Forests}: A popular black-box method that makes predictions using a large ensemble of weak classification trees. The method was originally developed by \citet{breimanRF} but is widely used for recidivism prediction \citep[see e.g.,][]{berk2009forecasting,ritter2013predicting}.
\item\textbf{Support Vector Machines}: A popular black-box method for non-parametric linear classification. The Radial Basis Function (RBF) kernel lets the method to handle classification problems where the decision-boundary may be non-linear \cite[see e.g.,][]{cristianini2000introduction,berk2014forecasts}.
\item\textbf{Stochastic Gradient Boosting}: A popular black-box method that create prediction models in the form of an ensemble of weaker prediction models \citep{friedman2001greedy, freund1997decision}.
%
%
\end{itemize}
\subsubsection{Details on Experimental Design, Parameter Tuning, and Computation}
\label{ExperimentalDetails}

We summarize the methods, software, and settings that we used in our experiments in Table \ref{Table::TrainingSetup}. 

For each of the 6 recidivism prediction problems and each of the 9 methods, we constructed ROC curves by running the algorithm with 19 values of $\wplus$. The values of $\wplus$ were chosen to produce models across the full ROC curves. By default, we chose values of $\wplus \in \{0.1,0.2,\ldots,1.9\}$ and set $\wminus = 2 - \wplus$. These values of $\wplus$ were inappropriate for problems with a significant class imbalance as all methods produced trivial models. Thus, for significantly imbalanced problems, such as \textds{domestic\_violence} and \textds{sexual\_violence}, we used values of $\wplus \in \{1.815,1.820,\ldots,1.995\}$. For \textds{fatal\_violence}, which was extremely imbalanced, we used $\wplus \in \{1.975,1.976,\ldots,1.995\}$.

This setup requires us to produce a total of 1,026 recidivism prediction models (6 recidivism problems $\times$ 9 methods $\times$ 19 imbalance ratios). Each of the 1,026 models were built on a training set and their performance was assessed out-of-sample. In particular, 1/3 of the data was reserved as the \textit{test set}. The remaining 2/3 of the data was the \textit{training set}. During training, we used 5-fold nested cross-validation (5-CV) for parameter tuning. Explicitly, the training data were split into 5 folds, and one of those 5 was reserved as the validation fold. The validation fold was rotated in order to select free parameter values, and a \textit{final model} was trained on the full training set (2/3) with the selected parameter values and its performance was assessed on the test set (1/3). The folds were generated once to allow for comparisons across methods and prediction problems. The parameters were chosen during nested cross validation to minimize the mean weighted 5-CV validation error on the training set. \textit{Having obtained a set of 19 different models for each method and each problem, we then constructed an ROC curve for that method on that problem by plotting the test TPR and test FPR of the 19 final models.}

We trained all baseline methods using publicly available packages in R 3.2.2 \citep{Rcitation} without imposing any time constraints. In comparison, we trained SLIM by solving integer programming problems (IP) with the CPLEX 12.6 API in MATLAB 2013a. We solved each IP through the following procedure: (i) we trained the solver on the formulation in Section \ref{Sec::BSLIMFormulation} for a total of 4 hours on a local computing cluster with 2.7GHz CPUs. Each time we solved a IP we kept 500 feasible solutions, and polished them using the formulation in Section \ref{Sec::SolutionPolishing}. We then used the same nested cross-validation procedure as the other methods to tune the number of terms in the final model. Polishing all 500 solutions took less than one minute of computing time. Thus, the total number of optimization problems we solved were 500 polishing IP's $\times$ (5 folds +  1 final model) $\times$ 6 problems $\times$ 19 values of $\wplus$ = 342,000 integer programming problems.

\begin{table}
\caption{\label{Table::TrainingSetup}Methods, software and free parameters used to train models for all 6 recidivism prediction problems. We ran each method for 19 values of $\wplus$ and all combinations of free parameters listed in the table. For each value of $\wplus$, we selected the model that minimized the mean weighted 5-CV validation error. The values of $\wplus$ are problem-specific (see Section \ref{ExperimentalDetails} for details)}
\centering{\resizebox{\textwidth}{!} {
\begin{tabular}{lccl@{}}

\toprule 
\textbf{Method} & \textbf{Acronym} & \textbf{Software} & \textbf{\cell{l}{Free Parameters and Settings}} \\ 

\toprule

CART Decision Trees & CART & \cell{c}{\pkg{rpart}\\\citep{the2012rpart}} & \cell{l}{minSplit $\in (3,5,10,15,20)$ $\times$ \\ CP $\in (0.0001, 0.001, 0.01)$}\\ 

\midrule

C5.0 Decision Trees & C5.0T & \cell{c}{\pkg{c50}\\\citep{kuhn2012c50}} & default settings\\ 

\midrule

C5.0 Decision Rules  & C5.0R & \cell{c}{\pkg{c50}\\\citep{kuhn2012c50}} & default settings\\ 

\midrule

\cell{l}{Logistic Regression\\($L_1$-Penalty)} & Lasso & \cell{c}{\pkg{glmnet}\\\citep{friedman2010glmnet}} & \cell{l}{100 values of $L_1$-penalty chosen by \pkg{glmnet}} \\

\midrule

\cell{l}{Logistic Regression\\($L_2$-Penalty)} & Ridge & \cell{c}{\pkg{glmnet}\\\citep{friedman2010glmnet}} & \cell{l}{100 values of $L_2$-penalty chosen by \pkg{glmnet}} \\

\midrule

Random Forests  & RF & \cell{c}{\pkg{randomForest}\\\citep{randomForest}} & \cell{l}{sampsize $\in (0.632N, 0.4N, 0.2N)$ $\times$ \\ nodesize $\in (1,5,10,20)$\\ with unbounded tree depth}\\ 

\midrule

\cell{l}{Support Vector Machines\\(Radial Basis Kernel)} & SVM RBF & \cell{c}{\pkg{e1071}\\\citep{meyer2012e1071}} & \cell{l}{$C \in (0.01, 0.1, 1, 10)$ $\times$ \\ $\gamma \in (\frac{1}{10P}, \frac{1}{5P}, \frac{1}{2P}, \frac{1}{P}, \frac{2}{P}, \frac{5}{P}, \frac{10}{P})$} \\ 

\midrule

\cell{l}{Stochastic Gradient Boosting\\(Adaboost)} & SGB & \cell{c}{\pkg{gbm}\\\citep{ridgeway2006gbm}} & \cell{l}{shrinkage $\in (0.001,0.01,0.1)$ $\times$ \\ interaction.depth $\in (1, 2, 3, 4)$ $\times$ \\ ntrees $\in (100, 500, 1500, 3000)$} \\  

\midrule

SLIM Scoring Systems & SLIM  & \cell{c}{\pkg{CPLEX 12.6}\\\citep{berk_ustun_2016_47964}} & \cell{l}{$C_0$ and $\epsilon$ set to find most accurate model with $\leq$ 8 coefficients\\where $\lambda_0 \in \{-100,\ldots,100\}$ and $\lambda_j \in \{-10,\ldots,10\}$ \\ } \\
\bottomrule 
\end{tabular}
}
}
\end{table}
%

%
%
%
\subsection{Observations on Predictive Accuracy}
\label{Sec::AccuracyComparison}

We show ROC curves for all methods and prediction problems in Figure \ref{Figure::ROCCurves1} and summarize the test AUC of each method in Table 
\ref{Table::testAUC}. Tables with the training and 5-CV validation AUC's for all methods are included in Appendix \ref{Sec::AdditionalResultsOnPredictivePerformance}. 

We make the following important observations, which we believe carry over to a large class of problems beyond recidivism prediction:

\begin{itemize}

\item All methods did well on the general recidivism prediction problem \textds{arrest}. In this case, we observe only small differences in predictive accuracy of different methods: all methods other than CART attain a test AUC above 0.72; the highest test AUC of 0.73 was achieved by SGB, Ridge, and RF. This multiplicity of good models reflects the \textit{Rashomon effect} of \citet{breimanRF}.

\item Major differences between methods appeared in their performance on imbalanced prediction problems. We expected different methods to respond differently to changes in the misclassification costs, and therefore trained each method over a large range of possible misclassification costs. Even so, it was difficult (if not impossible) to tune certain methods to produce models at certain points of the ROC curve (see e.g., problems with significant imbalance, such as \textds{fatal\_violence}). 

\item SVM RBF, SGB, Lasso and Ridge were able to produce accurate models at different points on the ROC curve for most problems. SGB usually achieved the highest AUC on most problems (e.g., \textds{arrest}, \textds{drug}, \textds{general\_violence}, \textds{domestic\_violence}, \textds{fatal\_violence}). Lasso, Ridge, and SVM RBF often produce comparable AUCs. We find that these methods respond well to cost-sensitive tuning, but it is difficult to tune the misclassification costs for highly imbalanced problems, such as \textds{fatal\_violence}, to get models at specific points on the ROC curve.

\item C5.0T, C5.0R and CART were unable to produce accurate models at different points on the ROC curve on any imbalanced problems. We found that these methods do not respond well to cost-sensitive tuning. The issue becomes markedly more severe as problems become more imbalanced.  For \textds{drug} and \textds{general\_violence}, for instance, these methods could not produce models with high TPR. For \textds{fatal\_violence}, \textds{sexual\_violence}, and \textds{domestic\_violence}, these methods almost always produced trivial models that predict $y=-1$ (resulting in AUCs of 0.5). This result may be attributed to the greedy nature of the algorithms used to fit the trees, as opposed to the use of tree models in general. The issue is unlikely to be software-related as it affects both C5.0 and CART, and has been observed by others \citep[see e.g.,][]{goh2014box}. This problem might not occur if trees were better optimized.

\item  
In general, SLIM produced models that are close to or on the efficient frontier of the ROC curve, despite being restricted to a relatively small class of simple linear models (at most 8 non-zero coefficients from -10 to 10). 
Even on highly imbalanced problems such as \textds{domestic\_violence} and \textds{sexual\_violence}, it responds well to changes in misclassification costs (as expected, by nature of its formulation).

\end{itemize}


In addition to predictive accuracy, we also examine the risk calibration of the models. Figure \ref{Figure::rearrestCalibPlot} show the risk calibration for \textds{arrest}, constructed using the binning method from \citet{zadrozny2002transforming}. We include calibration plots for all other problems in Appendix \ref{Appendix::ModelBasedComparisons}. We see that SLIM is well-calibrated, even though there is no reason it should be; it is a decision-making tool, not a risk assessment tool. For \textds{arrest}, Lasso and Ridge are well-calibrated; however, they lose this quality once we consider only sparse models (see Appendix \ref{Appendix::LossInCalibration}). This property would also be lost if the Lasso and Ridge coefficients were rounded.  

\newcommand{\tabulartitle}[3]{\begin{tabularx}{#1}{>{\centering}X>{\centering}X}{\hspace{1cm}#2}&{\hspace{1cm}#3}\end{tabularx}}
\setlength{\fwidth}{0.5\textwidth}
\newcommand{\rocinclude}[1]{\includegraphics[trim=0cm 0.8cm 0cm 0.6cm,clip,width=.95\fwidth]{#1}}
\begin{figure}

\setlength{\tabcolsep}{3pt}
\centering
\tabulartitle{\textwidth}{\textds{arrest}}{\textds{drug}} \\ 
\begin{tabular}{lr}
\rocinclude{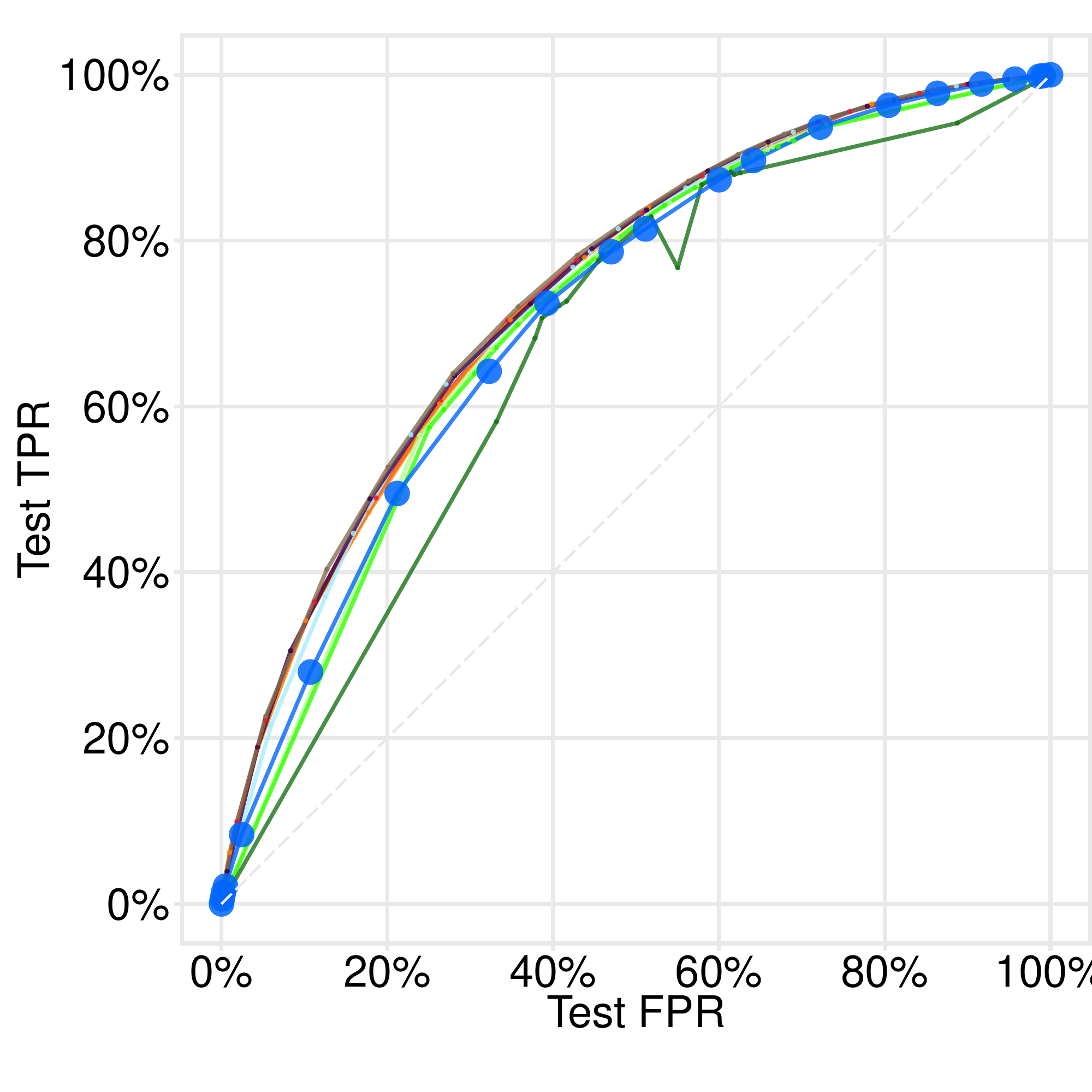} &
\rocinclude{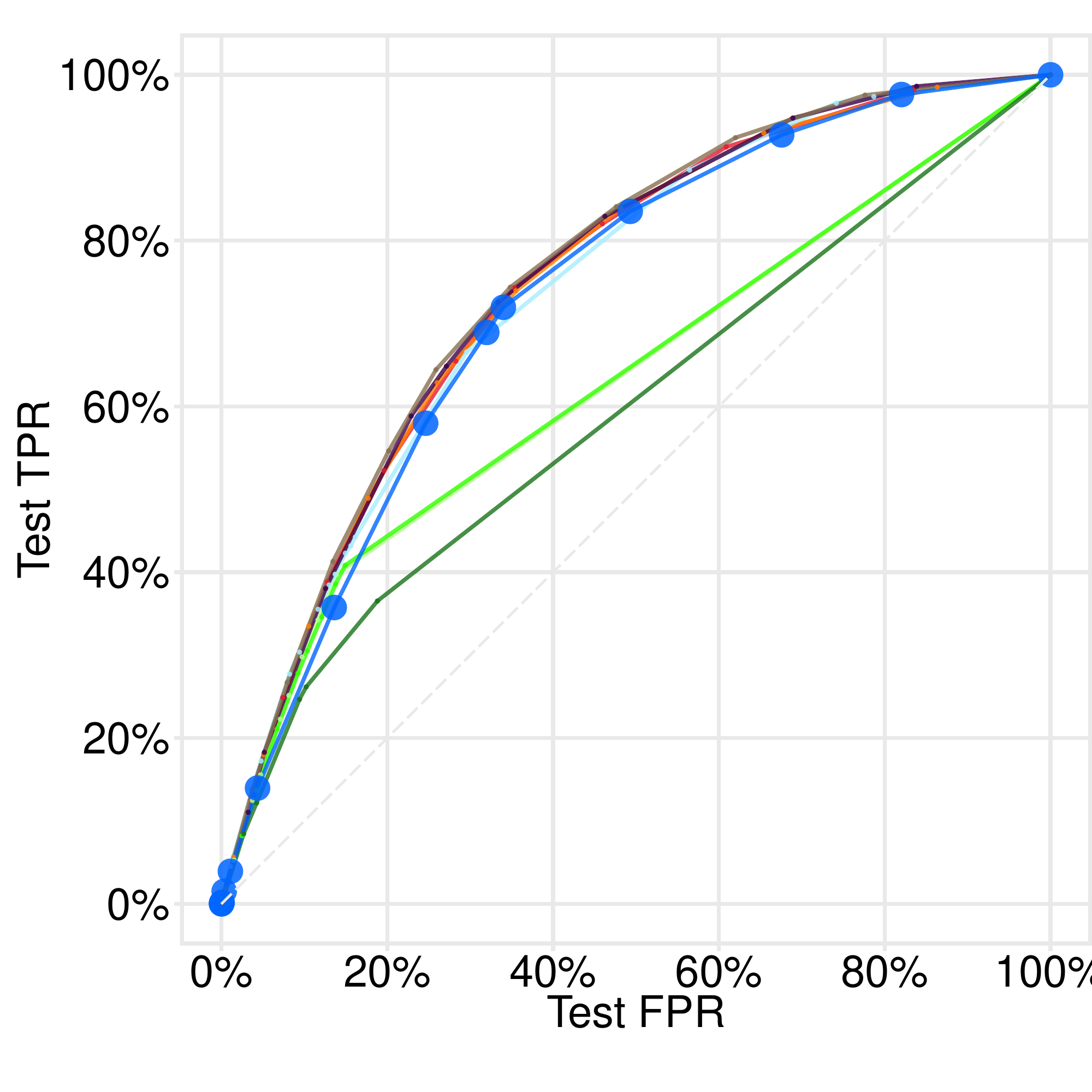} 
\end{tabular}

\vspace{.3cm}

%
%


\tabulartitle{\textwidth}{\textds{general\_violence}}{\textds{domestic\_violence}} \\ 
\begin{tabular}{lr}
	\rocinclude{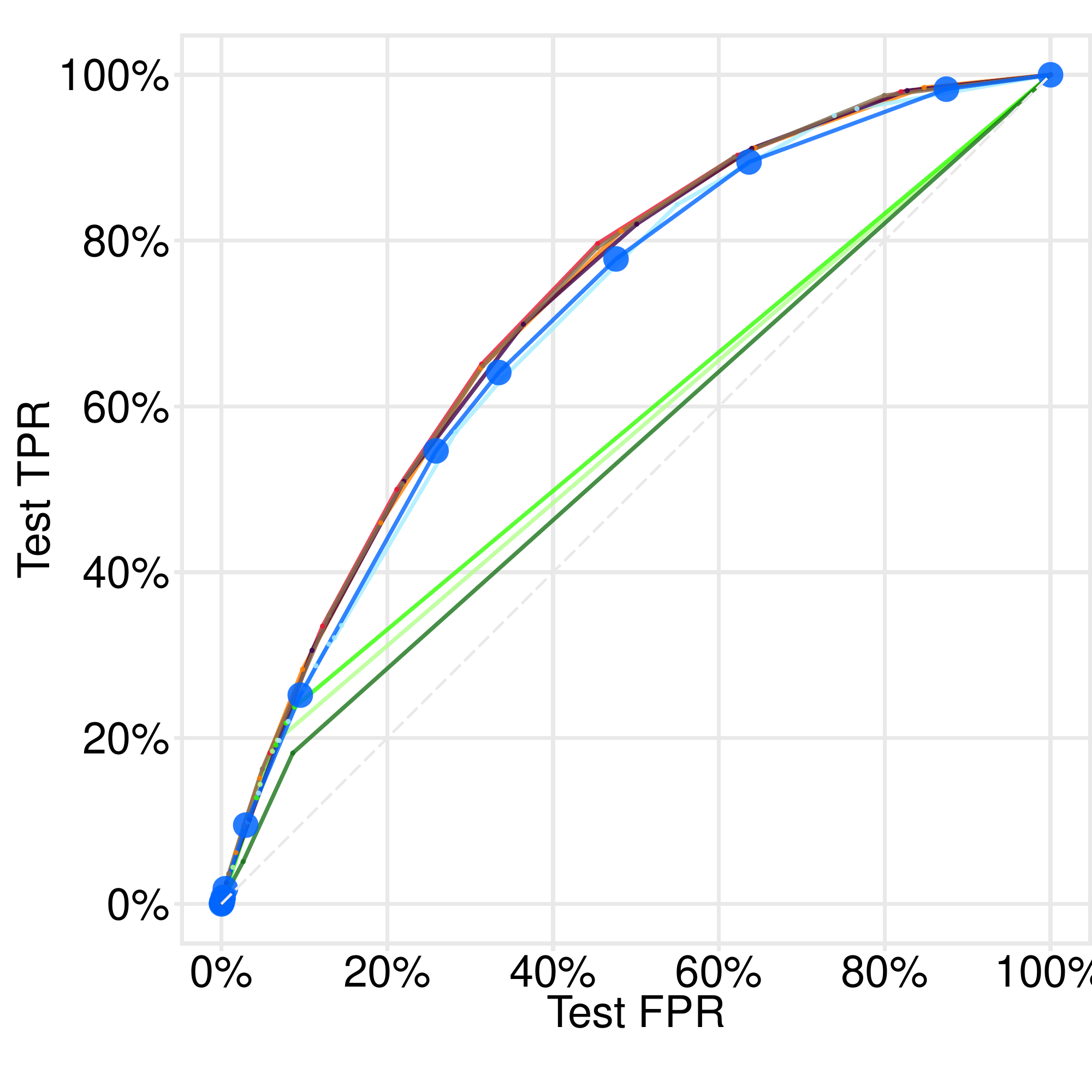} 
	\rocinclude{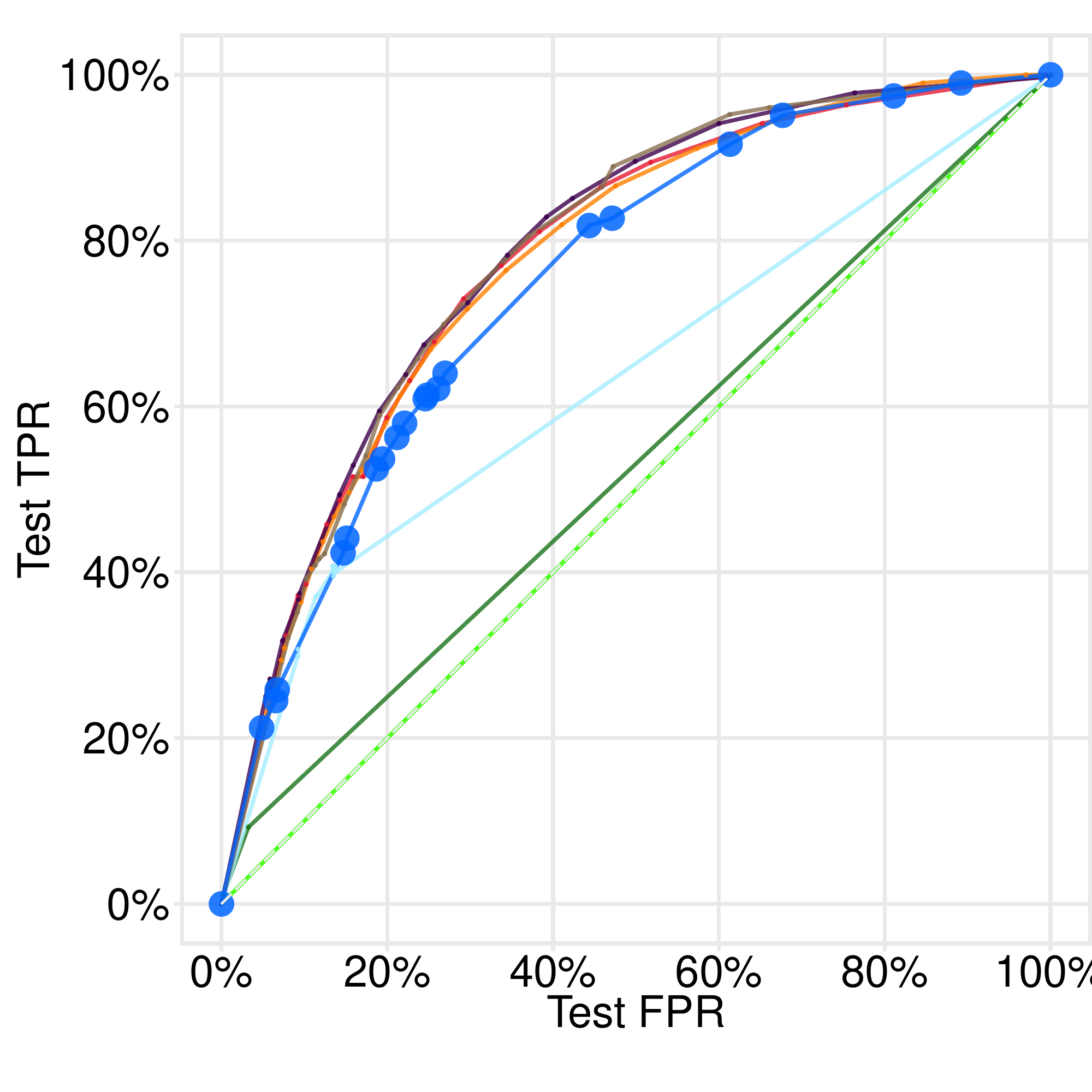} &
\end{tabular}

\vspace{.3cm}

\tabulartitle{\textwidth}{\textds{sexual\_violence}}{\textds{fatal\_violence}} \\ 
\begin{tabular}{lr}
	\rocinclude{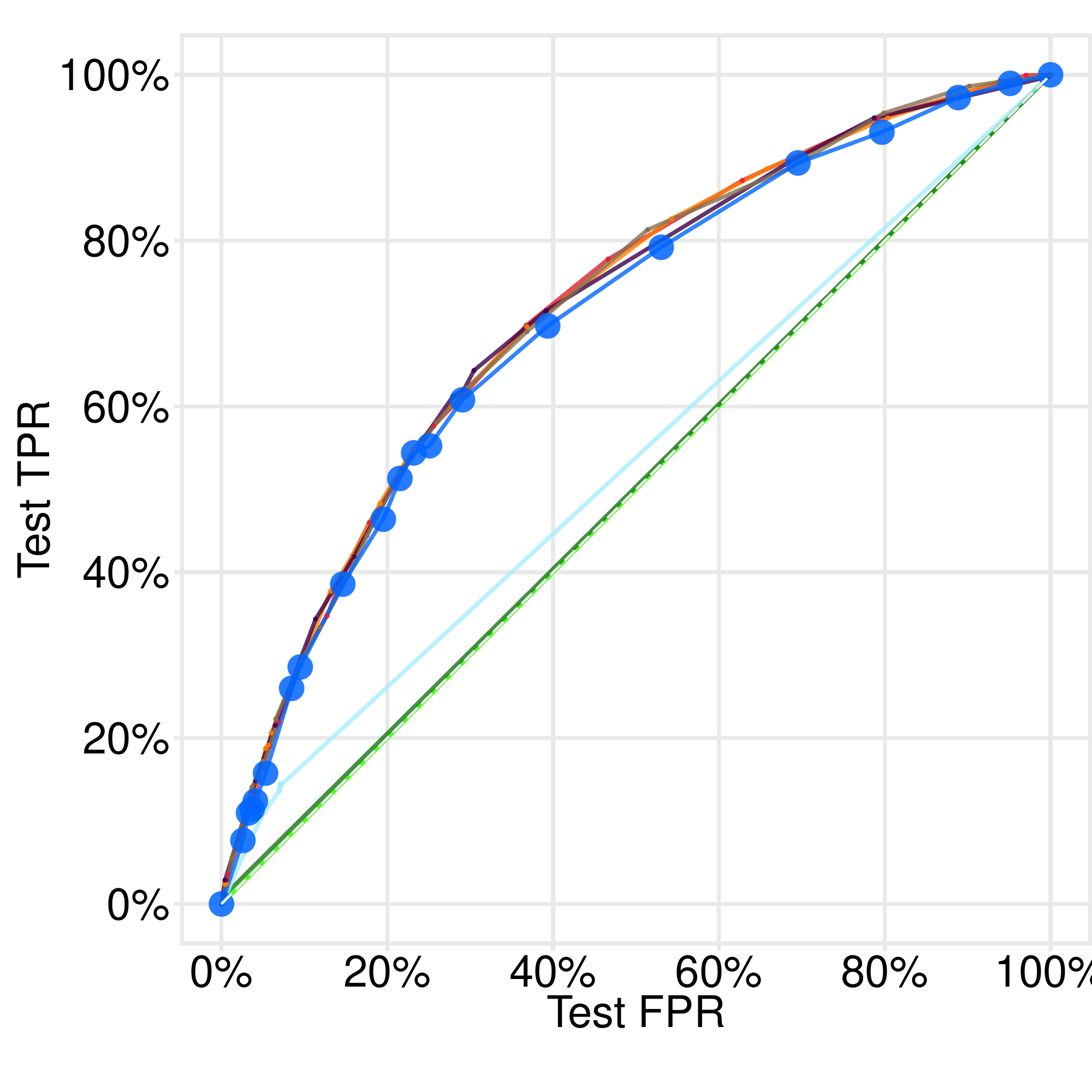} & 
	\rocinclude{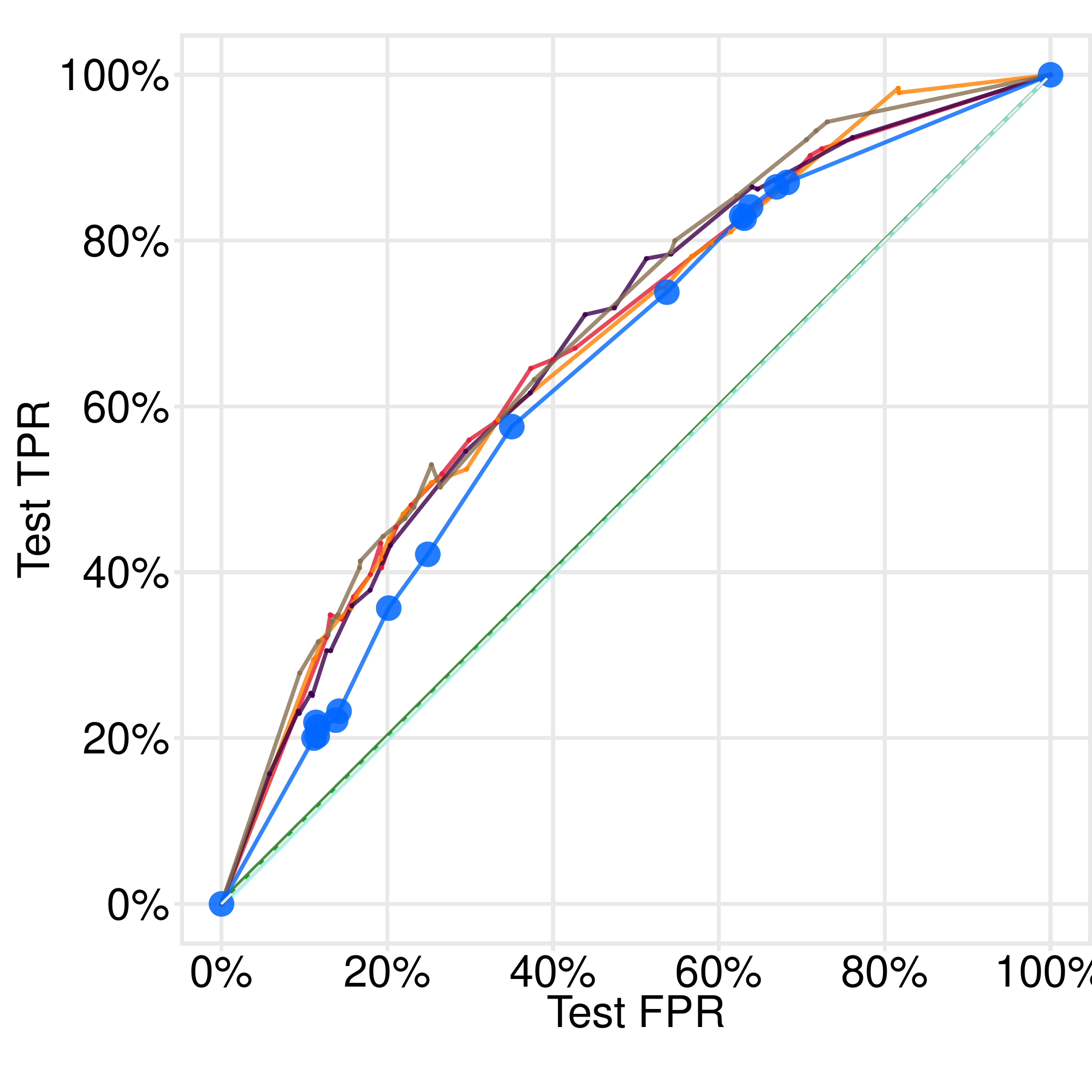}
\end{tabular}

\vspace{.2cm}

\begin{tabular}{c}
\includegraphics[trim=0.5in 0in 0.5in 0.3in,clip,width=0.95\textwidth]{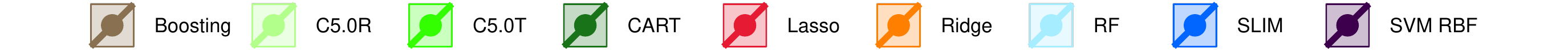}
\end{tabular}
\vspace{-0.1in}
\caption{ROC curves for general recidivism-related prediction problems with test data. We plot SLIM models using large blue dots. All models perform similarly except for C5.0R, C5.0T, and CART.}
\label{Figure::ROCCurves1}
\end{figure}
\begin{figure}
\newcommand{\calibinclude}[1]{\includegraphics[trim=0.2cm 0.4cm 0.0cm 1.1cm,clip,width=\fwidth]{#1}}
\centering
\begin{tabular}{lr}
\calibinclude{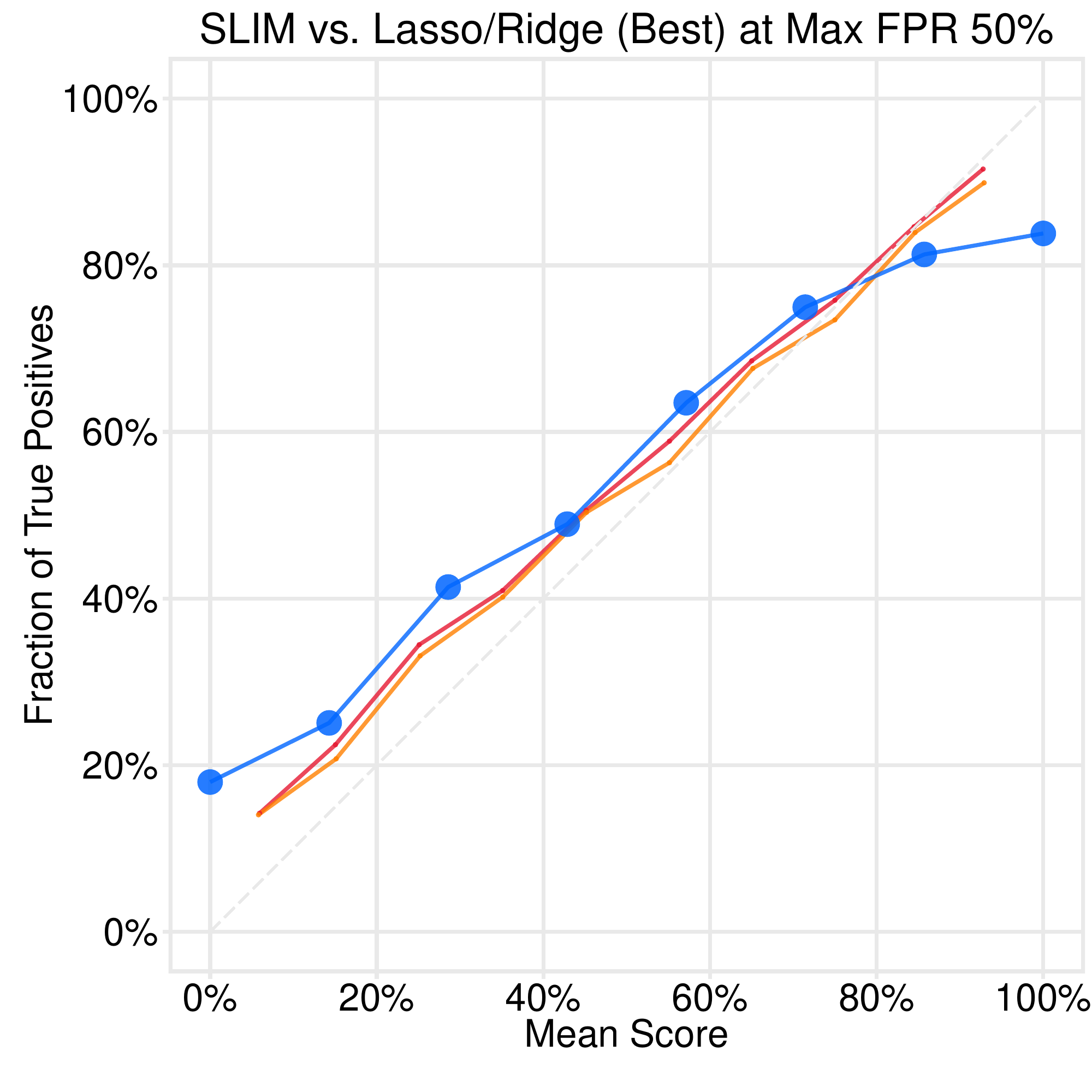}
\end{tabular}
\begin{tabular}{c}
\includegraphics[trim=0.5in 0in 0.5in 0.3in,clip,width=0.9\textwidth]{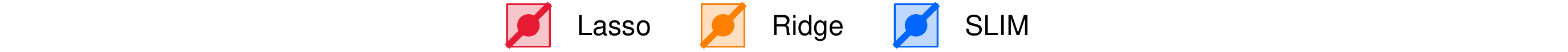}
\end{tabular}
\vspace{-0.1in}
\caption{Risk calibration plot for \textds{arrest} based on test data. We compare 3 models chosen at a similar decision point, with test FPR$\leq50\%$. Although it is not a risk assessment tool, we see that SLIM is well calibrated.}
\label{Figure::rearrestCalibPlot}
\end{figure}

\begin{table}
\caption{\label{Table::testAUC}Test AUC for all methods on all prediction problems. Each cell contains the test AUC. }
\scriptsize
\renewcommand{\arraystretch}{1.2}
\resizebox{\textwidth}{!} {\centering
\begin{tabular}{lccccccccc}
  \toprule
\bfcell{c}{Prediction Problem} & \bfcell{c}{Lasso} & \bfcell{c}{Ridge} & \bfcell{c}{C5.0R} & \bfcell{c}{C5.0T} & \bfcell{c}{CART} & \bfcell{c}{RF} & \bfcell{c}{SVM RBF} & \bfcell{c}{SGB} & \bfcell{c}{SLIM} \\ 
  \toprule
  \texttt{arrest} & \cell{c}{0.72} & \cell{c}{0.73} & \cell{c}{0.72} & \cell{c}{0.72} & \cell{c}{0.68} & \cell{c}{0.73} & \cell{c}{0.72} & \cell{c}{0.73} & \cell{c}{0.72} \\ 
   \midrule \texttt{drug} & \cell{c}{0.74} & \cell{c}{0.74} & \cell{c}{0.63} & \cell{c}{0.63} & \cell{c}{0.59} & \cell{c}{0.75} & \cell{c}{0.73} & \cell{c}{0.75} & \cell{c}{0.74} \\ 
   \midrule \texttt{general\_violence} & \cell{c}{0.72} & \cell{c}{0.72} & \cell{c}{0.56} & \cell{c}{0.57} & \cell{c}{0.56} & \cell{c}{0.71} & \cell{c}{0.70} & \cell{c}{0.72} & \cell{c}{0.71} \\ 
   \midrule \texttt{domestic\_violence} & \cell{c}{0.77} & \cell{c}{0.77} & \cell{c}{0.50} & \cell{c}{0.50} & \cell{c}{0.53} & \cell{c}{0.64} & \cell{c}{0.77} & \cell{c}{0.78} & \cell{c}{0.76} \\
   \midrule \texttt{sexual\_violence} & \cell{c}{0.72} & \cell{c}{0.72} & \cell{c}{0.50} & \cell{c}{0.50} & \cell{c}{0.51} & \cell{c}{0.54} & \cell{c}{0.69} & \cell{c}{0.70} & \cell{c}{0.70} \\ 
   \midrule \texttt{fatal\_violence} & \cell{c}{0.67} & \cell{c}{0.68} & \cell{c}{0.50} & \cell{c}{0.50} & \cell{c}{0.50} & \cell{c}{0.50} & \cell{c}{0.69} & \cell{c}{0.70} & \cell{c}{0.62} \\
   \bottomrule \end{tabular}
}
\end{table}
%
%
\clearpage
\FloatBarrier
\subsection{Trade-offs Between Accuracy and Interpretability}
\label{Sec::tradeoffsAccuracyInterpretability}
In Appendix \ref{Appendix::InterpretabilityTradeoffs}, we show that the baseline methods are unable to maintain the same level of accuracy as they have in Section \ref{Sec::AccuracyComparison} when their model size was constrained. For Lasso, Ridge and SLIM, model size is defined as the number of features in the model. For CART and C5.0, model size is the number of leaves or rules. In fact, we find the only methods that can consistently produce accurate models along the full ROC curve and also have the potential for interpretability are SLIM and (non-sparse) Lasso. 

%




Tree and rule-based methods such as CART, C5.0T and C5.0R were generally unable to produce models that attain high degrees of accuracy. Worse, even for balanced problems such as \textds{arrest}, where these methods did produce accurate models, the models are complicated and use a very large number of rules or leaves  \citep[similar behavior for C5.0T/C5.0R is also observed by, for instance,][]{lim2000comparison}. As we show in Appendix \ref{Appendix::InterpretabilityTradeoffs}, it was not reasonably possible to obtain a C5.0R/C5.0T/CART model with at most 8 rules or 8 leaves for almost every prediction problem. 



\subsection{On the Interpretability of Equally Accurate Transparent Models}
\label{Sec::InterpretabilityResults}

\begin{figure}
\centering
\tikzstyle{block} = [fill=white!20, text centered, minimum height=1em,font=\it]
\tikzstyle{line} = [draw, color=gray!40, -latex']
\tikzstyle{lnode} = [below=-5.5pt, near end, color=gray, fill=white!20, text centered, font=\tiny]
\tikzstyle{rcloud} = [fill=red!20, text centered,font=\scriptsize]
\tikzstyle{gcloud} = [fill=green!20, text centered,font=\scriptsize]
    
\begin{tikzpicture}
\footnotesize
\node [block] (root) {prior\_arrests$\geq$5};
\node [block,left of=root,left=10em,below of=root] (D1) {age\_at\_release\_18\_to\_24};
\node [gcloud, left of=D1, left=3em, below of=D1] (end1no) {not rearrested}; 
\node [rcloud, right of=D1, right=3em, below of=D1] (end1yes) {rearrested};
\node [block, right of=root, right=10em,below of=root] (D2) {age\_at\_release $\geq$ 40};
\node [rcloud,right of=D2, right=3em, below of=D2] (end2yes) {rearrested};
\node [block, left of=D2, left=4em, below of=D2, below=.3em] (D3) {prior\_arrest\_for\_misdemeanor};
\node [rcloud, right of=D3, right=4em, below of=D3] (end3yes) {rearrested};
\node [gcloud, left of=D3, left=4em, below of=D3] (end3no) {not rearrested};

\path [line] (root) -| node [lnode] {NO} (D1);
\path [line] (D1) -| node [lnode] {YES} (end1yes);
\path [line] (D1) -| node [lnode] {NO} (end1no);
\path [line] (root) -| node [lnode] {YES} (D2);
\path [line] (D2) -| node [lnode] {NO} (end2yes);
\path [line] (D2) -| node [lnode] {YES} (D3);
\path [line] (D3) -| node [lnode] {YES} (end3yes);
\path [line] (D3) -| node [lnode] {NO} (end3no);
   
\end{tikzpicture}
\caption{CART model for \textds{arrest}. This model has a test TPR/FPR of 79.1$\%$/47.9$\%$, and a mean 5-CV validation TPR/FPR of 79.9$\%$/48.5$\%$.}
\label{Model::rearrestCARTModel}
\end{figure}
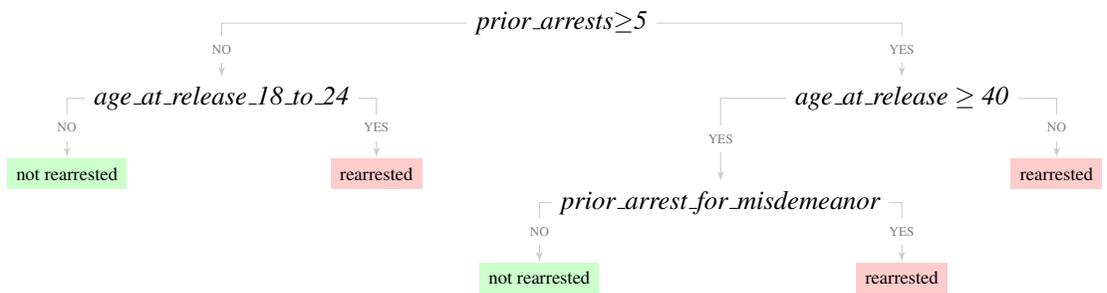

To assess the interpretability of different models, we provide a comparison of predictive models produced by SLIM, Lasso and CART for the \textds{arrest} problem in Figures \ref{Model::rearrestSLIMModel}--\ref{Model::rearrestCARTModel}. This setup provides a nice basis for comparison as all three methods produce models at roughly the same decision point, and with the same degree of sparsity. For this comparison, we considered any transparent model with at most 8 coefficients (Lasso), 8 rules (C5.0R) or 8 leaves (C5.0T, CART) and had a test FPR of below 50\%. We report the models with the minimum weighted test error. Here, neither C5.0R nor C5.0T could produce an acceptable model with at most 8 rules or 8 leaves, so only models from SLIM, CART and Lasso could be displayed.
As described before, it is rare for Lasso and CART to produce models with a similar degree of accuracy to SLIM when model size is constrained. We make the following observations:

\begin{itemize}[leftmargin=*]

\item All three models attain similar levels of predictive accuracy.  Test TPR values ranged between 70-79\% and test FPR values ranged between 43-48\%. There may not exist a classification model that can attain substantially higher accuracy. 

\item The SLIM model uses 5 input variables and small integer coefficients 
(see e.g., Figure \ref{Model::rearrestSLIMModel}). There is a natural rule-based interpretation. In this case, the model implies that if the prisoner is young (\textfn{age\_at\_release\_of\_18\_to\_24}) or has a history of arrests (\textfn{prior\_arrests$\geq$5}), he is highly likely to be rearrested. On the other hand, if he is relatively older (\textfn{age\_at\_release$\geq$40}) or has no history of arrests (\textfn{no\_prior\_arrests}), he is unlikely to commit another crime. 


\item The CART model also allows users to make predictions without a calculator. In comparison to the SLIM model, however, the hierarchical structure of the CART model makes it difficult to gauge the relationship of each input variable on the predicted outcome. Consider, for instance, the relationship between age at release and the outcome. In this case, users are immediately aware that there is an effect, as the model branches on the variables \textfn{age\_at\_release$\geq$40} and \textfn{age\_at\_release\_18\_to\_24}. However, the effect is difficult to comprehend since it depends on prior arrests for misdemeanor: if \textfn{prior\_arrests$\geq$5} = 1 and \textfn{age\_at\_release\_18\_to\_24} = 1 then the model predicts $\hat{y}=+1$; if \textfn{prior\_arrests$\geq$5} = 0 and \textfn{age\_at\_release$\geq$40} = 0 then $\hat{y}=+1$; however, if \textfn{prior\_arrests$\geq$5} = 0 and \textfn{age\_at\_release$\geq$40} = 1 then $\hat{y}=+1$ only if \textfn{prior\_arrest\_for\_misdemeanor} = 1. Such issues do not affect linear models such as SLIM and Lasso, where users can immediately gauge the direction and strength of the relationship between a input variable and the predicted outcome by the size and sign of a coefficient. The literature on interpretability in machine learning indicates that interpretability is domain-specific; there are some domains where logical models are preferred over linear models, and vice versa \citep[e.g.,][]{Freitas:2014ic}.

\end{itemize}

\begin{figure}[]
\footnotesize
\centering{
\predcell{c}{PREDICT ARREST FOR ANY OFFENSE IF SCORE $> 1$}
\begin{tabular}{|l l  c | c |}
   \hline
1. & $\textfn{age\_at\_release\_18\_to\_24}$ & 2 points & $\phantom{+}\quad\cdots\cdots$ \\ 
  2. & $\textfn{prior\_arrests$\geq$5}$ & 2 points & $+\quad\cdots\cdots$ \\ 
  3. & $\textfn{prior\_arrest\_for\_misdemeanor}$ & 1 point & $+\quad\cdots\cdots$ \\ 
  4. & $\textfn{no\_prior\_arrests}$ & -1 point & $+\quad\cdots\cdots$ \\ 
  5. & $\textfn{age\_at\_release$\geq$40}$ & -1 point & $+\quad\cdots\cdots$ \\ 
   \hline
 & \small{\textbf{ADD POINTS FROM ROWS 1--5}} & \small{\textbf{SCORE}} & $=\quad\cdots\cdots$ \\ 
   \hline
\end{tabular}
}
\caption{SLIM scoring system for \textds{arrest}. This model has a test TPR/FPR of 76.6$\%$/44.5$\%$, and a mean 5-CV validation TPR/FPR of 78.3$\%$/46.5$\%$.}
\label{Model::rearrestSLIMModel}
\end{figure}
%
\begin{figure}[]
\footnotesize
\centering
\predcell{c}{PREDICT ARREST FOR ANY OFFENSE IF SCORE $> 0.31$}
\begin{tabular}{|l l  c | c |}
   \hline
1. &  $\textfn{prior\_arrests$\geq$5}$ & 0.63 points & $\phantom{+}\quad\cdots\cdots$ \\
2. & $\textfn{age\_1st\_confinement\_18\_to\_24}$ & 0.15 points & $+\quad\cdots\cdots$ \\ 
3. & $\textfn{prior\_arrest\_for\_property}$ & 0.09 points & $+\quad\cdots\cdots$ \\
4. & $\textfn{prior\_arrest\_for\_misdemeanor}$ & 0.05 points & $+\quad\cdots\cdots$ \\
5. & $\textfn{age\_at\_release$\geq$40}$ & -0.20 points & $+\quad\cdots\cdots$\\
  \hline
 & \small{\textbf{ADD POINTS FROM ROWS 1--5}} & \small{\textbf{SCORE}} & $=\quad\cdots\cdots$ \\ 
   \hline
\end{tabular}
\centering\footnotesize
\caption{Lasso model for \textds{arrest}, with coefficients rounded to two significant digits. This model has a test TPR/FPR of 70.9$\%$/43.8$\%$, and a mean 5-CV validation TPR/FPR of 72.2$\%$/44.0$\%$.}
\label{Model::rearrestLassoModel}
\end{figure}
%

\FloatBarrier
\subsection{Scoring Systems for Recidivism Prediction}
\label{Sec::ScoringSystems}
We show a SLIM scoring system for each of the prediction problems that we consider in Figures \ref{Model::drugMaxFPR4000SLIMModel1ScoringSystem}--\ref{Model::fatalviolenceMaxFPR6500SLIMModel1ScoringSystem}. The models are chosen at specific decision points, with the constraint that 5-CV FPR$\leq50\%$ except for \textds{sexual\_violence}, which is chosen at 5-CV FPR$\leq20\%$. The models presented here may be suitable for screening tasks. To obtain a model suitable for sentencing, a point on the ROC curve with a much higher TPR would be needed. We note that these models generalize well from the dataset, evident by the close match between test TPR/FPR (Table \ref{Table::testAUC}) and training TPR/FPR (Table \ref{Table::TrainingAUC}).

Many of these models exhibit the same ``rule-like" tendencies discussed in Section \ref{Sec::InterpretabilityResults}. 
For example,  the model for \textds{drug} in Figure \ref{Model::drugMaxFPR4000SLIMModel1ScoringSystem} predicts that a person will be arrested for a drug-related offense if he/she has ever had any prior drug offenses. Similarly, model for \textds{sexual\_violence} in Figure \ref{Model::sexualMaxFPR5000SLIMModel1ScoringSystem} effectively states that a person will be rearrested for a sexual offense if and only if he/she has prior history of sexual crimes. 
For completeness, we include comparisons with other models in Appendix \ref{Appendix::ModelBasedComparisons}. Additional risk calibration plots for models with constrained model size are included in Appendix \ref{Appendix::LossInCalibration}. 

%
\begin{figure}[]
\footnotesize\centering 
\renewcommand{\arraystretch}{1.2}
\predcell{c}{PREDICT ARREST FOR DRUG OFFENSE IF SCORE $> 7$}

\begin{tabular}{|l l  c | c |}
   \hline
1. & $\textfn{prior\_arrest\_for\_drugs}$ & 9 points & $\phantom{+}\quad\cdots\cdots$ \\ 
  2. & $\textfn{age\_at\_release\_18\_to\_24}$ & 5 points & $+\quad\cdots\cdots$ \\ 
  3. & $\textfn{age\_at\_release\_25\_to\_29}$ & 3 points & $+\quad\cdots\cdots$ \\ 
  4. & $\textfn{prior\_arrest\_for\_multiple\_types\_of\_crime}$ & 2 points & $+\quad\cdots\cdots$ \\ 
  5. & $\textfn{prior\_arrest\_for\_property}$ & 1 points & $+\quad\cdots\cdots$ \\  
  6. & $\textfn{age\_at\_release\_30\_to\_39}$ & -1 point & $+\quad\cdots\cdots$ \\
  7. & $\textfn{no\_prior\_arrests}$ & -6 points & $+\quad\cdots\cdots$ \\ 
   \hline
 & \small{\textbf{ADD POINTS FROM ROWS 1-7}} & \small{\textbf{SCORE}} & $=\quad\cdots\cdots$ \\ 
   \hline
\end{tabular}
\caption{SLIM scoring system for \textds{drug}. This model has a test TPR/FPR of 85.7$\%$/51.1$\%$, and a mean 5-CV validation TPR/FPR of 82.3$\%$/49.7$\%$.}
\label{Model::drugMaxFPR4000SLIMModel1ScoringSystem}
\end{figure}

\begin{figure}[]
\footnotesize
\centering
\renewcommand{\arraystretch}{1.2}
\predcell{c}{PREDICT ARREST FOR GENERAL VIOLENCE OFFENSE IF SCORE $> 7$}
\begin{tabular}{|l l  c | c |}
   \hline
1. & $\textfn{prior\_arrest\_for\_general\_violence}$ & 8 points & $\phantom{+}\quad\cdots\cdots$ \\ 
  2. & $\textfn{prior\_arrest\_for\_misdemeanor}$ & 5 points & $+\quad\cdots\cdots$ \\ 
  3. & $\textfn{infraction\_in\_prison}$ & 3 points & $+\quad\cdots\cdots$ \\ 
  4. & $\textfn{prior\_arrest\_for\_local\_ord}$ & 3 points & $+\quad\cdots\cdots$ \\ 
  5. & $\textfn{prior\_arrest\_for\_property}$ & 2 points & $+\quad\cdots\cdots$ \\ 
  6. & $\textfn{prior\_arrest\_for\_fatal\_violence}$ & 2 points & $+\quad\cdots\cdots$ \\ 
  7. & $\textfn{prior\_arrest\_with\_firearms\_involved}$ & 1 point & $+\quad\cdots\cdots$ \\ 
  8. & $\textfn{age\_at\_release$\geq$40}$ & -7 points & $+\quad\cdots\cdots$ \\ 
   \hline
 & \small{\textbf{ADD POINTS FROM ROWS 1-8}} & \small{\textbf{SCORE}} & $=\quad\cdots\cdots$ \\ 
   \hline
\end{tabular}
\caption{SLIM scoring system for \textds{general\_violence}. This model has a test TPR/FPR of 76.7$\%$/45.4$\%$, and a mean 5-CV validation TPR/FPR of 76.8$\%$/47.6$\%$.}
\label{Model::otherviolenceMaxFPR5000SLIMModel1ScoringSystem}
\end{figure}
\begin{figure}
\footnotesize
\centering{
{\renewcommand{\arraystretch}{1.2}
\predcell{c}{PREDICT ARREST FOR DOMESTIC VIOLENCE OFFENSE IF SCORE $> 3$}

\begin{tabular}{|l l  c | c |}
   \hline
1. & $\textfn{prior\_arrest\_for\_misdemeanor}$ & 4 points & $\phantom{+}\quad\cdots\cdots$ \\ 
  2. & $\textfn{prior\_arrest\_for\_felony}$ & 3 points & $+\quad\cdots\cdots$ \\ 
  3. & $\textfn{prior\_arrest\_for\_domestic\_violence}$ & 2 points & $+\quad\cdots\cdots$ \\ 
  4. & $\textfn{age\_1st\_confinement\_18\_to\_24}$ & 1 point & $+\quad\cdots\cdots$ \\ 
  5. & $\textfn{infraction\_in\_prison}$ & -5 points & $+\quad\cdots\cdots$ \\ 
   \hline
 & \small{\textbf{ADD POINTS FROM ROWS 1-5}} & \small{\textbf{SCORE}} & $=\quad\cdots\cdots$ \\ 
   \hline
\end{tabular}
}
}\caption{SLIM scoring system for \textds{domestic\_violence}. This model has a test TPR/FPR of 85.5$\%$/46.0$\%$, and a mean 5-CV validation TPR/FPR of 81.4$\%$/48.0$\%$.}
\label{Model::domesticviolenceMaxFPR6000SLIMModel1ScoringSystem}
\end{figure}
\begin{figure}
\footnotesize
\centering{
{\renewcommand{\arraystretch}{1.2}
\predcell{c}{PREDICT ARREST FOR SEXUAL VIOLENCE OFFENSE IF SCORE $> 2$}

\begin{tabular}{|l l  c | c |}
   \hline
1. & $\textfn{prior\_arrest\_for\_sexual}$ & 3 points & $\phantom{+}\quad\cdots\cdots$ \\ 
  2. & $\textfn{prior\_arrests$\geq$5}$ & 1 point & $+\quad\cdots\cdots$ \\ 
  3. & $\textfn{multiple\_prior\_jail\_time}$ & 1 point & $+\quad\cdots\cdots$ \\ 
  4. & $\textfn{prior\_arrest\_for\_multiple\_types\_of\_crime}$ & -1 point & $+\quad\cdots\cdots$ \\ 
  5. & $\textfn{no\_prior\_arrests}$ & -2 points & $+\quad\cdots\cdots$ \\ 
   \hline
 & \small{\textbf{ADD POINTS FROM ROWS 1-5}} & \small{\textbf{SCORE}} & $=\quad\cdots\cdots$ \\ 
   \hline
\end{tabular}
}
}\caption{SLIM scoring system for \textds{sexual\_violence}. This model has a test TPR/FPR of 44.3$\%$/17.7$\%$, and a mean 5-CV validation TPR/FPR of 43.7$\%$/19.9$\%$.}
\label{Model::sexualMaxFPR5000SLIMModel1ScoringSystem}
\end{figure}
\begin{figure}[]
\footnotesize
\centering{
{\renewcommand{\arraystretch}{1.2}
\predcell{c}{PREDICT ARREST FOR FATAL VIOLENCE OFFENSE IF SCORE $> 4$}

\begin{tabular}{|l l  c | c |}
   \hline
1. & $\textfn{age\_1st\_confinement$\leq$17}$ & 5 points & $\phantom{+}\quad\cdots\cdots$ \\ 
  2. & $\textfn{prior\_arrest\_with\_firearms\_involved}$ & 3 points & $+\quad\cdots\cdots$ \\ 
  3. & $\textfn{age\_1st\_confinement\_18\_to\_24}$ & 2 points & $+\quad\cdots\cdots$ \\ 
  4. & $\textfn{prior\_arrest\_for\_felony}$ & 2 points & $+\quad\cdots\cdots$ \\ 
  5. & $\textfn{age\_at\_release\_18\_to\_24}$ & 1 point & $+\quad\cdots\cdots$ \\ 
  6. & $\textfn{prior\_arrest\_for\_drugs}$ & 1 point & $+\quad\cdots\cdots$ \\ 
   \hline
 & \small{\textbf{ADD POINTS FROM ROWS 1-6}} & \small{\textbf{SCORE}} & $=\quad\cdots\cdots$ \\ 
   \hline
\end{tabular}
}
}
\caption{SLIM scoring system for \textds{fatal\_violence}. This model has a test TPR/FPR of 55.4$\%$/35.5$\%$, and a mean 5-CV validation TPR/FPR of 64.2$\%$/42.4$\%$.
\label{Model::fatalviolenceMaxFPR6500SLIMModel1ScoringSystem}}
\end{figure}
%

\FloatBarrier
\section{Discussion}


Our paper merges two perspectives on recidivism modeling: the first is to obtain accurate predictive models using the most powerful machine learning tools available, and the second is to create models that are easy to use and understand. 

We used a set of features that are commonly accessible to police officers and judges, and compared the ability of different machine learning methods to produce models at different decision points across the ROC curve. Our results suggest that it is possible for traditional methods, such as Ridge Regression, to perform just as well as more modern methods, such as Stochastic Gradient Boosting -- a finding that is in line with the work of \citet{tollenaar2013method} and \citet{yang2010efficacy}. Further, we found that even simple models may perform surprisingly well, even when they are fitting from a heavily constrained space -- a finding that is in line with work on the surprising performance of simple models \citep[see e.g.,][]{dawes1979robust,holte1993very,Holte:2006fp}. 

Our study shows that there may be major advantages of using SLIM for recidivism prediction, as it can dependably produce a simple scoring system that is accurate and interpretable on any decision point along the ROC curve. Interpretability is crucial for many of the high-stakes applications where recidivism prediction models are being used. In such applications, it is not enough for the decision-maker to know what input variables are being used to train the model, or how individual input variables are related to the outcome; decision-makers should know how the model combines all the input variables to generate its predictions, and whether this mechanism aligns with their ethical values. SLIM not only shows this mechanism, but also accommodates constraints that are designed to align the prediction model with the ethical values of the decision-maker. 



In comparison to current machine learning methods, the main drawback of running SLIM is increased computation involved in solving an integer programming problem. To this end, we proposed two new techniques to reduce computation involved in training high quality SLIM scoring systems: (i) a polishing procedure that improves the quality of feasible solutions found by an IP solver; and (ii) an IP formulation that makes it easier for an IP solver to provide a certificate of optimality. In our experiments, the time required to train SLIM was ultimately comparable to the time required to train random forests or stochastic gradient boosting. However, it was still significant compared to the time required for other methods such as CART, C5.0 and penalized logistic regression. In theory, the computation required to find an optimal solution to the SLIM integer program is NP-hard, meaning that the runtime increases exponentially with the number of features. In practice, the runtime depends on several factors: such as the number of samples, the number of dimensions, the underlying ease of the classification, and how the data are encoded. Since most criminological problems cannot by nature involve massive datasets (since each observation is a person), and since computer speed of solving MIPs is also increasing exponentially, it is possible that mathematical programming techniques like SLIM are well-suited for criminological problems that are substantially larger and more complex than the one discussed in this work.


\bibliographystyle{plainnat}
\bibliography{recidivism}

\appendix

\clearpage
\section{Additional Results on Predictive Accuracy}
\label{Sec::AdditionalResultsOnPredictivePerformance}
To supplement the experimental results in Section \ref{Sec::AccuracyComparison}, we include the training and 5-CV validation results. Table \ref{Table::TrainingAUC} shows the training AUC performance for all methods on all prediction problems, and Table \ref{Table::ValidationAUC} shows the 5-CV validation AUC performance for all methods. A table of test AUC for all methods on all prediction problems can be found in Table \ref{Table::testAUC}.
\begin{table}
\caption{\label{Table::TrainingAUC}Training AUC for all methods on all prediction problems.}
\scriptsize
\renewcommand{\arraystretch}{1.2}
\resizebox{\textwidth}{!} {\centering
\begin{tabular}{lccccccccc}
  \toprule
\bfcell{c}{Prediction Problem} & \bfcell{c}{Lasso} & \bfcell{c}{Ridge} & \bfcell{c}{C5.0R} & \bfcell{c}{C5.0T} & \bfcell{c}{CART} & \bfcell{c}{RF} & \bfcell{c}{SVM RBF} & \bfcell{c}{SGB} & \bfcell{c}{SLIM} \\ 
  \toprule
  \texttt{arrest} & \cell{c}{0.73} & \cell{c}{0.73} & \cell{c}{0.73} & \cell{c}{0.73} & \cell{c}{0.81} & \cell{c}{0.73} & \cell{c}{0.87} & \cell{c}{0.75} & \cell{c}{0.72} \\
   \midrule \texttt{drug} & \cell{c}{0.74} & \cell{c}{0.73} & \cell{c}{0.65} & \cell{c}{0.66} & \cell{c}{0.76} & \cell{c}{0.73} & \cell{c}{0.85} & \cell{c}{0.77} & \cell{c}{0.73} \\ 
   \midrule \texttt{general\_violence} & \cell{c}{0.71} & \cell{c}{0.71} & \cell{c}{0.58} & \cell{c}{0.59} & \cell{c}{0.77} & \cell{c}{0.71} & \cell{c}{0.84} & \cell{c}{0.74} & \cell{c}{0.71} \\ 
   \midrule \texttt{domestic\_violence} & \cell{c}{0.77} & \cell{c}{0.77} & \cell{c}{0.50} & \cell{c}{0.50} & \cell{c}{0.75} & \cell{c}{0.64} & \cell{c}{0.88} & \cell{c}{0.81} & \cell{c}{0.76} \\
   \midrule \texttt{sexual\_violence} & \cell{c}{0.71} & \cell{c}{0.71} & \cell{c}{0.50} & \cell{c}{0.50} & \cell{c}{0.84} & \cell{c}{0.55} & \cell{c}{0.86} & \cell{c}{0.77} & \cell{c}{0.71} \\ 
   \midrule \texttt{fatal\_violence} & \cell{c}{0.75} & \cell{c}{0.74} & \cell{c}{0.50} & \cell{c}{0.50} & \cell{c}{0.50} & \cell{c}{0.51} & \cell{c}{0.90} & \cell{c}{0.84} & \cell{c}{0.73} \\ 
   \bottomrule \end{tabular}

}
\end{table}
\begin{table}
\caption{\label{Table::ValidationAUC}5-CV validation AUC for all methods on all prediction problems. We report the 5-CV mean validation AUC. The ranges underneath each cell represent the 5-CV minimum and maximum.}
\scriptsize
\renewcommand{\arraystretch}{1.2}
\resizebox{\textwidth}{!} {\centering
\begin{tabular}{lccccccccc}
  \toprule
\bfcell{c}{Prediction Problem} & \bfcell{c}{Lasso} & \bfcell{c}{Ridge} & \bfcell{c}{C5.0R} & \bfcell{c}{C5.0T} & \bfcell{c}{CART} & \bfcell{c}{RF} & \bfcell{c}{SVM RBF} & \bfcell{c}{SGB} & \bfcell{c}{SLIM} \\ 
  \toprule
\texttt{arrest} & \cell{c}{\scriptsize{0.72}\\\tiny{0.72 - 0.74}} & \cell{c}{\scriptsize{0.73}\\\tiny{0.72 - 0.74}} & \cell{c}{\scriptsize{0.71}\\\tiny{0.71 - 0.73}} & \cell{c}{\scriptsize{0.71}\\\tiny{0.70 - 0.72}} & \cell{c}{\scriptsize{0.67}\\\tiny{0.66 - 0.69}} & \cell{c}{\scriptsize{0.73}\\\tiny{0.72 - 0.74}} & \cell{c}{\scriptsize{0.71}\\\tiny{0.70 - 0.72}} & \cell{c}{\scriptsize{0.73}\\\tiny{0.72 - 0.74}} & \cell{c}{\scriptsize{0.72}\\\tiny{0.71 - 0.73}} \\ 
   \midrule \texttt{drug} & \cell{c}{\scriptsize{0.73}\\\tiny{0.72 - 0.74}} & \cell{c}{\scriptsize{0.73}\\\tiny{0.71 - 0.74}} & \cell{c}{\scriptsize{0.62}\\\tiny{0.61 - 0.64}} & \cell{c}{\scriptsize{0.62}\\\tiny{0.61 - 0.64}} & \cell{c}{\scriptsize{0.59}\\\tiny{0.58 - 0.60}} & \cell{c}{\scriptsize{0.73}\\\tiny{0.72 - 0.74}} & \cell{c}{\scriptsize{0.72}\\\tiny{0.71 - 0.73}} & \cell{c}{\scriptsize{0.74}\\\tiny{0.72 - 0.74}} & \cell{c}{\scriptsize{0.72}\\\tiny{0.71 - 0.73}} \\  
   \midrule \texttt{general\_violence} & \cell{c}{\scriptsize{0.71}\\\tiny{0.70 - 0.71}} & \cell{c}{\scriptsize{0.71}\\\tiny{0.70 - 0.71}} & \cell{c}{\scriptsize{0.56}\\\tiny{0.55 - 0.57}} & \cell{c}{\scriptsize{0.57}\\\tiny{0.55 - 0.59}} & \cell{c}{\scriptsize{0.56}\\\tiny{0.55 - 0.58}} & \cell{c}{\scriptsize{0.70}\\\tiny{0.69 - 0.71}} & \cell{c}{\scriptsize{0.69}\\\tiny{0.69 - 0.70}} & \cell{c}{\scriptsize{0.71}\\\tiny{0.70 - 0.71}} & \cell{c}{\scriptsize{0.70}\\\tiny{0.69 - 0.71}} \\ 
   \midrule \texttt{domestic\_violence} & \cell{c}{\scriptsize{0.76}\\\tiny{0.75 - 0.79}} & \cell{c}{\scriptsize{0.76}\\\tiny{0.75 - 0.78}} & \cell{c}{\scriptsize{0.50}\\\tiny{0.50 - 0.50}} & \cell{c}{\scriptsize{0.50}\\\tiny{0.50 - 0.50}} & \cell{c}{\scriptsize{0.53}\\\tiny{0.51 - 0.54}} & \cell{c}{\scriptsize{0.63}\\\tiny{0.59 - 0.66}} & \cell{c}{\scriptsize{0.76}\\\tiny{0.74 - 0.78}} & \cell{c}{\scriptsize{0.77}\\\tiny{0.75 - 0.79}} & \cell{c}{\scriptsize{0.75}\\\tiny{0.72 - 0.78}} \\
   \midrule \texttt{sexual\_violence} & \cell{c}{\scriptsize{0.70}\\\tiny{0.68 - 0.74}} & \cell{c}{\scriptsize{0.69}\\\tiny{0.66 - 0.74}} & \cell{c}{\scriptsize{0.50}\\\tiny{0.50 - 0.50}} & \cell{c}{\scriptsize{0.50}\\\tiny{0.50 - 0.50}} & \cell{c}{\scriptsize{0.51}\\\tiny{0.50 - 0.51}} & \cell{c}{\scriptsize{0.54}\\\tiny{0.53 - 0.55}} & \cell{c}{\scriptsize{0.67}\\\tiny{0.63 - 0.70}} & \cell{c}{\scriptsize{0.68}\\\tiny{0.65 - 0.72}} & \cell{c}{\scriptsize{0.68}\\\tiny{0.66 - 0.72}} \\  
   \midrule \texttt{fatal\_violence} & \cell{c}{\scriptsize{0.66}\\\tiny{0.59 - 0.74}} & \cell{c}{\scriptsize{0.67}\\\tiny{0.62 - 0.75}} & \cell{c}{\scriptsize{0.50}\\\tiny{0.50 - 0.50}} & \cell{c}{\scriptsize{0.50}\\\tiny{0.50 - 0.50}} & \cell{c}{\scriptsize{0.50}\\\tiny{0.50 - 0.52}} & \cell{c}{\scriptsize{0.51}\\\tiny{0.50 - 0.53}} & \cell{c}{\scriptsize{0.67}\\\tiny{0.63 - 0.73}} & \cell{c}{\scriptsize{0.67}\\\tiny{0.61 - 0.74}} & \cell{c}{\scriptsize{0.65}\\\tiny{0.61 - 0.69}} \\
   \bottomrule \end{tabular}

}
\end{table}


\clearpage
\section{Model-Based Comparisons}
\label{Appendix::ModelBasedComparisons}
\FloatBarrier

In Section \ref{Sec::Results}, we included a comparison of transparent models produced for the \textds{arrest} problem. Here, we include a similar comparison for all other recidivism prediction problems. 

The models and calibration plots shown here correspond to the \textit{best} models we produced using Lasso and Ridge (i.e., the ones that were plotted as points in Figure \ref{Figure::ROCCurves1}). We omit CART and C5.0 models are shown because all models that were produced were either trivial or contained too many leaves to be printed. For any given problem, the models operate at similar decision points (TPR), and are constrained to the same FPR criteria as in Section \ref{Sec::ScoringSystems}.

Note that the calibration plots will appear to be flat for problems with significant class imbalance. Typically, a well-calibrated classifier on a problem without class imbalance should fall on the $x=y$ line. However, because the $y$-axis is defined as $P(y= +1|s(x)=s)$, where $s$ is predicted score of a model, the slope of the graph will be less than $P(y= +1)$ by definition. Therefore, for a highly imbalanced problem such as \textds{fatal\_violence}, where $P(y=+1)=0.7\%$, the plot will be flat. 

\subsection{\textds{drug}}

This is the SLIM model for \textds{drug}. This model has a test TPR/FPR of 85.7$\%$/51.1$\%$, and a mean 5-CV validation TPR/FPR of 82.3$\%$/49.7$\%$.
\begin{center}
\scriptsize
\begin{tabularx}{\textwidth}{p{1mm}lp{1mm}lp{1mm}l}
   & $9.00 ~\textfn{prior\_arrest\_for\_drugs}$ & $\scriptsize{+}$ & $5.00 ~\textfn{age\_at\_release\_18\_to\_24}$ & $\scriptsize{+}$ & $4.00 ~\textfn{age\_at\_release\_25\_to\_29}$ \\ 
  $\scriptsize{+}$ & $3.00 ~\textfn{prior\_arrest\_for\_multiple\_types\_of\_crime}$ & $\scriptsize{+}$ & $1.00 ~\textfn{prior\_arrest\_for\_property}$ & $\scriptsize{-}$ & $6.00 ~\textfn{no\_prior\_arrests}$ \\ 
  $\scriptsize{-}$ & $1.00 ~\textfn{age\_at\_release\_30\_to\_39}$ & $\scriptsize{-}$ & $7.00$ &  &  \\ 
  \end{tabularx}

\end{center}
%
%
This is the best Lasso model for \textds{drug}. This model has a test TPR/FPR of 82.0$\%$/45.9$\%$, and a mean 5-CV validation TPR/FPR of 81.2$\%$/45.9$\%$.
\begin{center}
\scriptsize
\begin{tabularx}{\textwidth}{p{1mm}lp{1mm}lp{1mm}l}
   & $1.14 ~\textfn{prior\_arrest\_for\_drugs}$ & $\scriptsize{+}$ & $0.27 ~\textfn{prior\_arrest\_for\_property}$ & $\scriptsize{+}$ & $0.26 ~\textfn{time\_served$\leq$6mo}$ \\ 
  $\scriptsize{+}$ & $0.19 ~\textfn{prior\_arrest\_for\_other\_violence}$ & $\scriptsize{+}$ & $0.18 ~\textfn{prior\_arrest\_for\_multiple\_types\_of\_crime}$ & $\scriptsize{+}$ & $0.17 ~\textfn{prior\_arrest\_for\_misdemeanor}$ \\ 
  $\scriptsize{+}$ & $0.16 ~\textfn{age\_at\_release\_18\_to\_24}$ & $\scriptsize{+}$ & $0.14 ~\textfn{prior\_arrests$\geq$5}$ & $\scriptsize{+}$ & $0.13 ~\textfn{age\_1st\_confinement\_18\_to\_24}$ \\ 
  $\scriptsize{+}$ & $0.12 ~\textfn{prior\_arrest\_for\_public\_order}$ & $\scriptsize{+}$ & $0.10 ~\textfn{prior\_arrest\_with\_firearms\_involved}$ & $\scriptsize{+}$ & $0.08 ~\textfn{any\_prior\_jail\_time}$ \\ 
  $\scriptsize{+}$ & $0.06 ~\textfn{age\_1st\_arrest$\leq$17}$ & $\scriptsize{+}$ & $0.04 ~\textfn{multiple\_prior\_jail\_time}$ & $\scriptsize{+}$ & $0.04 ~\textfn{drug\_abuse}$ \\ 
  $\scriptsize{+}$ & $0.03 ~\textfn{multiple\_prior\_prison\_time}$ & $\scriptsize{+}$ & $0.03 ~\textfn{any\_prior\_prb\_or\_fine}$ & $\scriptsize{-}$ & $0.62 ~\textfn{age\_at\_release$\geq$40}$ \\ 
  $\scriptsize{-}$ & $0.25 ~\textfn{prior\_arrest\_for\_sexual}$ & $\scriptsize{-}$ & $0.23 ~\textfn{age\_at\_release\_30\_to\_39}$ & $\scriptsize{-}$ & $0.12 ~\textfn{time\_served\_25\_to\_60mo}$ \\ 
  $\scriptsize{-}$ & $0.11 ~\textfn{prior\_arrest\_with\_child\_involved}$ & $\scriptsize{-}$ & $0.08 ~\textfn{alcohol\_abuse}$ & $\scriptsize{-}$ & $0.07 ~\textfn{age\_1st\_confinement$\geq$40}$ \\ 
  $\scriptsize{-}$ & $1.11 \times 10^{-03} ~\textfn{time\_served$\geq$61mo}$ & $\scriptsize{-}$ & $1.01$ &  &  \\ 
  \end{tabularx}
\end{center}
This is the best Ridge model for \textds{drug}. This model has a test TPR/FPR of 84.0$\%$/48.2$\%$, and a mean 5-CV validation TPR/FPR of 83.1$\%$/48.4$\%$.
\begin{center}
\scriptsize
\begin{tabularx}{\textwidth}{p{1mm}lp{1mm}lp{1mm}l}
   & $0.91 ~\textfn{prior\_arrest\_for\_drugs}$ & $\scriptsize{+}$ & $0.25 ~\textfn{time\_served$\leq$6mo}$ & $\scriptsize{+}$ & $0.24 ~\textfn{age\_at\_release\_18\_to\_24}$ \\ 
  $\scriptsize{+}$ & $0.21 ~\textfn{prior\_arrest\_for\_multiple\_types\_of\_crime}$ & $\scriptsize{+}$ & $0.20 ~\textfn{prior\_arrest\_for\_property}$ & $\scriptsize{+}$ & $0.17 ~\textfn{prior\_arrest\_for\_misdemeanor}$ \\ 
  $\scriptsize{+}$ & $0.17 ~\textfn{prior\_arrest\_for\_other\_violence}$ & $\scriptsize{+}$ & $0.17 ~\textfn{age\_1st\_confinement\_18\_to\_24}$ & $\scriptsize{+}$ & $0.14 ~\textfn{prior\_arrests$\geq$5}$ \\ 
  $\scriptsize{+}$ & $0.13 ~\textfn{prior\_arrest\_with\_firearms\_involved}$ & $\scriptsize{+}$ & $0.12 ~\textfn{age\_at\_release\_25\_to\_29}$ & $\scriptsize{+}$ & $0.11 ~\textfn{drug\_abuse}$ \\ 
  $\scriptsize{+}$ & $0.11 ~\textfn{prior\_arrest\_for\_public\_order}$ & $\scriptsize{+}$ & $0.09 ~\textfn{age\_1st\_arrest$\leq$17}$ & $\scriptsize{+}$ & $0.08 ~\textfn{age\_1st\_confinement$\leq$17}$ \\ 
  $\scriptsize{+}$ & $0.08 ~\textfn{any\_prior\_jail\_time}$ & $\scriptsize{+}$ & $0.07 ~\textfn{multiple\_prior\_jail\_time}$ & $\scriptsize{+}$ & $0.07 ~\textfn{age\_at\_release$\leq$17}$ \\ 
  $\scriptsize{+}$ & $0.06 ~\textfn{multiple\_prior\_prison\_time}$ & $\scriptsize{+}$ & $0.06 ~\textfn{released\_unconditonal}$ & $\scriptsize{+}$ & $0.05 ~\textfn{any\_prior\_prb\_or\_fine}$ \\ 
  $\scriptsize{+}$ & $0.05 ~\textfn{prior\_arrests$\geq$2}$ & $\scriptsize{+}$ & $0.04 ~\textfn{time\_served\_7\_to\_12mo}$ & $\scriptsize{+}$ & $0.04 ~\textfn{multiple\_prior\_prb\_or\_fine}$ \\ 
  $\scriptsize{+}$ & $0.02 ~\textfn{prior\_arrests$\geq$1}$ & $\scriptsize{+}$ & $0.01 ~\textfn{age\_1st\_confinement\_25\_to\_29}$ & $\scriptsize{+}$ & $0.01 ~\textfn{released\_conditonal}$ \\ 
  $\scriptsize{+}$ & $2.52 \times 10^{-03} ~\textfn{prior\_arrest\_for\_felony}$ & $\scriptsize{+}$ & $1.76 \times 10^{-03} ~\textfn{age\_1st\_arrest\_18\_to\_24}$ & $\scriptsize{+}$ & $9.58 \times 10^{-04} ~\textfn{prior\_arrest\_for\_fatal\_violence}$ \\ 
  $\scriptsize{-}$ & $0.33 ~\textfn{age\_at\_release$\geq$40}$ & $\scriptsize{-}$ & $0.25 ~\textfn{prior\_arrest\_for\_sexual}$ & $\scriptsize{-}$ & $0.19 ~\textfn{age\_1st\_confinement$\geq$40}$ \\ 
  $\scriptsize{-}$ & $0.16 ~\textfn{prior\_arrest\_with\_child\_involved}$ & $\scriptsize{-}$ & $0.15 ~\textfn{time\_served\_25\_to\_60mo}$ & $\scriptsize{-}$ & $0.14 ~\textfn{alcohol\_abuse}$ \\ 
  $\scriptsize{-}$ & $0.13 ~\textfn{time\_served$\geq$61mo}$ & $\scriptsize{-}$ & $0.10 ~\textfn{prior\_arrest\_for\_domestic\_violence}$ & $\scriptsize{-}$ & $0.09 ~\textfn{age\_at\_release\_30\_to\_39}$ \\ 
  $\scriptsize{-}$ & $0.05 ~\textfn{age\_1st\_arrest$\geq$40}$ & $\scriptsize{-}$ & $0.04 ~\textfn{female}$ & $\scriptsize{-}$ & $0.04 ~\textfn{infraction\_in\_prison}$ \\ 
  $\scriptsize{-}$ & $0.03 ~\textfn{age\_1st\_arrest\_30\_to\_39}$ & $\scriptsize{-}$ & $0.02 ~\textfn{age\_1st\_confinement\_30\_to\_39}$ & $\scriptsize{-}$ & $0.02 ~\textfn{no\_prior\_arrests}$ \\ 
  $\scriptsize{-}$ & $4.71 \times 10^{-03} ~\textfn{prior\_arrest\_for\_local\_ord}$ & $\scriptsize{-}$ & $4.45 \times 10^{-03} ~\textfn{time\_served\_13\_to\_24mo}$ & $\scriptsize{-}$ & $2.23 \times 10^{-03} ~\textfn{age\_1st\_arrest\_25\_to\_29}$ \\ 
  $\scriptsize{-}$ & $1.09$ &  &  &  &  \\ 
  \end{tabularx}
\end{center}
\begin{figure}
	\newcommand{\calibinclude}[1]{\includegraphics[trim=0.2cm 0.4cm 0.0cm 1.6cm,clip,width=.9\fwidth]{#1}}
	\centering
	\calibinclude{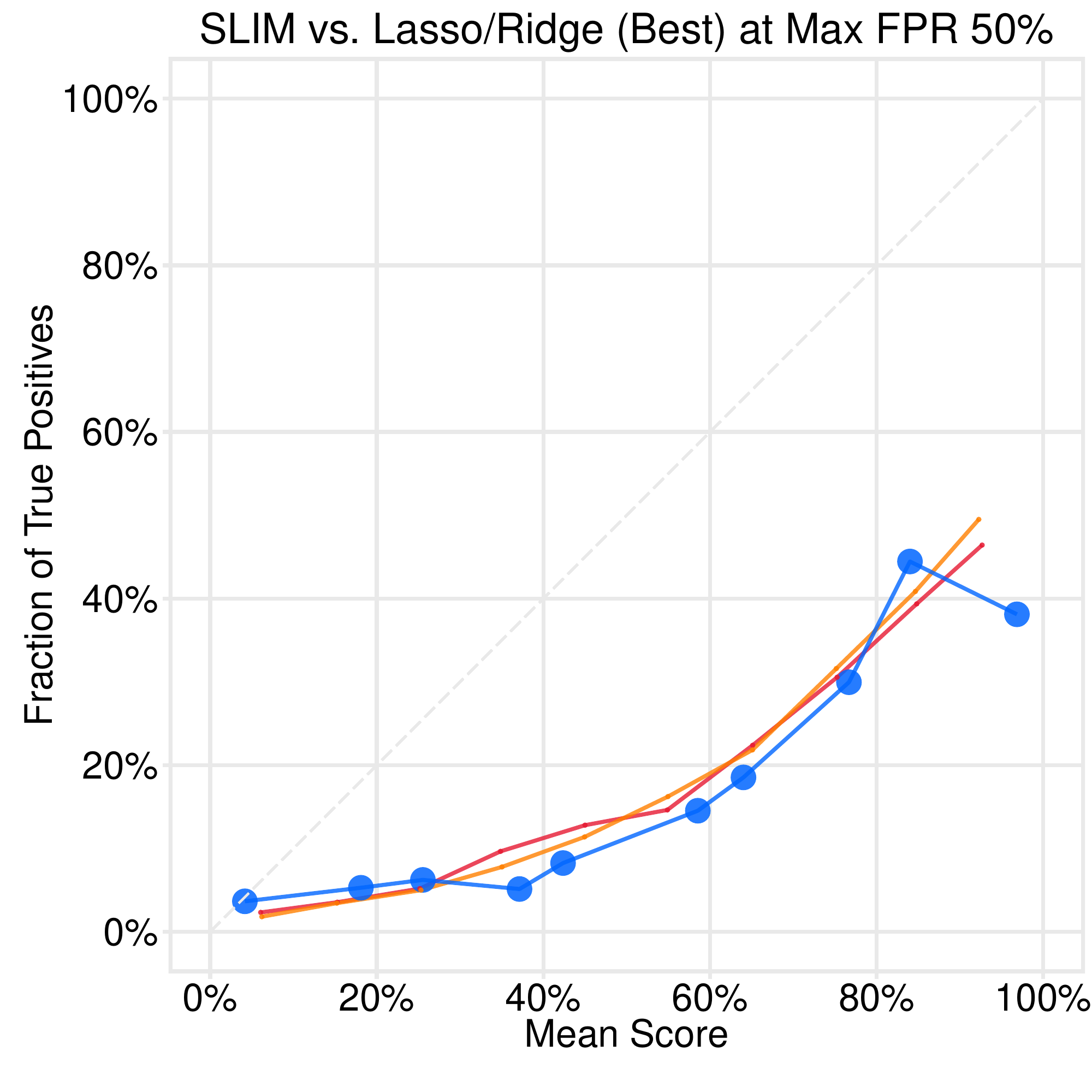}
	\includegraphics[trim=0.5in 0in 0.5in 0.3in,clip,width=0.9\textwidth]{figure/roc_plot_legend_calib}
	\vspace{-0.1in}
	\caption{Risk calibration plot for \textds{drug}.}
	\label{Figure::drugCalibPlot}
\end{figure}

\clearpage
\FloatBarrier
\subsection{\textds{general\_violence}}
SLIM model for \textds{general\_violence}. This model has a test TPR/FPR of 76.7$\%$/45.4$\%$, and a mean 5-CV validation TPR/FPR of 76.8$\%$/47.6$\%$.
\begin{center}
\scriptsize
\begin{tabularx}{\textwidth}{p{1mm}lp{1mm}lp{1mm}l}
   & $8 ~\textfn{prior\_arrest\_for\_other\_violence}$ & $\scriptsize{+}$ & $5 ~\textfn{prior\_arrest\_for\_misdemeanor}$ & $\scriptsize{+}$ & $3 ~\textfn{infraction\_in\_prison}$ \\ 
  $\scriptsize{+}$ & $3 ~\textfn{prior\_arrest\_for\_local\_ord}$ & $\scriptsize{+}$ & $2 ~\textfn{prior\_arrest\_for\_property}$ & $\scriptsize{+}$ & $2 ~\textfn{prior\_arrest\_for\_fatal\_violence}$ \\ 
  $\scriptsize{+}$ & $\textfn{prior\_arrest\_with\_firearms\_involved}$ & $\scriptsize{-}$ & $7 ~\textfn{age\_at\_release$\geq$40}$ & $\scriptsize{-}$ & $7$ \\ 
  \end{tabularx}
\end{center}
%
%
This is the best Lasso model for \textds{general\_violence}. This model has a test TPR/FPR of 79.7$\%$/45.5$\%$, and a mean 5-CV validation TPR/FPR of 77.3$\%$/45.7$\%$.
\begin{center}
\scriptsize
\begin{tabularx}{\textwidth}{p{1mm}lp{1mm}lp{1mm}l}
   & $0.90 ~\textfn{prior\_arrest\_for\_other\_violence}$ & $\scriptsize{+}$ & $0.35 ~\textfn{prior\_arrest\_for\_property}$ & $\scriptsize{+}$ & $0.28 ~\textfn{prior\_arrest\_for\_misdemeanor}$ \\ 
  $\scriptsize{+}$ & $0.28 ~\textfn{age\_at\_release\_18\_to\_24}$ & $\scriptsize{+}$ & $0.24 ~\textfn{prior\_arrest\_for\_public\_order}$ & $\scriptsize{+}$ & $0.20 ~\textfn{age\_1st\_arrest$\leq$17}$ \\ 
  $\scriptsize{+}$ & $0.20 ~\textfn{released\_unconditonal}$ & $\scriptsize{+}$ & $0.17 ~\textfn{age\_1st\_confinement\_18\_to\_24}$ & $\scriptsize{+}$ & $0.16 ~\textfn{alcohol\_abuse}$ \\ 
  $\scriptsize{+}$ & $0.14 ~\textfn{prior\_arrest\_for\_fatal\_violence}$ & $\scriptsize{+}$ & $0.14 ~\textfn{age\_1st\_confinement$\leq$17}$ & $\scriptsize{+}$ & $0.10 ~\textfn{prior\_arrest\_for\_felony}$ \\ 
  $\scriptsize{+}$ & $0.10 ~\textfn{prior\_arrests$\geq$5}$ & $\scriptsize{+}$ & $0.10 ~\textfn{prior\_arrest\_with\_firearms\_involved}$ & $\scriptsize{+}$ & $0.10 ~\textfn{age\_1st\_arrest\_18\_to\_24}$ \\ 
  $\scriptsize{+}$ & $0.09 ~\textfn{infraction\_in\_prison}$ & $\scriptsize{+}$ & $0.04 ~\textfn{time\_served$\leq$6mo}$ & $\scriptsize{+}$ & $0.03 ~\textfn{time\_served\_7\_to\_12mo}$ \\ 
  $\scriptsize{+}$ & $2.89 \times 10^{-03} ~\textfn{prior\_arrest\_for\_drugs}$ & $\scriptsize{-}$ & $0.72 ~\textfn{age\_at\_release$\geq$40}$ & $\scriptsize{-}$ & $0.41 ~\textfn{female}$ \\ 
  $\scriptsize{-}$ & $0.27 ~\textfn{age\_at\_release\_30\_to\_39}$ & $\scriptsize{-}$ & $0.15 ~\textfn{prior\_arrest\_with\_child\_involved}$ & $\scriptsize{-}$ & $0.07 ~\textfn{age\_1st\_confinement$\geq$40}$ \\ 
  $\scriptsize{-}$ & $0.05 ~\textfn{age\_1st\_arrest$\geq$40}$ & $\scriptsize{-}$ & $0.01 ~\textfn{time\_served\_25\_to\_60mo}$ & $\scriptsize{-}$ & $1.84 \times 10^{-03} ~\textfn{age\_1st\_confinement\_30\_to\_39}$ \\ 
  $\scriptsize{-}$ & $1.19$ &  &  &  &  \\ 
  \end{tabularx}
\end{center}
This is the best Ridge model for \textds{general\_violence}. This model has a test TPR/FPR of 81.4$\%$/48.1$\%$, and a mean 5-CV validation TPR/FPR of 80.0$\%$/48.5$\%$.
\begin{center}
\scriptsize
\begin{tabularx}{\textwidth}{p{1mm}lp{1mm}lp{1mm}l}
   & $0.62 ~\textfn{prior\_arrest\_for\_other\_violence}$ & $\scriptsize{+}$ & $0.27 ~\textfn{age\_at\_release\_18\_to\_24}$ & $\scriptsize{+}$ & $0.24 ~\textfn{prior\_arrest\_for\_property}$ \\ 
  $\scriptsize{+}$ & $0.23 ~\textfn{prior\_arrest\_for\_misdemeanor}$ & $\scriptsize{+}$ & $0.19 ~\textfn{age\_1st\_confinement\_18\_to\_24}$ & $\scriptsize{+}$ & $0.18 ~\textfn{prior\_arrest\_for\_public\_order}$ \\ 
  $\scriptsize{+}$ & $0.17 ~\textfn{age\_1st\_arrest$\leq$17}$ & $\scriptsize{+}$ & $0.14 ~\textfn{prior\_arrest\_for\_multiple\_types\_of\_crime}$ & $\scriptsize{+}$ & $0.13 ~\textfn{released\_unconditonal}$ \\ 
  $\scriptsize{+}$ & $0.13 ~\textfn{prior\_arrests$\geq$5}$ & $\scriptsize{+}$ & $0.13 ~\textfn{prior\_arrest\_for\_felony}$ & $\scriptsize{+}$ & $0.12 ~\textfn{prior\_arrest\_with\_firearms\_involved}$ \\ 
  $\scriptsize{+}$ & $0.11 ~\textfn{age\_1st\_confinement$\leq$17}$ & $\scriptsize{+}$ & $0.11 ~\textfn{alcohol\_abuse}$ & $\scriptsize{+}$ & $0.10 ~\textfn{age\_at\_release\_25\_to\_29}$ \\ 
  $\scriptsize{+}$ & $0.10 ~\textfn{prior\_arrest\_for\_fatal\_violence}$ & $\scriptsize{+}$ & $0.09 ~\textfn{infraction\_in\_prison}$ & $\scriptsize{+}$ & $0.08 ~\textfn{age\_1st\_arrest\_18\_to\_24}$ \\ 
  $\scriptsize{+}$ & $0.07 ~\textfn{prior\_arrest\_for\_domestic\_violence}$ & $\scriptsize{+}$ & $0.05 ~\textfn{drug\_abuse}$ & $\scriptsize{+}$ & $0.05 ~\textfn{time\_served$\leq$6mo}$ \\ 
  $\scriptsize{+}$ & $0.05 ~\textfn{prior\_arrest\_for\_local\_ord}$ & $\scriptsize{+}$ & $0.04 ~\textfn{time\_served\_7\_to\_12mo}$ & $\scriptsize{+}$ & $0.04 ~\textfn{age\_at\_release$\leq$17}$ \\ 
  $\scriptsize{+}$ & $0.03 ~\textfn{prior\_arrests$\geq$2}$ & $\scriptsize{+}$ & $0.03 ~\textfn{multiple\_prior\_prb\_or\_fine}$ & $\scriptsize{+}$ & $0.02 ~\textfn{multiple\_prior\_jail\_time}$ \\ 
  $\scriptsize{+}$ & $0.01 ~\textfn{prior\_arrest\_for\_drugs}$ & $\scriptsize{+}$ & $3.41 \times 10^{-03} ~\textfn{no\_prior\_arrests}$ & $\scriptsize{-}$ & $0.32 ~\textfn{age\_at\_release$\geq$40}$ \\ 
  $\scriptsize{-}$ & $0.20 ~\textfn{female}$ & $\scriptsize{-}$ & $0.18 ~\textfn{age\_1st\_confinement$\geq$40}$ & $\scriptsize{-}$ & $0.12 ~\textfn{prior\_arrest\_with\_child\_involved}$ \\ 
  $\scriptsize{-}$ & $0.12 ~\textfn{age\_1st\_arrest$\geq$40}$ & $\scriptsize{-}$ & $0.11 ~\textfn{age\_1st\_arrest\_30\_to\_39}$ & $\scriptsize{-}$ & $0.09 ~\textfn{age\_1st\_confinement\_30\_to\_39}$ \\ 
  $\scriptsize{-}$ & $0.08 ~\textfn{age\_at\_release\_30\_to\_39}$ & $\scriptsize{-}$ & $0.05 ~\textfn{age\_1st\_arrest\_25\_to\_29}$ & $\scriptsize{-}$ & $0.04 ~\textfn{prior\_arrest\_for\_sexual}$ \\ 
  $\scriptsize{-}$ & $0.04 ~\textfn{time\_served\_25\_to\_60mo}$ & $\scriptsize{-}$ & $0.03 ~\textfn{time\_served$\geq$61mo}$ & $\scriptsize{-}$ & $0.03 ~\textfn{released\_conditonal}$ \\ 
  $\scriptsize{-}$ & $0.03 ~\textfn{age\_1st\_confinement\_25\_to\_29}$ & $\scriptsize{-}$ & $0.02 ~\textfn{any\_prior\_prb\_or\_fine}$ & $\scriptsize{-}$ & $0.02 ~\textfn{time\_served\_13\_to\_24mo}$ \\ 
  $\scriptsize{-}$ & $5.89 \times 10^{-03} ~\textfn{multiple\_prior\_prison\_time}$ & $\scriptsize{-}$ & $3.60 \times 10^{-03} ~\textfn{any\_prior\_jail\_time}$ & $\scriptsize{-}$ & $3.47 \times 10^{-03} ~\textfn{prior\_arrests$\geq$1}$ \\ 
  $\scriptsize{-}$ & $1.13$ &  &  &  &  \\ 
  \end{tabularx}
\end{center}
\begin{figure}
	\newcommand{\calibinclude}[1]{\includegraphics[trim=0.2cm 0.4cm 0.0cm 1.1cm,clip,width=\fwidth]{#1}}
	\centering
	\calibinclude{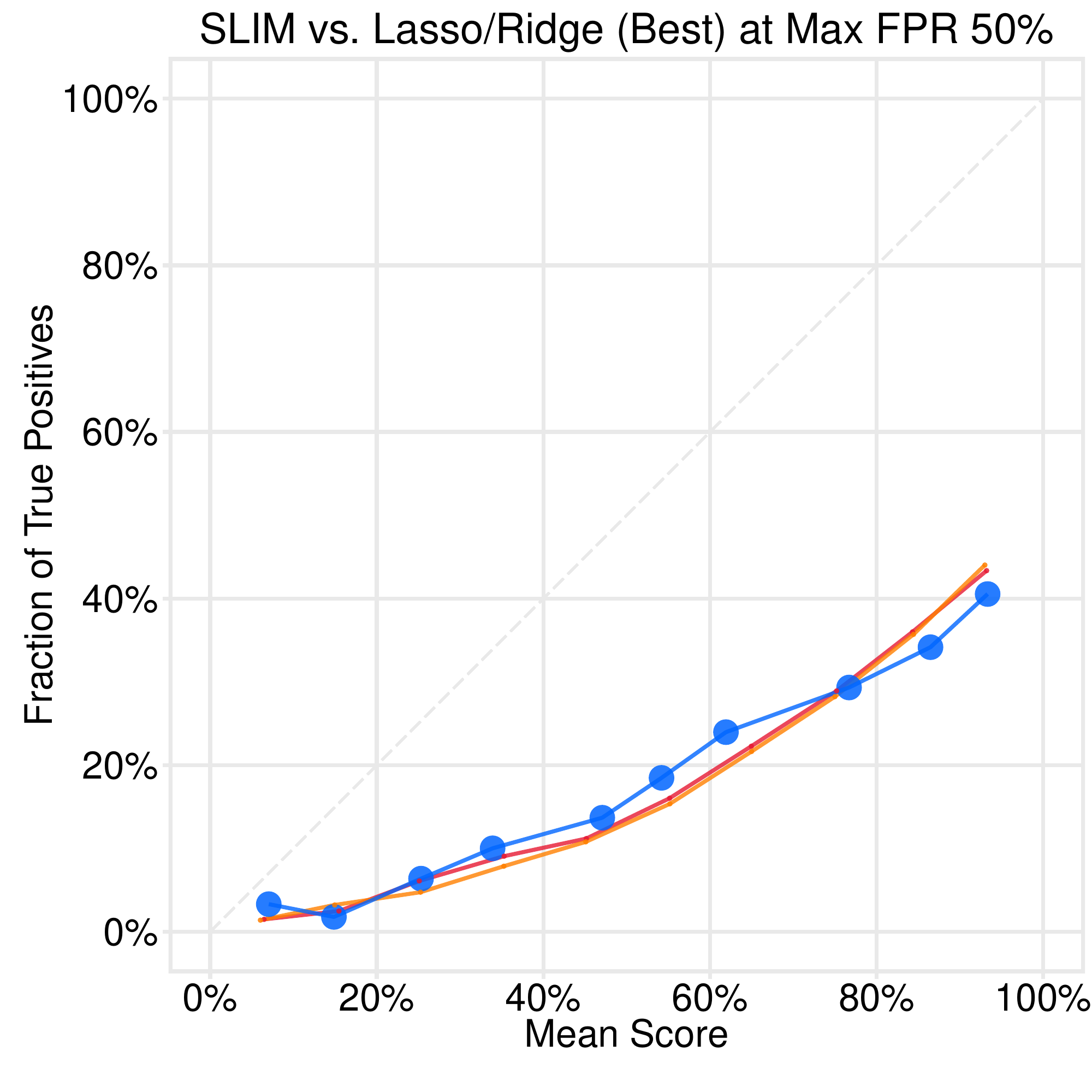}
	\includegraphics[trim=0.5in 0in 0.5in 0.3in,clip,width=0.9\textwidth]{figure/roc_plot_legend_calib}
	\vspace{-0.1in}
	\caption{Risk calibration plot for \textds{general\_violence}.}
	\label{Figure::generalViolenceCalibPlot}
\end{figure}

\clearpage
\FloatBarrier
\subsection{\textds{domestic\_violence}}
This is the SLIM model for \textds{domestic\_violence}. This model has a test TPR/FPR of 85.5$\%$/46.0$\%$, and a mean 5-CV validation TPR/FPR of 81.4$\%$/48.0$\%$.
\begin{center}
\scriptsize
\begin{tabularx}{\textwidth}{p{1mm}lp{1mm}lp{1mm}l}
   & $4 ~\textfn{prior\_arrest\_for\_misdemeanor}$ & $\scriptsize{+}$ & $3 ~\textfn{prior\_arrest\_for\_felony}$ & $\scriptsize{+}$ & $2 ~\textfn{prior\_arrest\_for\_domestic\_violence}$ \\ 
  $\scriptsize{+}$ & $\textfn{age\_1st\_confinement\_18\_to\_24}$ & $\scriptsize{-}$ & $5 ~\textfn{infraction\_in\_prison}$ & $\scriptsize{-}$ & $3$ \\ 
  \end{tabularx}
\end{center}
%
%
This is the best Lasso model for \textds{domestic\_violence}. This model has a test TPR/FPR of 87.0$\%$/45.8$\%$, and a mean 5-CV validation TPR/FPR of 84.5$\%$/45.8$\%$.
\begin{center}
\scriptsize
\begin{tabularx}{\textwidth}{p{1mm}lp{1mm}lp{1mm}l}
   & $0.88 ~\textfn{prior\_arrest\_for\_misdemeanor}$ & $\scriptsize{+}$ & $0.73 ~\textfn{prior\_arrest\_for\_domestic\_violence}$ & $\scriptsize{+}$ & $0.73 ~\textfn{prior\_arrest\_for\_felony}$ \\ 
  $\scriptsize{+}$ & $0.66 ~\textfn{prior\_arrest\_for\_other\_violence}$ & $\scriptsize{+}$ & $0.54 ~\textfn{released\_unconditonal}$ & $\scriptsize{+}$ & $0.32 ~\textfn{age\_1st\_confinement\_18\_to\_24}$ \\ 
  $\scriptsize{+}$ & $0.24 ~\textfn{multiple\_prior\_prb\_or\_fine}$ & $\scriptsize{+}$ & $0.21 ~\textfn{alcohol\_abuse}$ & $\scriptsize{+}$ & $0.17 ~\textfn{prior\_arrest\_for\_sexual}$ \\ 
  $\scriptsize{+}$ & $0.16 ~\textfn{prior\_arrests$\geq$5}$ & $\scriptsize{+}$ & $0.16 ~\textfn{prior\_arrest\_with\_firearms\_involved}$ & $\scriptsize{+}$ & $0.08 ~\textfn{age\_at\_release\_18\_to\_24}$ \\ 
  $\scriptsize{+}$ & $0.06 ~\textfn{no\_prior\_arrests}$ & $\scriptsize{+}$ & $0.05 ~\textfn{time\_served\_7\_to\_12mo}$ & $\scriptsize{+}$ & $0.03 ~\textfn{prior\_arrest\_for\_property}$ \\ 
  $\scriptsize{+}$ & $0.01 ~\textfn{age\_1st\_arrest\_18\_to\_24}$ & $\scriptsize{+}$ & $0.01 ~\textfn{prior\_arrest\_for\_public\_order}$ & $\scriptsize{-}$ & $1.09 ~\textfn{infraction\_in\_prison}$ \\ 
  $\scriptsize{-}$ & $0.54 ~\textfn{age\_at\_release$\geq$40}$ & $\scriptsize{-}$ & $0.47 ~\textfn{drug\_abuse}$ & $\scriptsize{-}$ & $0.40 ~\textfn{multiple\_prior\_prison\_time}$ \\ 
  $\scriptsize{-}$ & $0.31 ~\textfn{prior\_arrest\_with\_child\_involved}$ & $\scriptsize{-}$ & $0.28 ~\textfn{multiple\_prior\_jail\_time}$ & $\scriptsize{-}$ & $0.26 ~\textfn{female}$ \\ 
  $\scriptsize{-}$ & $0.20 ~\textfn{age\_1st\_confinement$\geq$40}$ & $\scriptsize{-}$ & $0.16 ~\textfn{any\_prior\_jail\_time}$ & $\scriptsize{-}$ & $0.07 ~\textfn{age\_1st\_arrest\_30\_to\_39}$ \\ 
  $\scriptsize{-}$ & $0.07 ~\textfn{any\_prior\_prb\_or\_fine}$ & $\scriptsize{-}$ & $0.06 ~\textfn{prior\_arrest\_for\_drugs}$ & $\scriptsize{-}$ & $0.06 ~\textfn{time\_served$\geq$61mo}$ \\ 
  $\scriptsize{-}$ & $4.48 \times 10^{-04} ~\textfn{time\_served\_25\_to\_60mo}$ & $\scriptsize{-}$ & $1.04$ &  &  \\ 
  \end{tabularx}
\end{center}
This is the best Ridge model for \textds{domestic\_violence}. This model has a test TPR/FPR of 87.0$\%$/47.7$\%$, and a mean 5-CV validation TPR/FPR of 85.2$\%$/47.5$\%$.
\begin{center}
\scriptsize
\begin{tabularx}{\textwidth}{p{1mm}lp{1mm}lp{1mm}l}
   & $0.76 ~\textfn{prior\_arrest\_for\_misdemeanor}$ & $\scriptsize{+}$ & $0.59 ~\textfn{prior\_arrest\_for\_other\_violence}$ & $\scriptsize{+}$ & $0.57 ~\textfn{prior\_arrest\_for\_domestic\_violence}$ \\ 
  $\scriptsize{+}$ & $0.54 ~\textfn{prior\_arrest\_for\_felony}$ & $\scriptsize{+}$ & $0.40 ~\textfn{released\_unconditonal}$ & $\scriptsize{+}$ & $0.27 ~\textfn{age\_1st\_confinement\_18\_to\_24}$ \\ 
  $\scriptsize{+}$ & $0.27 ~\textfn{multiple\_prior\_prb\_or\_fine}$ & $\scriptsize{+}$ & $0.21 ~\textfn{prior\_arrest\_for\_sexual}$ & $\scriptsize{+}$ & $0.19 ~\textfn{prior\_arrest\_with\_firearms\_involved}$ \\ 
  $\scriptsize{+}$ & $0.18 ~\textfn{alcohol\_abuse}$ & $\scriptsize{+}$ & $0.18 ~\textfn{prior\_arrests$\geq$5}$ & $\scriptsize{+}$ & $0.17 ~\textfn{age\_at\_release\_18\_to\_24}$ \\ 
  $\scriptsize{+}$ & $0.15 ~\textfn{prior\_arrest\_for\_local\_ord}$ & $\scriptsize{+}$ & $0.12 ~\textfn{age\_at\_release\_25\_to\_29}$ & $\scriptsize{+}$ & $0.11 ~\textfn{time\_served\_7\_to\_12mo}$ \\ 
  $\scriptsize{+}$ & $0.10 ~\textfn{prior\_arrest\_for\_property}$ & $\scriptsize{+}$ & $0.10 ~\textfn{prior\_arrest\_for\_fatal\_violence}$ & $\scriptsize{+}$ & $0.10 ~\textfn{no\_prior\_arrests}$ \\ 
  $\scriptsize{+}$ & $0.08 ~\textfn{age\_at\_release\_30\_to\_39}$ & $\scriptsize{+}$ & $0.07 ~\textfn{prior\_arrest\_for\_multiple\_types\_of\_crime}$ & $\scriptsize{+}$ & $0.07 ~\textfn{age\_1st\_arrest$\leq$17}$ \\ 
  $\scriptsize{+}$ & $0.07 ~\textfn{age\_1st\_arrest\_18\_to\_24}$ & $\scriptsize{+}$ & $0.07 ~\textfn{prior\_arrest\_for\_public\_order}$ & $\scriptsize{+}$ & $0.05 ~\textfn{age\_1st\_arrest\_25\_to\_29}$ \\ 
  $\scriptsize{+}$ & $0.05 ~\textfn{time\_served$\leq$6mo}$ & $\scriptsize{+}$ & $0.05 ~\textfn{time\_served\_13\_to\_24mo}$ & $\scriptsize{+}$ & $0.05 ~\textfn{prior\_arrests$\geq$2}$ \\ 
  $\scriptsize{+}$ & $3.08 \times 10^{-03} ~\textfn{age\_1st\_confinement\_30\_to\_39}$ & $\scriptsize{-}$ & $0.86 ~\textfn{infraction\_in\_prison}$ & $\scriptsize{-}$ & $0.40 ~\textfn{drug\_abuse}$ \\ 
  $\scriptsize{-}$ & $0.39 ~\textfn{multiple\_prior\_prison\_time}$ & $\scriptsize{-}$ & $0.36 ~\textfn{age\_at\_release$\geq$40}$ & $\scriptsize{-}$ & $0.26 ~\textfn{prior\_arrest\_with\_child\_involved}$ \\ 
  $\scriptsize{-}$ & $0.25 ~\textfn{multiple\_prior\_jail\_time}$ & $\scriptsize{-}$ & $0.25 ~\textfn{female}$ & $\scriptsize{-}$ & $0.24 ~\textfn{age\_1st\_confinement$\geq$40}$ \\ 
  $\scriptsize{-}$ & $0.19 ~\textfn{any\_prior\_jail\_time}$ & $\scriptsize{-}$ & $0.14 ~\textfn{time\_served$\geq$61mo}$ & $\scriptsize{-}$ & $0.12 ~\textfn{age\_1st\_arrest\_30\_to\_39}$ \\ 
  $\scriptsize{-}$ & $0.10 ~\textfn{any\_prior\_prb\_or\_fine}$ & $\scriptsize{-}$ & $0.10 ~\textfn{age\_1st\_arrest$\geq$40}$ & $\scriptsize{-}$ & $0.10 ~\textfn{prior\_arrests$\geq$1}$ \\ 
  $\scriptsize{-}$ & $0.08 ~\textfn{prior\_arrest\_for\_drugs}$ & $\scriptsize{-}$ & $0.06 ~\textfn{age\_1st\_confinement\_25\_to\_29}$ & $\scriptsize{-}$ & $0.05 ~\textfn{time\_served\_25\_to\_60mo}$ \\ 
  $\scriptsize{-}$ & $0.04 ~\textfn{released\_conditonal}$ & $\scriptsize{-}$ & $0.04 ~\textfn{age\_at\_release$\leq$17}$ & $\scriptsize{-}$ & $0.02 ~\textfn{age\_1st\_confinement$\leq$17}$ \\ 
  $\scriptsize{-}$ & $1.01$ &  &  &  &  \\ 
  \end{tabularx}
\end{center}
\begin{figure}
	\newcommand{\calibinclude}[1]{\includegraphics[trim=0.2cm 0.4cm 0.0cm 1.1cm,clip,width=\fwidth]{#1}}
	\centering
	\calibinclude{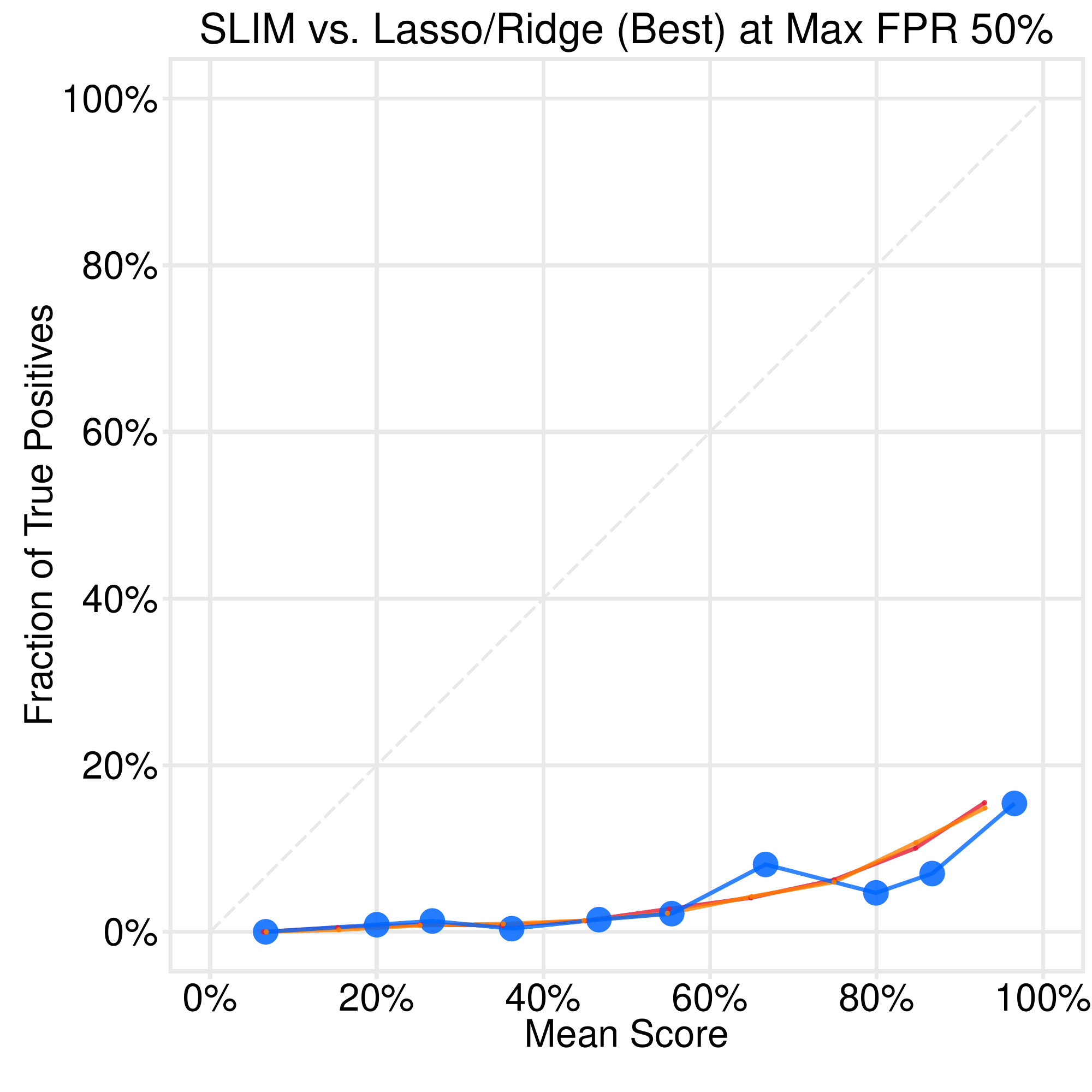}
	\includegraphics[trim=0.5in 0in 0.5in 0.3in,clip,width=0.9\textwidth]{figure/roc_plot_legend_calib}
	\vspace{-0.1in}
	\caption{Risk calibration plot for \textds{domestic\_violence}.}
	\label{Figure::domesticViolenceCalibPlot}
\end{figure}

\clearpage
\FloatBarrier
\subsection{\textds{sexual\_violence}}
This is the SLIM model for \textds{sexual\_violence}. This model has a test TPR/FPR of 44.3$\%$/17.7$\%$, and a mean 5-CV validation TPR/FPR of 43.7$\%$/19.9$\%$.
\begin{center}
\scriptsize
\begin{tabularx}{\textwidth}{p{1mm}lp{1mm}lp{1mm}l}
   & $3 ~\textfn{prior\_arrest\_for\_sexual}$ & $\scriptsize{+}$ & $\textfn{prior\_arrests$\geq$5}$ & $\scriptsize{+}$ & $\textfn{multiple\_prior\_jail\_time}$ \\ 
  $\scriptsize{-}$ & $2 ~\textfn{no\_prior\_arrests}$ & $\scriptsize{-}$ & $\textfn{prior\_arrest\_for\_multiple\_types\_of\_crime}$ & $\scriptsize{-}$ & $2$ \\ 
  \end{tabularx}
\end{center}
%
%
This is the best Lasso model for \textds{sexual\_violence}. This model has a test TPR/FPR of 46.9$\%$/18.1$\%$, and a mean 5-CV validation TPR/FPR of 43.7$\%$/17.9$\%$.
\begin{center}
\scriptsize
\begin{tabularx}{\textwidth}{p{1mm}lp{1mm}lp{1mm}l}
   & $1.10 ~\textfn{prior\_arrest\_for\_sexual}$ & $\scriptsize{+}$ & $0.40 ~\textfn{prior\_arrest\_for\_other\_violence}$ & $\scriptsize{+}$ & $0.27 ~\textfn{age\_1st\_confinement\_18\_to\_24}$ \\ 
  $\scriptsize{+}$ & $0.27 ~\textfn{prior\_arrest\_for\_felony}$ & $\scriptsize{+}$ & $0.19 ~\textfn{prior\_arrest\_with\_child\_involved}$ & $\scriptsize{+}$ & $0.19 ~\textfn{infraction\_in\_prison}$ \\ 
  $\scriptsize{+}$ & $0.12 ~\textfn{prior\_arrest\_for\_property}$ & $\scriptsize{+}$ & $0.09 ~\textfn{prior\_arrest\_for\_public\_order}$ & $\scriptsize{+}$ & $0.07 ~\textfn{prior\_arrests$\geq$5}$ \\ 
  $\scriptsize{+}$ & $0.03 ~\textfn{age\_1st\_confinement$\leq$17}$ & $\scriptsize{+}$ & $0.02 ~\textfn{age\_1st\_arrest$\leq$17}$ & $\scriptsize{+}$ & $8.11 \times 10^{-04} ~\textfn{prior\_arrest\_for\_fatal\_violence}$ \\ 
  $\scriptsize{-}$ & $0.58 ~\textfn{female}$ & $\scriptsize{-}$ & $0.25 ~\textfn{age\_at\_release$\geq$40}$ & $\scriptsize{-}$ & $0.23 ~\textfn{prior\_arrest\_for\_drugs}$ \\ 
  $\scriptsize{-}$ & $0.05 ~\textfn{any\_prior\_prb\_or\_fine}$ & $\scriptsize{-}$ & $0.05 ~\textfn{drug\_abuse}$ & $\scriptsize{-}$ & $0.01 ~\textfn{time\_served\_25\_to\_60mo}$ \\ 
  $\scriptsize{-}$ & $0.01 ~\textfn{prior\_arrest\_for\_misdemeanor}$ & $\scriptsize{-}$ & $5.85 \times 10^{-03} ~\textfn{age\_1st\_confinement\_30\_to\_39}$ & $\scriptsize{-}$ & $1.63$ \\ 
  \end{tabularx}
\end{center}%
This is the best Ridge model for \textds{sexual\_violence}. This model has a test TPR/FPR of 48.6$\%$/19.3$\%$, and a mean 5-CV validation TPR/FPR of 44.9$\%$/19.4$\%$.
\begin{center}
\scriptsize
\begin{tabularx}{\textwidth}{p{1mm}lp{1mm}lp{1mm}l}
   & $0.92 ~\textfn{prior\_arrest\_for\_sexual}$ & $\scriptsize{+}$ & $0.35 ~\textfn{prior\_arrest\_for\_other\_violence}$ & $\scriptsize{+}$ & $0.30 ~\textfn{prior\_arrest\_for\_felony}$ \\ 
  $\scriptsize{+}$ & $0.28 ~\textfn{prior\_arrest\_with\_child\_involved}$ & $\scriptsize{+}$ & $0.20 ~\textfn{age\_1st\_confinement\_18\_to\_24}$ & $\scriptsize{+}$ & $0.18 ~\textfn{infraction\_in\_prison}$ \\ 
  $\scriptsize{+}$ & $0.14 ~\textfn{prior\_arrest\_for\_property}$ & $\scriptsize{+}$ & $0.14 ~\textfn{prior\_arrest\_for\_public\_order}$ & $\scriptsize{+}$ & $0.13 ~\textfn{age\_1st\_confinement$\leq$17}$ \\ 
  $\scriptsize{+}$ & $0.12 ~\textfn{prior\_arrests$\geq$5}$ & $\scriptsize{+}$ & $0.10 ~\textfn{prior\_arrest\_for\_fatal\_violence}$ & $\scriptsize{+}$ & $0.07 ~\textfn{age\_at\_release\_18\_to\_24}$ \\ 
  $\scriptsize{+}$ & $0.07 ~\textfn{time\_served$\geq$61mo}$ & $\scriptsize{+}$ & $0.07 ~\textfn{age\_1st\_arrest$\leq$17}$ & $\scriptsize{+}$ & $0.07 ~\textfn{prior\_arrest\_for\_local\_ord}$ \\ 
  $\scriptsize{+}$ & $0.06 ~\textfn{any\_prior\_jail\_time}$ & $\scriptsize{+}$ & $0.05 ~\textfn{age\_at\_release\_30\_to\_39}$ & $\scriptsize{+}$ & $0.04 ~\textfn{age\_at\_release\_25\_to\_29}$ \\ 
  $\scriptsize{+}$ & $0.04 ~\textfn{multiple\_prior\_prb\_or\_fine}$ & $\scriptsize{+}$ & $0.03 ~\textfn{time\_served\_13\_to\_24mo}$ & $\scriptsize{+}$ & $0.03 ~\textfn{released\_conditonal}$ \\ 
  $\scriptsize{+}$ & $0.03 ~\textfn{released\_unconditonal}$ & $\scriptsize{+}$ & $0.02 ~\textfn{age\_1st\_arrest\_18\_to\_24}$ & $\scriptsize{+}$ & $9.63 \times 10^{-03} ~\textfn{age\_1st\_arrest\_30\_to\_39}$ \\ 
  $\scriptsize{+}$ & $7.60 \times 10^{-03} ~\textfn{prior\_arrests$\geq$1}$ & $\scriptsize{+}$ & $6.27 \times 10^{-03} ~\textfn{age\_at\_release$\leq$17}$ & $\scriptsize{-}$ & $0.37 ~\textfn{female}$ \\ 
  $\scriptsize{-}$ & $0.25 ~\textfn{prior\_arrest\_for\_drugs}$ & $\scriptsize{-}$ & $0.16 ~\textfn{age\_at\_release$\geq$40}$ & $\scriptsize{-}$ & $0.11 ~\textfn{age\_1st\_confinement$\geq$40}$ \\ 
  $\scriptsize{-}$ & $0.11 ~\textfn{any\_prior\_prb\_or\_fine}$ & $\scriptsize{-}$ & $0.11 ~\textfn{age\_1st\_confinement\_30\_to\_39}$ & $\scriptsize{-}$ & $0.09 ~\textfn{drug\_abuse}$ \\ 
  $\scriptsize{-}$ & $0.09 ~\textfn{age\_1st\_arrest$\geq$40}$ & $\scriptsize{-}$ & $0.07 ~\textfn{prior\_arrest\_for\_misdemeanor}$ & $\scriptsize{-}$ & $0.06 ~\textfn{multiple\_prior\_jail\_time}$ \\ 
  $\scriptsize{-}$ & $0.05 ~\textfn{time\_served\_25\_to\_60mo}$ & $\scriptsize{-}$ & $0.04 ~\textfn{prior\_arrests$\geq$2}$ & $\scriptsize{-}$ & $0.04 ~\textfn{alcohol\_abuse}$ \\ 
  $\scriptsize{-}$ & $0.04 ~\textfn{time\_served\_7\_to\_12mo}$ & $\scriptsize{-}$ & $0.03 ~\textfn{prior\_arrest\_for\_multiple\_types\_of\_crime}$ & $\scriptsize{-}$ & $0.02 ~\textfn{prior\_arrest\_for\_domestic\_violence}$ \\ 
  $\scriptsize{-}$ & $0.02 ~\textfn{time\_served$\leq$6mo}$ & $\scriptsize{-}$ & $0.02 ~\textfn{age\_1st\_confinement\_25\_to\_29}$ & $\scriptsize{-}$ & $0.02 ~\textfn{multiple\_prior\_prison\_time}$ \\ 
  $\scriptsize{-}$ & $7.46 \times 10^{-03} ~\textfn{no\_prior\_arrests}$ & $\scriptsize{-}$ & $5.79 \times 10^{-03} ~\textfn{age\_1st\_arrest\_25\_to\_29}$ & $\scriptsize{-}$ & $4.60 \times 10^{-03} ~\textfn{prior\_arrest\_with\_firearms\_involved}$ \\ 
  $\scriptsize{-}$ & $1.47$ &  &  &  &  \\ 
  \end{tabularx}
\end{center}
\begin{figure}
	\newcommand{\calibinclude}[1]{\includegraphics[trim=0.2cm 0.4cm 0.0cm 1.1cm,clip,width=\fwidth]{#1}}
	\centering
	\calibinclude{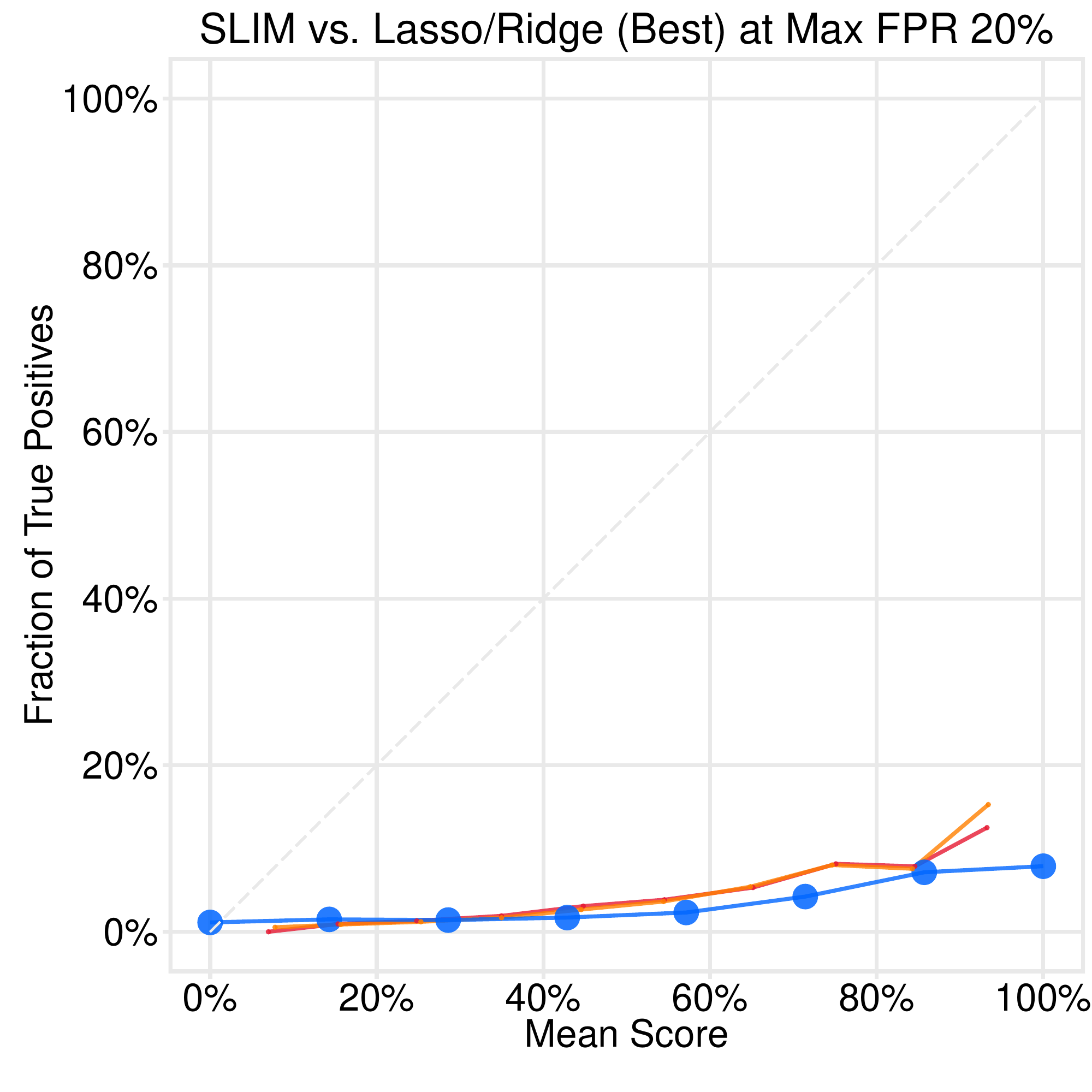}
	\includegraphics[trim=0.5in 0in 0.5in 0.3in,clip,width=0.9\textwidth]{figure/roc_plot_legend_calib}
	\vspace{-0.1in}
	\caption{Risk calibration plot for \textds{sexual\_violence}.}
	\label{Figure::sexualCalibPlot}
\end{figure}

\clearpage
\FloatBarrier
\subsection{\textds{fatal\_violence}}
This is the SLIM model for \textds{fatal\_violence}. This model has a test TPR/FPR of 55.4$\%$/35.5$\%$, and a mean 5-CV validation TPR/FPR of 64.2$\%$/42.4$\%$.
\begin{center}
\scriptsize
\begin{tabularx}{\textwidth}{p{1mm}lp{1mm}lp{1mm}l}
   & $5 ~\textfn{age\_1st\_confinement$\leq$17}$ & $\scriptsize{+}$ & $3 ~\textfn{prior\_arrest\_with\_firearms\_involved}$ & $\scriptsize{+}$ & $2 ~\textfn{age\_1st\_confinement\_18\_to\_24}$ \\ 
  $\scriptsize{+}$ & $2 ~\textfn{prior\_arrest\_for\_felony}$ & $\scriptsize{+}$ & $\textfn{age\_at\_release\_18\_to\_24}$ & $\scriptsize{+}$ & $\textfn{prior\_arrest\_for\_drugs}$ \\ 
  $\scriptsize{-}$ & $4$ &  &  &  &  \\ 
  \end{tabularx}
\end{center}
This is the best Lasso model for \textds{fatal\_violence}. This model has a test TPR/FPR of 68.9$\%$/44.5$\%$, and a mean 5-CV validation TPR/FPR of 67.6$\%$/42.4$\%$.
\begin{center}
\scriptsize
\begin{tabularx}{\textwidth}{p{1mm}lp{1mm}lp{1mm}l}
   & $1.52 ~\textfn{age\_1st\_confinement$\leq$17}$ & $\scriptsize{+}$ & $1.47 ~\textfn{age\_at\_release$\leq$17}$ & $\scriptsize{+}$ & $1.12 ~\textfn{prior\_arrest\_for\_felony}$ \\ 
  $\scriptsize{+}$ & $0.73 ~\textfn{age\_at\_release\_18\_to\_24}$ & $\scriptsize{+}$ & $0.69 ~\textfn{alcohol\_abuse}$ & $\scriptsize{+}$ & $0.66 ~\textfn{prior\_arrests$\geq$5}$ \\ 
  $\scriptsize{+}$ & $0.60 ~\textfn{prior\_arrest\_for\_fatal\_violence}$ & $\scriptsize{+}$ & $0.54 ~\textfn{age\_1st\_confinement\_18\_to\_24}$ & $\scriptsize{+}$ & $0.47 ~\textfn{prior\_arrest\_with\_firearms\_involved}$ \\ 
  $\scriptsize{+}$ & $0.39 ~\textfn{prior\_arrest\_for\_drugs}$ & $\scriptsize{+}$ & $0.38 ~\textfn{age\_1st\_confinement\_25\_to\_29}$ & $\scriptsize{+}$ & $0.35 ~\textfn{prior\_arrest\_for\_other\_violence}$ \\ 
  $\scriptsize{+}$ & $0.35 ~\textfn{age\_1st\_arrest$\leq$17}$ & $\scriptsize{+}$ & $0.34 ~\textfn{prior\_arrest\_for\_public\_order}$ & $\scriptsize{+}$ & $0.31 ~\textfn{prior\_arrest\_for\_multiple\_types\_of\_crime}$ \\ 
  $\scriptsize{+}$ & $0.28 ~\textfn{no\_prior\_arrests}$ & $\scriptsize{+}$ & $0.26 ~\textfn{age\_1st\_arrest\_25\_to\_29}$ & $\scriptsize{+}$ & $0.24 ~\textfn{age\_1st\_confinement\_30\_to\_39}$ \\ 
  $\scriptsize{+}$ & $0.20 ~\textfn{multiple\_prior\_prison\_time}$ & $\scriptsize{+}$ & $0.19 ~\textfn{prior\_arrest\_for\_property}$ & $\scriptsize{+}$ & $0.18 ~\textfn{prior\_arrest\_for\_sexual}$ \\ 
  $\scriptsize{+}$ & $0.11 ~\textfn{any\_prior\_prb\_or\_fine}$ & $\scriptsize{+}$ & $0.07 ~\textfn{time\_served\_7\_to\_12mo}$ & $\scriptsize{+}$ & $0.07 ~\textfn{time\_served$\leq$6mo}$ \\ 
  $\scriptsize{+}$ & $0.04 ~\textfn{age\_1st\_arrest\_18\_to\_24}$ & $\scriptsize{-}$ & $2.69 ~\textfn{age\_1st\_arrest$\geq$40}$ & $\scriptsize{-}$ & $1.68 ~\textfn{female}$ \\ 
  $\scriptsize{-}$ & $0.70 ~\textfn{drug\_abuse}$ & $\scriptsize{-}$ & $0.55 ~\textfn{infraction\_in\_prison}$ & $\scriptsize{-}$ & $0.50 ~\textfn{time\_served$\geq$61mo}$ \\ 
  $\scriptsize{-}$ & $0.42 ~\textfn{released\_conditonal}$ & $\scriptsize{-}$ & $0.39 ~\textfn{prior\_arrests$\geq$2}$ & $\scriptsize{-}$ & $0.36 ~\textfn{age\_at\_release$\geq$40}$ \\ 
  $\scriptsize{-}$ & $0.34 ~\textfn{prior\_arrest\_for\_misdemeanor}$ & $\scriptsize{-}$ & $0.33 ~\textfn{prior\_arrest\_with\_child\_involved}$ & $\scriptsize{-}$ & $0.29 ~\textfn{multiple\_prior\_prb\_or\_fine}$ \\ 
  $\scriptsize{-}$ & $0.24 ~\textfn{multiple\_prior\_jail\_time}$ & $\scriptsize{-}$ & $0.16 ~\textfn{released\_unconditonal}$ & $\scriptsize{-}$ & $0.13 ~\textfn{time\_served\_13\_to\_24mo}$ \\ 
  $\scriptsize{-}$ & $0.08 ~\textfn{age\_at\_release\_30\_to\_39}$ & $\scriptsize{-}$ & $0.08 ~\textfn{prior\_arrest\_for\_domestic\_violence}$ & $\scriptsize{-}$ & $0.02 ~\textfn{prior\_arrests$\geq$1}$ \\ 
  $\scriptsize{-}$ & $2.00$ &  &  &  &  \\ 
  \end{tabularx}
\end{center}
This is the best Ridge model for \textds{fatal\_violence}. This model has a test TPR/FPR of 62.2$\%$/34.0$\%$, and a mean 5-CV validation TPR/FPR of 60.1$\%$/33.0$\%$.
\begin{center}
\scriptsize
\begin{tabularx}{\textwidth}{p{1mm}lp{1mm}lp{1mm}l}
   & $0.55 ~\textfn{prior\_arrest\_for\_felony}$ & $\scriptsize{+}$ & $0.54 ~\textfn{age\_1st\_confinement$\leq$17}$ & $\scriptsize{+}$ & $0.45 ~\textfn{age\_at\_release\_18\_to\_24}$ \\ 
  $\scriptsize{+}$ & $0.39 ~\textfn{age\_1st\_arrest$\leq$17}$ & $\scriptsize{+}$ & $0.39 ~\textfn{prior\_arrest\_for\_fatal\_violence}$ & $\scriptsize{+}$ & $0.35 ~\textfn{prior\_arrests$\geq$5}$ \\ 
  $\scriptsize{+}$ & $0.35 ~\textfn{prior\_arrest\_with\_firearms\_involved}$ & $\scriptsize{+}$ & $0.29 ~\textfn{prior\_arrest\_for\_other\_violence}$ & $\scriptsize{+}$ & $0.29 ~\textfn{prior\_arrest\_for\_drugs}$ \\ 
  $\scriptsize{+}$ & $0.26 ~\textfn{prior\_arrest\_for\_public\_order}$ & $\scriptsize{+}$ & $0.25 ~\textfn{alcohol\_abuse}$ & $\scriptsize{+}$ & $0.24 ~\textfn{prior\_arrest\_for\_multiple\_types\_of\_crime}$ \\ 
  $\scriptsize{+}$ & $0.19 ~\textfn{age\_at\_release$\leq$17}$ & $\scriptsize{+}$ & $0.16 ~\textfn{multiple\_prior\_prison\_time}$ & $\scriptsize{+}$ & $0.16 ~\textfn{prior\_arrest\_for\_property}$ \\ 
  $\scriptsize{+}$ & $0.15 ~\textfn{time\_served\_7\_to\_12mo}$ & $\scriptsize{+}$ & $0.14 ~\textfn{time\_served$\leq$6mo}$ & $\scriptsize{+}$ & $0.12 ~\textfn{age\_1st\_confinement\_18\_to\_24}$ \\ 
  $\scriptsize{+}$ & $0.10 ~\textfn{any\_prior\_prb\_or\_fine}$ & $\scriptsize{+}$ & $0.08 ~\textfn{prior\_arrest\_for\_sexual}$ & $\scriptsize{+}$ & $0.06 ~\textfn{released\_unconditonal}$ \\ 
  $\scriptsize{+}$ & $0.06 ~\textfn{no\_prior\_arrests}$ & $\scriptsize{+}$ & $0.06 ~\textfn{time\_served\_25\_to\_60mo}$ & $\scriptsize{+}$ & $0.05 ~\textfn{age\_1st\_arrest\_25\_to\_29}$ \\ 
  $\scriptsize{+}$ & $0.03 ~\textfn{prior\_arrest\_for\_local\_ord}$ & $\scriptsize{-}$ & $0.51 ~\textfn{female}$ & $\scriptsize{-}$ & $0.42 ~\textfn{age\_at\_release$\geq$40}$ \\ 
  $\scriptsize{-}$ & $0.35 ~\textfn{drug\_abuse}$ & $\scriptsize{-}$ & $0.30 ~\textfn{infraction\_in\_prison}$ & $\scriptsize{-}$ & $0.29 ~\textfn{age\_1st\_arrest$\geq$40}$ \\ 
  $\scriptsize{-}$ & $0.28 ~\textfn{age\_1st\_confinement$\geq$40}$ & $\scriptsize{-}$ & $0.25 ~\textfn{time\_served$\geq$61mo}$ & $\scriptsize{-}$ & $0.20 ~\textfn{multiple\_prior\_prb\_or\_fine}$ \\ 
  $\scriptsize{-}$ & $0.19 ~\textfn{multiple\_prior\_jail\_time}$ & $\scriptsize{-}$ & $0.17 ~\textfn{prior\_arrest\_with\_child\_involved}$ & $\scriptsize{-}$ & $0.16 ~\textfn{prior\_arrest\_for\_misdemeanor}$ \\ 
  $\scriptsize{-}$ & $0.16 ~\textfn{age\_at\_release\_30\_to\_39}$ & $\scriptsize{-}$ & $0.15 ~\textfn{released\_conditonal}$ & $\scriptsize{-}$ & $0.14 ~\textfn{prior\_arrests$\geq$2}$ \\ 
  $\scriptsize{-}$ & $0.14 ~\textfn{age\_1st\_confinement\_30\_to\_39}$ & $\scriptsize{-}$ & $0.12 ~\textfn{age\_1st\_arrest\_30\_to\_39}$ & $\scriptsize{-}$ & $0.07 ~\textfn{time\_served\_13\_to\_24mo}$ \\ 
  $\scriptsize{-}$ & $0.06 ~\textfn{age\_at\_release\_25\_to\_29}$ & $\scriptsize{-}$ & $0.06 ~\textfn{age\_1st\_confinement\_25\_to\_29}$ & $\scriptsize{-}$ & $0.06 ~\textfn{prior\_arrests$\geq$1}$ \\ 
  $\scriptsize{-}$ & $0.01 ~\textfn{prior\_arrest\_for\_domestic\_violence}$ & $\scriptsize{-}$ & $0.01 ~\textfn{any\_prior\_jail\_time}$ & $\scriptsize{-}$ & $8.27 \times 10^{-03} ~\textfn{age\_1st\_arrest\_18\_to\_24}$ \\ 
  $\scriptsize{-}$ & $1.33$ &  &  &  &  \\ 
  \end{tabularx}
\end{center}
\begin{figure}
	\newcommand{\calibinclude}[1]{\includegraphics[trim=0.2cm 0.4cm 0.0cm 1.1cm,clip,width=\fwidth]{#1}}
	\centering
	\calibinclude{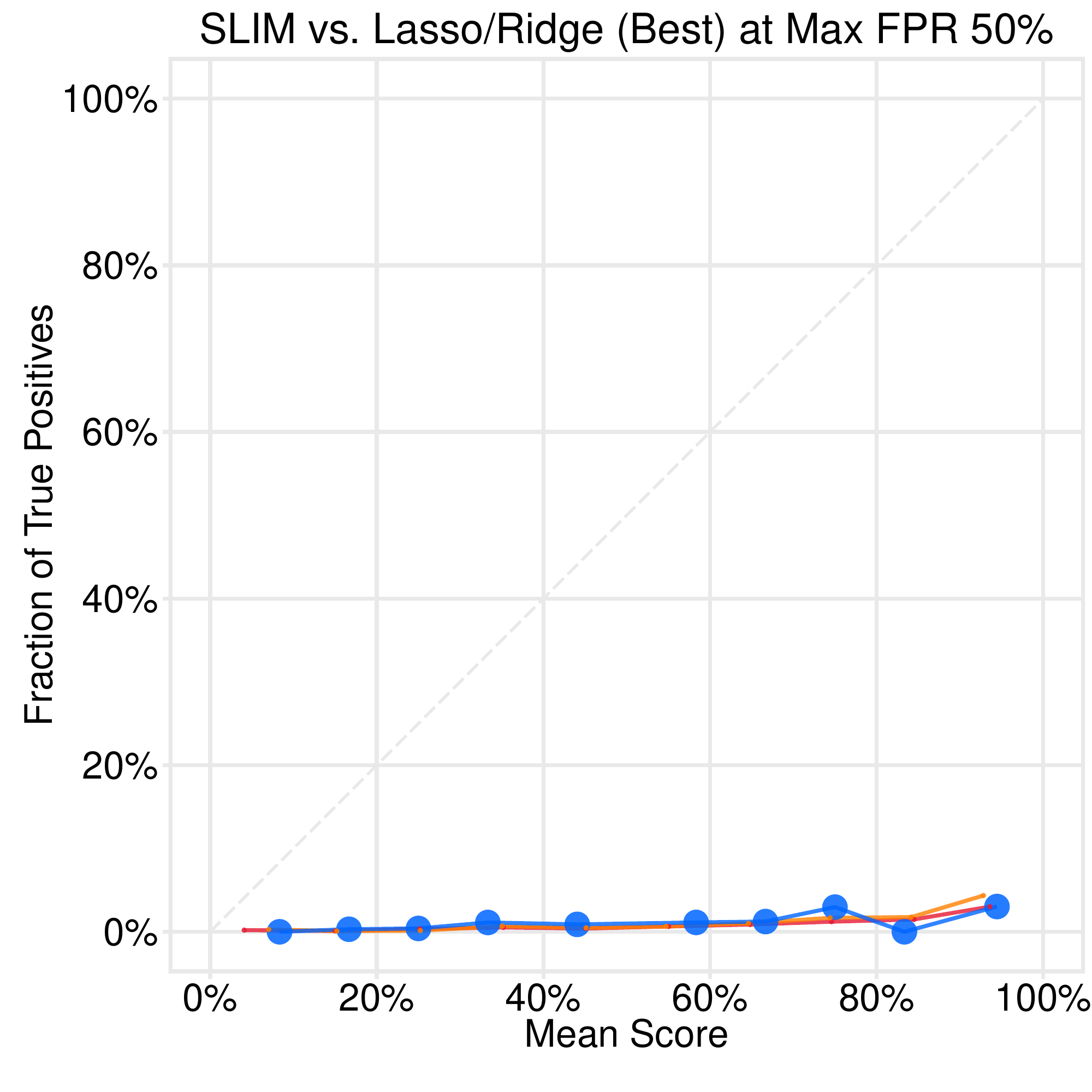}
	\includegraphics[trim=0.5in 0in 0.5in 0.3in,clip,width=0.9\textwidth]{figure/roc_plot_legend_calib}
	\vspace{-0.1in}
	\caption{Risk calibration plot for \textds{fatal\_violence}.}
	\label{Figure::fatalViolenceCalibPlot}
\end{figure}


\clearpage
\FloatBarrier
\section{Additional Results on the Trade-off between Accuracy and Interpretability}
\label{Appendix::InterpretabilityTradeoffs}
In the experiments in Section \ref{Sec::Results}, we used SLIM to fit models from a highly constrained space (i.e., models with at most 8 non-zero integer coefficients between -10 and 10). Here, we present evidence to show that baseline methods cannot attain the same level of accuracy or risk calibration when they are used to fit models from a slightly less constrained model space (i.e, model with at most 8 non-zero coefficients, 8 leaves or 8 rules).
%

Table \ref{Table::testAUCwithInterpretability} shows the test AUC of each method when they are used to fit a models with a model size of 8 or less. Trivial models of size 1 are also omitted. Table \ref{Table::differencetestAUCwithInterpretability} shows the percentage change in test AUC for the methods due to the model size restriction. For all models other than SLIM, the predictive accuracy was compromised with the size constraint. We see that C5.0R and C5.0T are unable to produce a suitably sparse model for some of the problems since their implementation does not provide control over model sparsity. Note that we have omitted results for Ridge because it could not produce a model with fewer than 8 coefficients for all prediction problems (see Section \ref{Sec::InterpretabilityResults} for explanation).  

\begin{table}
	\caption{\label{Table::testAUCwithInterpretability}Test AUC on all prediction problems when transparent methods are restricted to models with at most 8 coefficients, 8 leaves or 8 rules.}
	\scriptsize
	\renewcommand{\arraystretch}{1.2}
	\resizebox{0.7\textwidth}{!} {
	\centering
		\begin{tabular}{lccccc}
			\toprule
			\bfcell{c}{Prediction Problem} & \bfcell{c}{Lasso} & \bfcell{c}{C5.0R} & \bfcell{c}{C5.0T} & \bfcell{c}{CART} & \bfcell{c}{SLIM} \\ 
			\toprule
			\texttt{arrest} & \cell{c}{0.70} & - & - & \cell{c}{0.66} & \cell{c}{0.72} \\ 
			\midrule \texttt{drug} & \cell{c}{0.71} & - & - & \cell{c}{0.50} & \cell{c}{0.74} \\  
			\midrule \texttt{general\_violence} & \cell{c}{0.70} & \cell{c}{0.50} & \cell{c}{0.50} & \cell{c}{0.50} & \cell{c}{0.71} \\ 
			\midrule \texttt{domestic\_violence} & \cell{c}{0.74} & - & - & \cell{c}{0.50} & \cell{c}{0.76} \\
			\midrule \texttt{sexual\_violence} & \cell{c}{0.70} & - & - & \cell{c}{0.50} & \cell{c}{0.70} \\ 
			\midrule \texttt{fatal\_violence} & \cell{c}{0.60} & - & - & \cell{c}{0.50} & \cell{c}{0.62} \\
			\bottomrule \end{tabular}
		
	}
\end{table}

\begin{table}
	\caption{\label{Table::differencetestAUCwithInterpretability}Percentage in test AUC with respect to SLIM's model on all prediction problems when transparent methods are restricted to models with at most 8 coefficients, 8 leaves or 8 rules.}
	\scriptsize
	\renewcommand{\arraystretch}{1.2}
	\resizebox{0.7\textwidth}{!} {\centering
		\begin{tabular}{lccccc}
			\toprule
			\bfcell{c}{Prediction Problem} & \bfcell{c}{Lasso} & \bfcell{c}{C5.0R} & \bfcell{c}{C5.0T} & \bfcell{c}{CART} & \bfcell{c}{SLIM} \\ 
			\toprule
			\texttt{arrest} & \cell{c}{-3.8$\%$} & - & - & \cell{c}{-2.8$\%$} & \cell{c}{0.0$\%$} \\
			\midrule \texttt{drug} & \cell{c}{-4.0$\%$} & - & - & \cell{c}{-15.7$\%$} & \cell{c}{0.0$\%$} \\  
			\midrule \texttt{general\_violence} & \cell{c}{-2.2$\%$} & \cell{c}{-11.0$\%$} & \cell{c}{-12.7$\%$} & \cell{c}{-10.3$\%$} & \cell{c}{0.0$\%$} \\ 
			\midrule \texttt{domestic\_violence} & \cell{c}{-4.1$\%$} & - & - & \cell{c}{-5.4$\%$} & \cell{c}{0.0$\%$} \\ 
			\midrule \texttt{sexual\_violence} & \cell{c}{-2.2$\%$} & - & - & \cell{c}{-1.8$\%$} & \cell{c}{0.0$\%$} \\ 
			\midrule \texttt{fatal\_violence} & \cell{c}{-11.2$\%$} & - & - & \cell{c}{0.0$\%$} & \cell{c}{0.0$\%$} \\
			\bottomrule \end{tabular}	
	}
\end{table}

\section{Trade-off between Risk Calibration and Interpretability}
\label{Appendix::LossInCalibration}

\begin{figure}[]
	\newcommand{\calibinclude}[1]{\includegraphics[trim=0.2cm 0.5cm 0.0cm 1.1cm,clip,width=.95\fwidth]{#1}}
	\setlength{\tabcolsep}{0pt}
	\centering
	\tabulartitle{\textwidth}{\textds{arrest}}{\textds{drug}} \\ 
	\begin{tabular}{lr}
		\calibinclude{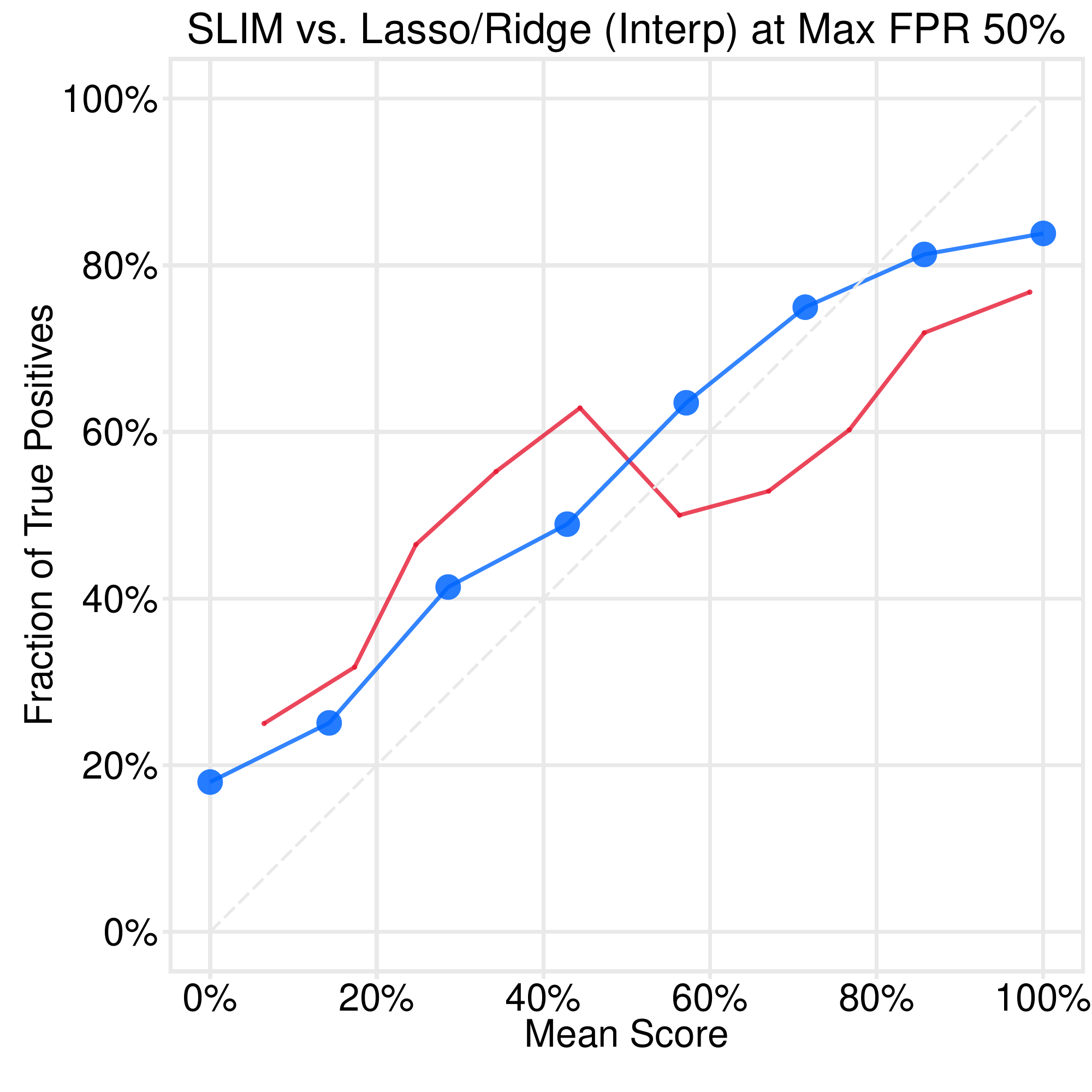} &
		\calibinclude{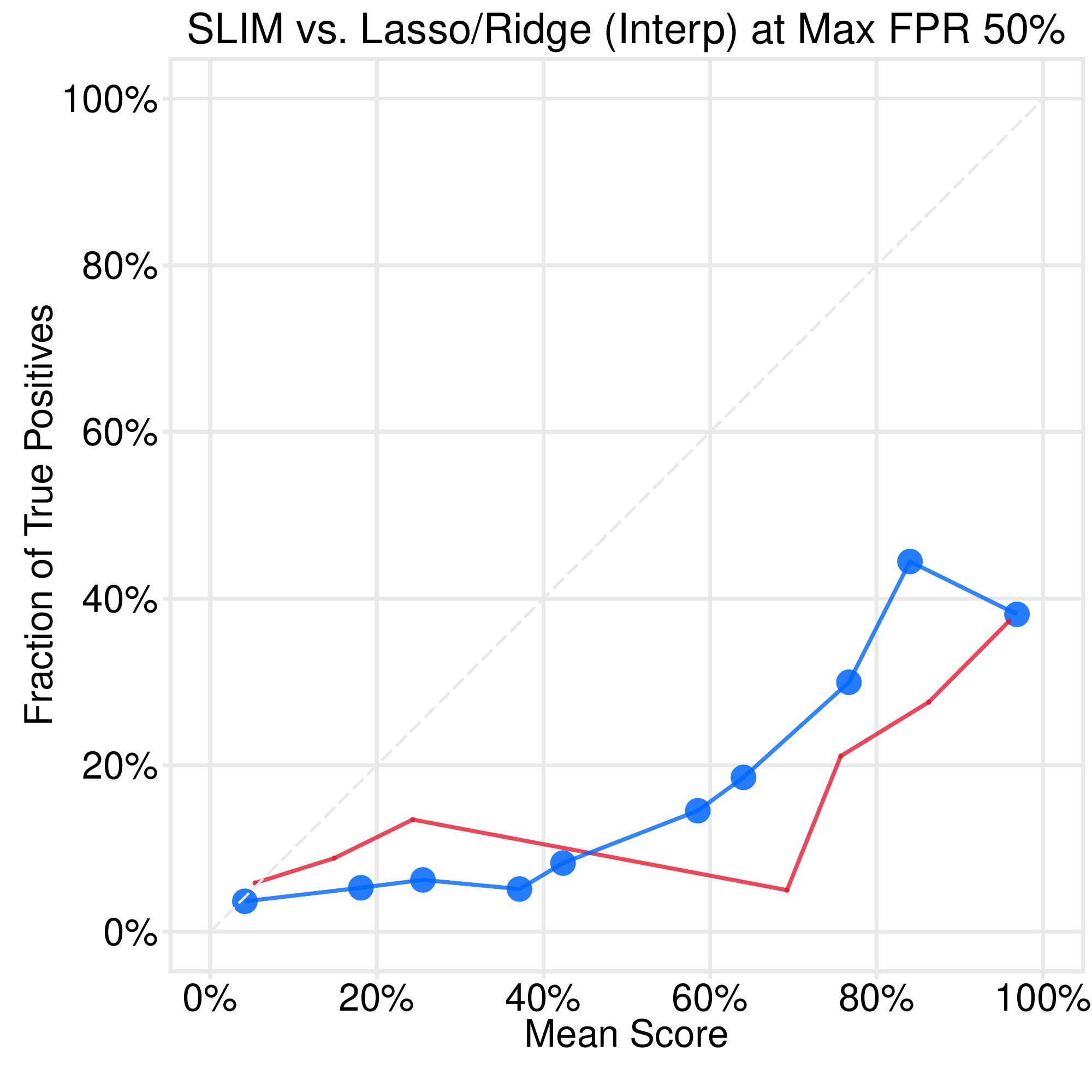}
	\end{tabular}
	\vspace{.3cm}
	\tabulartitle{\textwidth}{\textds{general\_violence}}{\textds{domestic\_violence}} \\ 
	\begin{tabular}{lr}
		\calibinclude{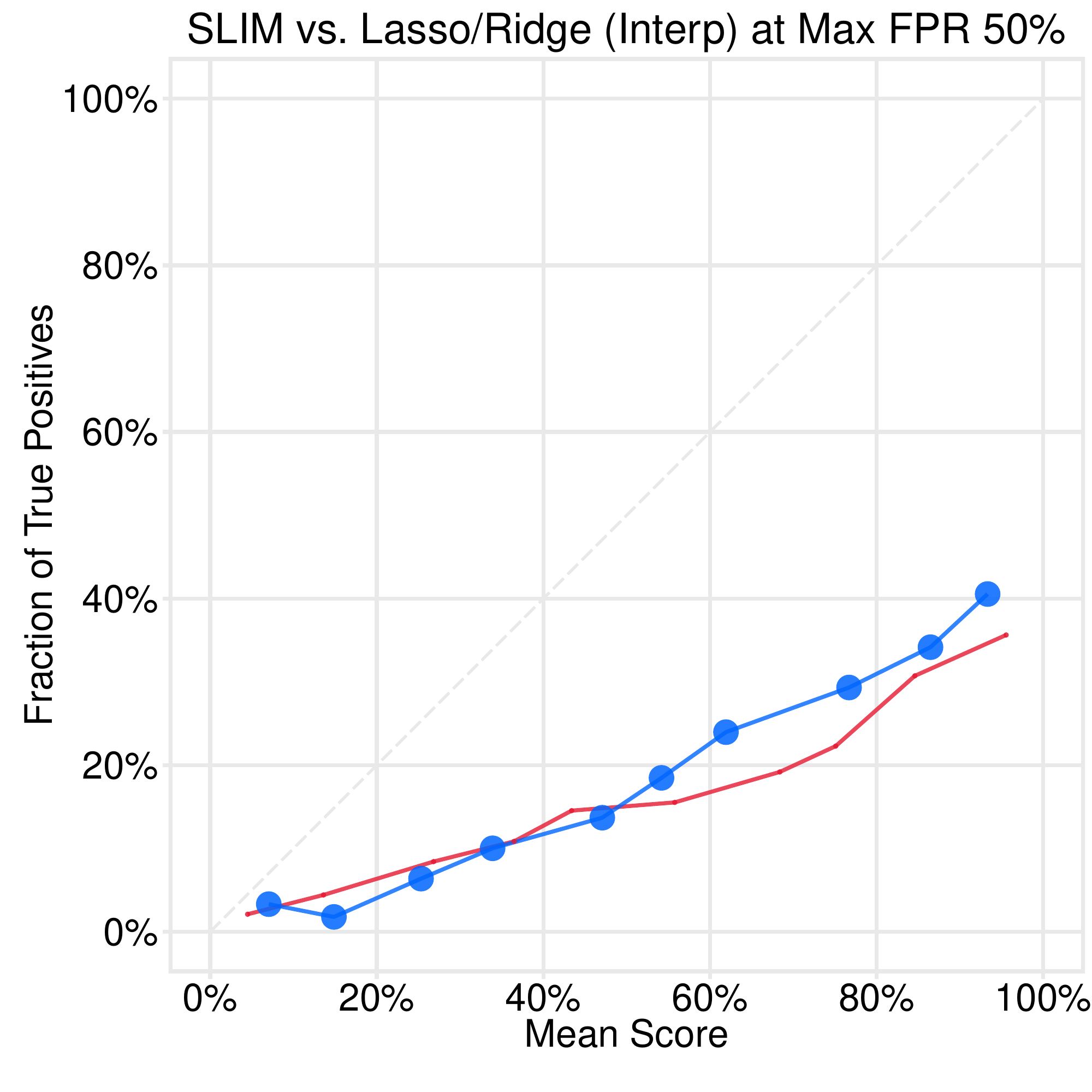} &
		\calibinclude{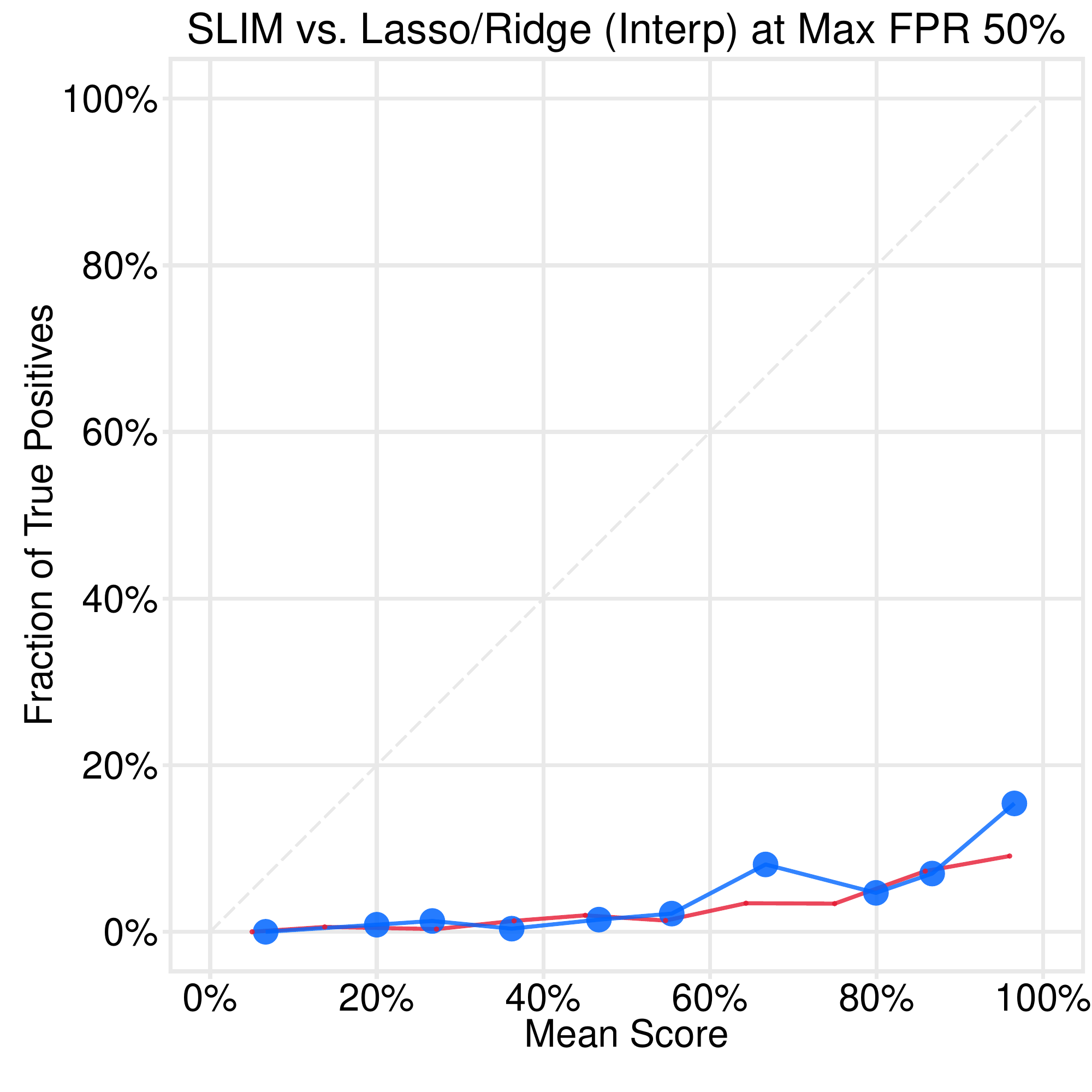}
	\end{tabular}
	\vspace{.3cm}
	\tabulartitle{\textwidth}{\textds{sexual\_violence}}{\textds{fatal\_violence}} \\ 
	\begin{tabular}{lr}
		\calibinclude{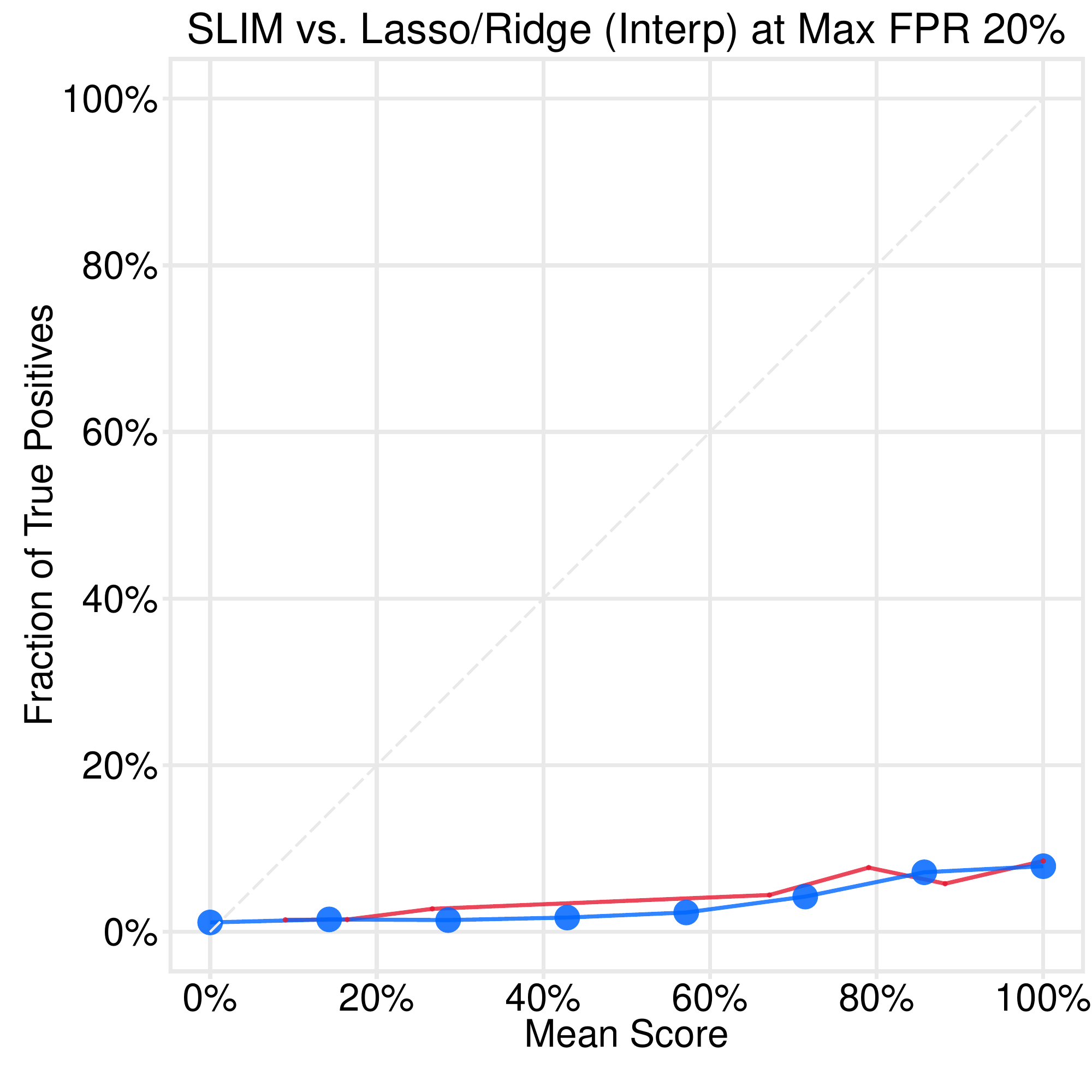} &
		\calibinclude{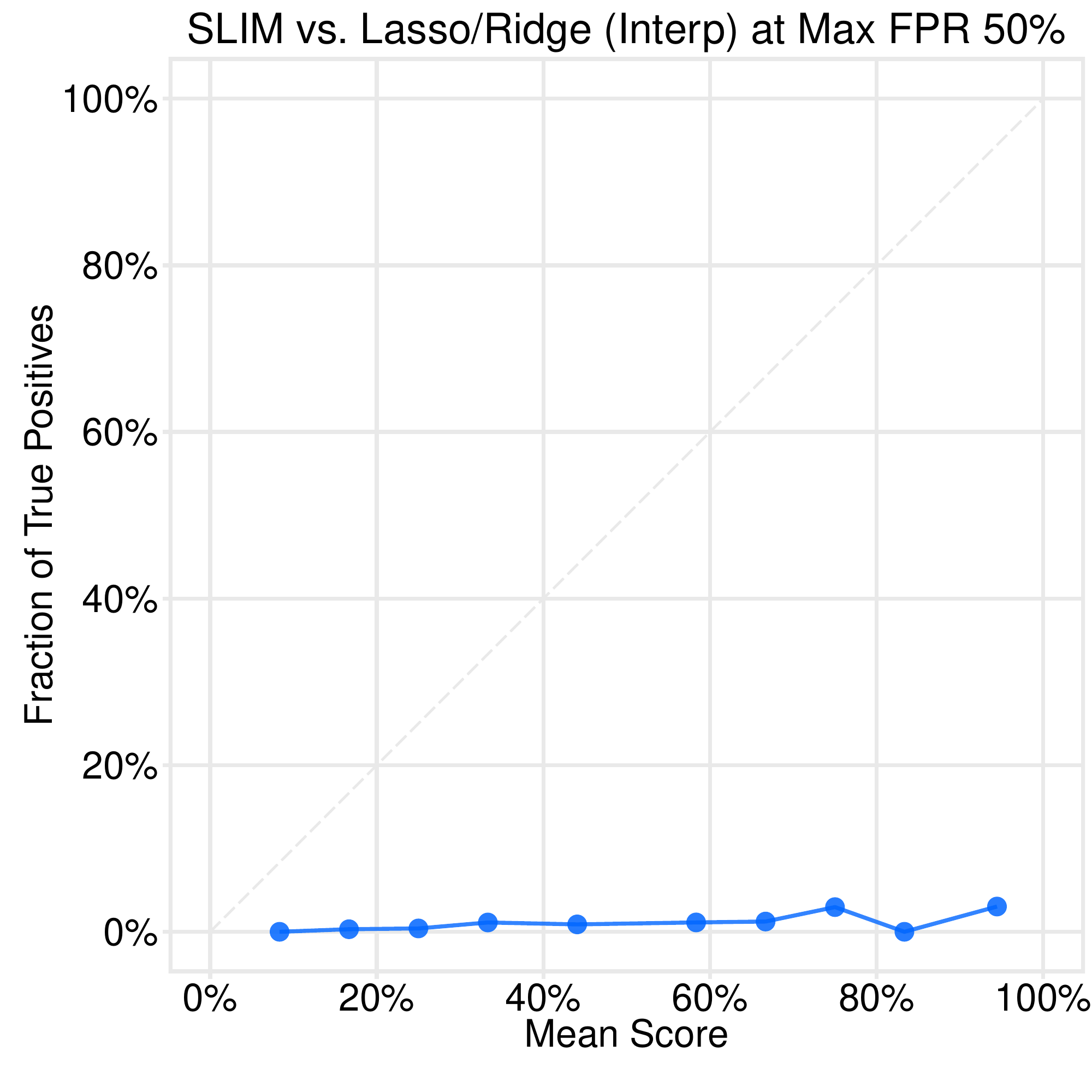}
	\end{tabular}
	\vspace{.2cm}
	\begin{tabular}{c}
		\includegraphics[trim=0.5in 0cm 0.5in 0.8cm,clip,width=0.9\textwidth] {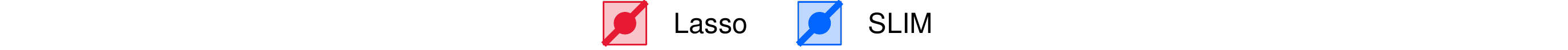}
	\end{tabular}
	\vspace{-0.2in}
	\caption{Risk calibration plots for transparent models with model size constrained to 8 or less.}
	\label{Figure::interpCalibPlots}
\end{figure}

Figure \ref{Figure::interpCalibPlots} shows the risk calibration plots of Lasso, Ridge, and SLIM for transparent models with model size constrained to 8 or less, chosen under the same decision criteria as Appendix \ref{Appendix::ModelBasedComparisons}. Ridge is not included because no such models are achievable, as discussed also in Appendix \ref{Appendix::InterpretabilityTradeoffs}. For Lasso, the risk calibration performance is worse in comparison to Figures \ref{Figure::drugCalibPlot}--\ref{Figure::fatalViolenceCalibPlot}. For \textds{fatal\_violence}, there was no Lasso model available at the desired decision point. 

\clearpage
\section{On the Predictive Accuracy of Baseline Methods with Continuous Input Variables}
\label{Sec::BinaryComparison}

In our experiments in Section \ref{Sec::Results}, we ran all methods with a dataset composed exclusively of binary input variables. That is, for each feature in the original database (e.g., $\textfn{prior\_arrests}$), we derived binary variables (e.g., $\textfn{no\_prior\_arrests}$, $\textfn{prior\_arrests $\geq$ 1}$ and so on)  and trained each method using these binary variables. It is possible that machine learning methods could potentially be hindered by this removal of information. Here, we investigate how the predictive accuracy of the baseline method would have been affected had we run these methods using continuous input variables (Appendix \ref{Sec::ContinuousComparison}) or both binary and continuous input variables (Appendix \ref{Sec::AllComparison}). In both cases, we find that the change in variable encoding results in a minor difference in performance.

\subsection{Change in Predictive Accuracy using Only Continuous Input Variables}
\label{Sec::ContinuousComparison}

Instead of using 48 input variables, we now have 25 continuous variables. Table \ref{Table::testContinuousAUC} summarizes the test AUC for all methods on all prediction problems when we use only continuous input variables. Table \ref{Table::pdifftestContinuousAUC} shows the percentage change in test AUC due to this change in encoding (i.e. from binary input variables to continuous input variables). The largest increases in predictive accuracy are 4.6\% for CART and 7.7\% for SVM RBF, while the biggest decrease in accuracy is $-19.6$\% for RF. 

Our results suggest that there is no uniform gain/loss in performance for most of the methods: for any given method, the test AUC increased slightly for at least one problem, and decreased slightly for at least another. Among the methods, CART saw the most uniform improvement, performing slightly better on 5 out of the 6 problems when continuous variables are used (though CART still performs poorly compared to other methods).

\begin{table}
\caption{\label{Table::testContinuousAUC}Test AUC for all methods on all datasets when features are encoded as continuous variables.}
\scriptsize
\renewcommand{\arraystretch}{1.2}
\resizebox{\textwidth}{!} {\centering
\begin{tabular}{lcccccccc}
  \toprule
\bfcell{c}{Prediction Problem} & \bfcell{c}{Lasso} & \bfcell{c}{Ridge} & \bfcell{c}{C5.0R} & \bfcell{c}{C5.0T} & \bfcell{c}{CART} & \bfcell{c}{RF} & \bfcell{c}{SVM RBF} & \bfcell{c}{SGB} \\ 
  \toprule
  \texttt{arrest} & \cell{c}{0.74} & \cell{c}{0.73} & \cell{c}{0.72} & \cell{c}{0.72} & \cell{c}{0.70} & \cell{c}{0.75} & \cell{c}{0.74} & \cell{c}{0.75} \\
   \midrule \texttt{drug} & \cell{c}{0.74} & \cell{c}{0.74} & \cell{c}{0.65} & \cell{c}{0.66} & \cell{c}{0.62} & \cell{c}{0.75} & \cell{c}{0.74} & \cell{c}{0.76} \\ 
   \midrule \texttt{general\_violence} & \cell{c}{0.71} & \cell{c}{0.70} & \cell{c}{0.54} & \cell{c}{0.58} & \cell{c}{0.55} & \cell{c}{0.69} & \cell{c}{0.69} & \cell{c}{0.71} \\ 
   \midrule \texttt{domestic\_violence} & \cell{c}{0.74} & \cell{c}{0.70} & \cell{c}{0.50} & \cell{c}{0.50} & \cell{c}{0.54} & \cell{c}{0.51} & \cell{c}{0.75} & \cell{c}{0.77} \\ 
   \midrule \texttt{sexual\_violence} & \cell{c}{0.70} & \cell{c}{0.68} & \cell{c}{0.50} & \cell{c}{0.50} & \cell{c}{0.52} & \cell{c}{0.51} & \cell{c}{0.68} & \cell{c}{0.71} \\ 
   \midrule \texttt{fatal\_violence} & \cell{c}{0.69} & \cell{c}{0.68} & \cell{c}{0.50} & \cell{c}{0.50} & \cell{c}{0.51} & \cell{c}{0.50} & \cell{c}{0.74} & \cell{c}{0.72} \\ 
   \bottomrule \end{tabular}
}
\end{table}
\begin{table}
\caption{\label{Table::pdifftestContinuousAUC}Percentage change in test AUC for all methods on all datasets when features are encoded as continuous variables instead of binary variables.}
\scriptsize
\renewcommand{\arraystretch}{1.2}
\resizebox{\textwidth}{!} {\centering
\begin{tabular}{lcccccccc}
  \toprule
\bfcell{c}{Prediction Problem} & \bfcell{c}{Lasso} & \bfcell{c}{Ridge} & \bfcell{c}{C5.0R} & \bfcell{c}{C5.0T} & \bfcell{c}{CART} & \bfcell{c}{RF} & \bfcell{c}{SVM RBF} & \bfcell{c}{SGB} \\ 
  \toprule 
  \texttt{arrest} & \cell{c}{1.7$\%$} & \cell{c}{0.1$\%$} & \cell{c}{0.0$\%$} & \cell{c}{0.1$\%$} & \cell{c}{2.4$\%$} & \cell{c}{2.7$\%$} & \cell{c}{3.0$\%$} & \cell{c}{1.7$\%$} \\ 
   \midrule \texttt{drug} & \cell{c}{-0.5$\%$} & \cell{c}{-0.3$\%$} & \cell{c}{2.5$\%$} & \cell{c}{4.2$\%$} & \cell{c}{4.6$\%$} & \cell{c}{0.6$\%$} & \cell{c}{0.7$\%$} & \cell{c}{1.9$\%$} \\  
   \midrule \texttt{general\_violence} & \cell{c}{-1.5$\%$} & \cell{c}{-2.6$\%$} & \cell{c}{-4.1$\%$} & \cell{c}{0.7$\%$} & \cell{c}{-1.3$\%$} & \cell{c}{-2.7$\%$} & \cell{c}{-2.2$\%$} & \cell{c}{-1.0$\%$} \\ 
   \midrule \texttt{domestic\_violence} & \cell{c}{-3.9$\%$} & \cell{c}{-8.7$\%$} & \cell{c}{-0.1$\%$} & \cell{c}{-0.1$\%$} & \cell{c}{1.6$\%$} & \cell{c}{-19.6$\%$} & \cell{c}{-3.1$\%$} & \cell{c}{-0.8$\%$} \\
   \midrule \texttt{sexual\_violence} & \cell{c}{-1.5$\%$} & \cell{c}{-5.1$\%$} & \cell{c}{0.0$\%$} & \cell{c}{0.0$\%$} & \cell{c}{2.0$\%$} & \cell{c}{-5.3$\%$} & \cell{c}{-2.7$\%$} & \cell{c}{0.9$\%$} \\ 
   \midrule \texttt{fatal\_violence} & \cell{c}{2.8$\%$} & \cell{c}{0.1$\%$} & \cell{c}{0.0$\%$} & \cell{c}{0.0$\%$} & \cell{c}{1.0$\%$} & \cell{c}{0.3$\%$} & \cell{c}{7.7$\%$} & \cell{c}{2.9$\%$} \\
   \bottomrule \end{tabular}
}
\end{table}

\clearpage
\subsection{Change in Predictive Accuracy using Both Binary and Continuous Input Variables}
\label{Sec::AllComparison}
\FloatBarrier

Instead of the original 48 variables, we now use a combination of 66 binary and continuous variables. Table \ref{Table::testAllAUC} summarizes the test AUC for all methods on all prediction problems when we used both binary and continuous input variables. Table \ref{Table::pdifftestAllAUC} shows the percentage change in test AUC due to this change in encoding (i.e., from binary input variables to both binary and continuous input variables). Most methods saw a slight AUC increase due to the addition of continuous variables, ranging from 0.2--6.3\%. The most significant increases are 3.3\% for CART and 6.3\% for C5.0T, while the largest decrease is $-16.0$\% for RF. In addition to RF, Ridge and SVM RBF all saw slight decreases with the inclusion. Similar to Appendix \ref{Sec::ContinuousComparison}, no uniform gain/loss in performance is seen. 

\begin{table}
\caption{\label{Table::testAllAUC}Test AUC for models created using both continuous and binary variables. }
\scriptsize
\renewcommand{\arraystretch}{1.2}
\resizebox{\textwidth}{!} {\centering
\begin{tabular}{lcccccccc}
  \toprule
\bfcell{c}{Prediction Problem} & \bfcell{c}{Lasso} & \bfcell{c}{Ridge} & \bfcell{c}{C5.0R} & \bfcell{c}{C5.0T} & \bfcell{c}{CART} & \bfcell{c}{RF} & \bfcell{c}{SVM RBF} & \bfcell{c}{SGB} \\ 
  \toprule
\texttt{arrest} & \cell{c}{0.74} & \cell{c}{0.73} & \cell{c}{0.72} & \cell{c}{0.72} & \cell{c}{0.69} & \cell{c}{0.75} & \cell{c}{0.73} & \cell{c}{0.75} \\ 
   \midrule \texttt{drug} & \cell{c}{0.75} & \cell{c}{0.74} & \cell{c}{0.65} & \cell{c}{0.67} & \cell{c}{0.61} & \cell{c}{0.76} & \cell{c}{0.75} & \cell{c}{0.76} \\  
   \midrule \texttt{general\_violence} & \cell{c}{0.72} & \cell{c}{0.71} & \cell{c}{0.58} & \cell{c}{0.58} & \cell{c}{0.56} & \cell{c}{0.72} & \cell{c}{0.71} & \cell{c}{0.73} \\ 
   \midrule \texttt{domestic\_violence} & \cell{c}{0.75} & \cell{c}{0.71} & \cell{c}{0.50} & \cell{c}{0.50} & \cell{c}{0.54} & \cell{c}{0.54} & \cell{c}{0.77} & \cell{c}{0.78} \\
   \midrule \texttt{sexual\_violence} & \cell{c}{0.71} & \cell{c}{0.69} & \cell{c}{0.50} & \cell{c}{0.50} & \cell{c}{0.52} & \cell{c}{0.50} & \cell{c}{0.71} & \cell{c}{0.71} \\ 
   \midrule \texttt{fatal\_violence} & \cell{c}{0.69} & \cell{c}{0.68} & \cell{c}{0.50} & \cell{c}{0.50} & \cell{c}{0.50} & \cell{c}{0.50} & \cell{c}{0.70} & \cell{c}{0.72} \\
   \bottomrule \end{tabular}

}
\end{table}
\begin{table}
\caption{\label{Table::pdifftestAllAUC}Percentage difference of test AUC for models created with both continuous and binary variables verses test AUC for models created with just binary variables.}
\scriptsize
\renewcommand{\arraystretch}{1.2}
\resizebox{\textwidth}{!} {\centering
\begin{tabular}{lcccccccc}
  \toprule
\bfcell{c}{Prediction Problem} & \bfcell{c}{Lasso} & \bfcell{c}{Ridge} & \bfcell{c}{C5.0R} & \bfcell{c}{C5.0T} & \bfcell{c}{CART} & \bfcell{c}{RF} & \bfcell{c}{SVM RBF} & \bfcell{c}{SGB} \\ 
  \toprule
\texttt{arrest} & \cell{c}{2.4$\%$} & \cell{c}{0.6$\%$} & \cell{c}{0.7$\%$} & \cell{c}{0.2$\%$} & \cell{c}{1.9$\%$} & \cell{c}{2.7$\%$} & \cell{c}{1.5$\%$} & \cell{c}{1.7$\%$} \\ 
   \midrule \texttt{drug} & \cell{c}{0.2$\%$} & \cell{c}{0.2$\%$} & \cell{c}{2.1$\%$} & \cell{c}{6.3$\%$} & \cell{c}{3.3$\%$} & \cell{c}{1.9$\%$} & \cell{c}{2.1$\%$} & \cell{c}{2.1$\%$} \\  
   \midrule \texttt{general\_violence} & \cell{c}{0.5$\%$} & \cell{c}{-1.7$\%$} & \cell{c}{2.5$\%$} & \cell{c}{0.7$\%$} & \cell{c}{-0.2$\%$} & \cell{c}{0.4$\%$} & \cell{c}{0.7$\%$} & \cell{c}{1.2$\%$} \\ 
   \midrule \texttt{domestic\_violence} & \cell{c}{-2.2$\%$} & \cell{c}{-7.4$\%$} & \cell{c}{0.0$\%$} & \cell{c}{0.0$\%$} & \cell{c}{1.3$\%$} & \cell{c}{-16.0$\%$} & \cell{c}{-0.5$\%$} & \cell{c}{0.4$\%$} \\
   \midrule \texttt{sexual\_violence} & \cell{c}{-0.4$\%$} & \cell{c}{-4.2$\%$} & \cell{c}{0.0$\%$} & \cell{c}{0.0$\%$} & \cell{c}{1.5$\%$} & \cell{c}{-6.4$\%$} & \cell{c}{1.9$\%$} & \cell{c}{0.6$\%$} \\ 
   \midrule \texttt{fatal\_violence} & \cell{c}{2.2$\%$} & \cell{c}{0.3$\%$} & \cell{c}{0.0$\%$} & \cell{c}{0.0$\%$} & \cell{c}{-0.3$\%$} & \cell{c}{0.3$\%$} & \cell{c}{2.7$\%$} & \cell{c}{2.8$\%$} \\
   \bottomrule \end{tabular}

}
\end{table}


\clearpage
\section{Association Rules}
\label{Sec::AssociationRules}
We produce insights more extensive than those in Section \ref{Sec::SimpleInsights} by mining \textit{association rules}. Association rules, also known as ``IF-THEN" rules, are small predictive models that can be produced using search techniques or optimization techniques.

\subsection{Terminology}

High quality association rules are characterized by large values of \textit{support}, \textit{confidence}, and \textit{lift}. To define this terminology, consider a rule such as ``IF $a$ THEN $b$." We denote this rule also as $a\rightarrow b$. 
The \textit{support} of $a\rightarrow b$ is the empirical probability $\hat{P}(a \textrm{ and } b)$,  that is, the proportion of observations where the conditions $a$ and $b$ are both satisfied. 
The \textit{confidence} of $a\rightarrow b$ is the empirical probability $\hat{P}(b|a)$, that is, the proportion of observations for which condition $b$ is satisfied given $a$ is satisfied. 
The \textit{lift} of $a\rightarrow b$ is the ratio $\frac{\hat{P}(b | a)}{\hat{P}(b)}$. Lift measures the ability of condition $a$ to ``target" the population where condition $b$ is satisfied: if the lift of $a \rightarrow b$ is equal to 1, then outcome $b$ could be predicted equally well if we had assumed that $a$ and $b$ were independent; if the lift of $a \rightarrow b$ greater than 1 then event $a$ has some effect on predicting event $b$. 

To illustrate these concepts, consider the following association rule:
\begin{center}
	IF \textfn{age\_at\_release\_18\_to\_24}  AND \textfn{prior\_arrests$\geq$5} THEN $y=+1$.
\end{center}
The support of this rule is 0.07, which means that 7\% of prisoners were released from prison between the ages of 18 to 24, had at least 5 prior arrests, and were arrested within 3 years of being released from prison.
The confidence of this rule is 0.83, which means that if a prisoner was released from prison between the ages of 18 to 24 and had at least 5 prior arrests, then there was an 83\% chance that this person would be arrested within 3 years of being released from prison.
Lastly, the lift of this rule is 1.41, which means that prisoners released from prison between the ages of 18 to 24 and had at least 5 prior arrests have a higher chance of being rearrested than other prisoners, i.e., the prisoners age at release and arrest history makes the conditional probability of arrest 1.41 times higher than if arrest was independent of these conditions.

\subsection{Rule Mining}\label{SecRuleMining}

We list 24 interesting association rules for the \textds{arrest} problem in Table \ref{Table::AssociationRules}. 
 These rules were generated with the apriori method in the \pkg{arules} package in R 3.1.1 \citep{arules2014}. Note that the choice of package does not matter, as mining rules through search techniques is deterministic, so all packages produce the same rules.

Here, the IF conditions are formulated using combinations of input variables (i.e. $x_j=1$ and $x_k=1$) and the THEN condition is that a prisoner is arrested within 3 years of being released from prison (i.e. a positive outcome $y=+1$).
The rules in Table \ref{Table::AssociationRules} have the highest levels of lift and confidence with a minimum support of 5\% (i.e., the rule applied to at least 1690 of the 33796 prisoners in our dataset). This threshold value was chosen so the rules do not reflect spurious correlations.
Rules A -- E were produced by mining the most powerful single-variable predictors for \textds{arrest}. Rules A -- E attain the highest lift among one-variable rules with a support of at least 5\% and a confidence of at least 0.70.
Rules F -- X were produced by mining two-variable rules that use at least one of the input variables from Rules A -- E that attain the highest possible lift, as well as support at least 5\% and confidence at least 0.75.
Out of all these rules, Rule F performs the best with a confidence of 0.83 and a lift of 1.41. As it turns out, Rule F is often exploited by some of the best models we find for \textds{arrest}, as we often find patterns similar to ``\textfn{age\_at\_release\_18\_to\_24} AND \textfn{prior\_arrests$\geq$5}" in our predictive models (see e.g., Figure \ref{Model::rearrestCARTModel} in Section \ref{Sec::InterpretabilityResults}).


Interesting observations can also be made from the discovered rules. Recall that jail is a much less severe punishment than prison. Considering Rule L and Rule M in Table \ref{Table::AssociationRules}, we can see that prisoners with multiple jail time and have \textit{any} past probations or fines are just as likely to be arrested as those with multiple jail time and multiple prior prison records -- despite \textfn{multiple\_prior\_prison\_time} being a indicator of much more severe past actions than \textfn{any\_prior\_probation\_or\_fine}. 
\newcolumntype{y}{>{\centering\arraybackslash}p{1.7cm}}
\newcommand{\and}[0]{AND}

\begin{table}
	\scriptsize
	\caption{\label{Table::AssociationRules}IF-THEN rules mined for \textds{arrest}. The THEN condition for each rules is the outcome $y = +1$, which indicates that a prisoner is arrested within 3 years of being released from prison.}
	\renewcommand{\arraystretch}{1}
	\centering
	\begin{tabular}{@{ }c@{}p{0mm}@{}r@{  }c@{  }l@{}y@{  }y@{     }y@{   }}
		\toprule  
		\textbf{Rule} & \multicolumn{4}{c}{\textbf{IF Condition}} &  \textbf{Lift} & \textbf{Support} & \textbf{Confidence} \\ 
		
		\toprule
		
		A & & \textfn{multiple\_prior\_jail\_time}  & & & 1.24 & 0.21 & 0.73  \\ 
		B & & \textfn{age\_1st\_arrest$\leq$17}  & & & 1.23 & 0.10 & 0.73    \\ 
		C & & \textfn{multiple\_prior\_probation\_or\_fine}  & & & 1.20 & 0.16 & 0.71    \\
		D & & \textfn{age\_at\_release\_18\_to\_24} & & & 1.20 & 0.14 & 0.71    \\ 
		E & & \textfn{prior\_arrests$\geq$5}  & & & 1.19 & 0.42 & 0.70    \\ 
		
		\midrule
		
		F & & \textfn{age\_at\_release\_18\_to\_24}   & \and & \textfn{prior\_arrests$\geq$5}  &   1.41 & 0.07 & 0.83 \\
		
		
		G & & \textfn{multiple\_prior\_jail\_time}  & \and & \textfn{multiple\_prior\_probation\_or\_fine} & 1.30 & 0.08 & 0.77  \\ 
		
		
		H & & \textfn{age\_1st\_arrest$\leq$17} &  \and  & \textfn{prior\_arrests$\geq$5} & 1.28 & 0.08 & 0.76 \\
		
		\midrule
		
		I & & \textfn{multiple\_prior\_jail\_time}  & \and & \textfn{time\_served$\leq$6mo}  & 1.34 & 0.06 & 0.79  \\ 
		
		J & & \textfn{multiple\_prior\_jail\_time}  & \and & \textfn{age\_1st\_confinement\_18\_to\_24 } & 1.29 & 0.12 & 0.76  \\ 
		
		K & & \textfn{multiple\_prior\_jail\_time}  & \and & \textfn{prior\_arrest\_for\_misdemeanor} & 1.28 & 0.15 & 0.76  \\ 
		
		L & & \textfn{multiple\_prior\_jail\_time}  & \and & \textfn{multiple\_prior\_prison\_time} & 1.28 & 0.13 & 0.75  \\ 
		
		M & & \textfn{multiple\_prior\_jail\_time} & \and & \textfn{any\_prior\_probation\_or\_fine} & 1.27 & 0.13 & 0.75  \\ 
		
		\midrule
		
		N & & \textfn{age\_1st\_arrest$\leq$17} & \and  & \textfn{prior\_arrest\_for\_misdemeanor} & 1.32 & 0.07 & 0.78 \\
		
		O & & \textfn{age\_1st\_arrest$\leq$17}  & \and & \textfn{any\_prior\_jail\_time}  & 1.28 & 0.06 & 0.76  \\ 
		
		P & &  \textfn{age\_1st\_arrest$\leq$17}  & \and & \textfn{age\_1st\_confinement\_18\_to\_24}  & 1.28 & 0.05 & 0.75  \\ 
		
		\midrule
		
		Q & & \textfn{multiple\_prior\_probation\_or\_fine}  & \and  & \textfn{age\_1st\_confinement\_18\_to\_24 } &   1.31 & 0.08 & 0.77 \\
		
		\midrule
		
		R & & \textfn{age\_at\_release\_18\_to\_24}   & \and  & \textfn{prior\_arrest\_for\_misdemeanor} &   1.34 & 0.06 & 0.79 \\
		
		S & &  \textfn{age\_at\_release\_18\_to\_24}  & \and & \textfn{any\_prior\_jail\_time}  & 1.34 & 0.06 & 0.79  \\ 
		
		T & &  \textfn{age\_at\_release\_18\_to\_24}   & \and & \textfn{prior\_arrests$\geq$2}  & 1.32 & 0.10 & 0.78  \\ 
		
		U & &  \textfn{age\_at\_release\_18\_to\_24}   & \and & \textfn{prior\_arrest\_for\_multiple\_types}  & 1.30 & 0.10 & 0.76  \\ 
		
		\midrule
		V & & \textfn{prior\_arrests$\geq$5}  & \and  & \textfn{age\_at\_release\_25\_to\_29} &   1.31 & 0.10 & 0.77 \\
		
		W & & \textfn{prior\_arrests$\geq$5}  & \and & \textfn{age\_1st\_confinement\_18\_to\_24}  & 1.28 & 0.21 & 0.76  \\
		
		X & & \textfn{prior\_arrests$\geq$5}  & \and & \textfn{time\_served$\leq$6mo}  & 1.28 & 0.11 & 0.76  \\
		
		
		
		
		
		
		\bottomrule
	\end{tabular}
\end{table}

\subsection{Falling Rule Lists for Imbalanced Problems}
\label{Sec::FallingRuleLists}

As we discuss in Section \ref{Sec::AccuracyComparison}, it is difficult to use traditional tree and rule-based methods to create non-trivial models on imbalanced classification problems such as \textds{sexual\_violence}. This is possibly because these algorithms employ greedy splitting and pruning procedures. Here, we aim to show that there exist rule-based models that perform well on such problems by training \textit{Falling Rule Lists} \citep{wang2015falling}.
%
%

Falling Rule Lists are ordered lists of IF-THEN rules. The confidence of each rule decreases as we go down the list. In this way, the highest rule applies to the group of individuals that have the highest risk, the second highest rule applies to a group of individuals with the second highest risk, and so on. The algorithm that produces Falling Rule List globally optimizes the list, without greedy splitting and pruning.

We present a Falling Rule List for the \textds{arrest} problem in Table \ref{Model::FRLarrest}, learned from the algorithm of \citet{wang2015falling}. This model was trained using rules with at most two input variables and a support of at least 5\%. The rules listed within this model have the form ``IF $a$ THEN $b$" where $b$ denotes a positive outcome $y = +1$. In Table \ref{Model::FRLarrest}, support refers to the percentage of remaining examples that satisfy the IF conditions and  \textit{probability} refers to percentage of these examples where the outcome variable is positive.  This model shows that the highest risk prisoners are those who were released between ages 18 and 24, and who have at least 5 prior arrests --  this is aligned with the association rule (Rule F) that we found in Section \ref{SecRuleMining}. Once those individuals are removed, the second highest risk prisoners are 25--29 year olds with at least 5 prior arrests, etc. The risk of each group decreases as one moves down the rules. Rule 15 represents the default rule. If an individual does not fall under any of risk groups determined by Rules 1-14, then his/her risk of arrest is 0.21. 

\begin{table}
\scriptsize
\caption{\label{Model::FRLarrest}Falling rule list for \textds{arrest}.} 
\centering
\renewcommand{\arraystretch}{1.2}
\begin{tabular}{@{}p{0mm}@{ }l@{  }l@{  }c@{   }lcc}
\toprule 
& \multicolumn{4}{c}{\textbf{Conditions}}  & \textbf{Probability} & \textbf{Support} \\ 
\toprule
& IF & \textfn{age\_at\_release\_18\_to\_24}  & AND  & \textfn{prior\_arrests$\geq$5} & 0.83 & 0.08 \\
& ELSE IF & \textfn{age\_at\_releaser\_25\_to\_29} & AND & \textfn{prior\_arrests$\geq$5} & 0.77 & 0.13 \\
& ELSE IF & \textfn{multiple\_prior\_jail\_time} & AND  & \textfn{prior\_arrests\_for\_drugs} & 0.73 & 0.18 \\
& ELSE IF & \textfn{age\_at\_release\_30\_to\_39} & AND & \textfn{prior\_arrests$\geq$5} & 0.67 & 0.26 \\
& ELSE IF & \textfn{age\_at\_release\_18\_to\_24} & AND & \textfn{prior\_arrests$\geq$1} & 0.66 & 0.16 \\
& ELSE IF & \textfn{prior\_arrests\_for\_drugs} & AND & \textfn{prior\_arrests\_for\_misdemeanor} & 0.55 & 0.29 \\
& ELSE IF & \textfn{age\_at\_release\_25\_to\_29} & AND & \textfn{prior\_arrests$\geq$2} & 0.54 & 0.17 \\
& ELSE IF & \textfn{multiple\_prior\_jail\_time} & AND & \textfn{prior\_arrests$\geq$5} & 0.54 & 0.27 \\
& ELSE IF & \textfn{age\_1st\_arrest$\leq$17} & & & 0.53 & 0.14 \\
& ELSE IF & \textfn{age\_at\_release\_18\_to\_24} & & & 0.50 & 0.19 \\
& ELSE IF & \textfn{time\_served$\leq$6mo} & AND & \textfn{prior\_arrests\_for\_property} & 0.48 & 0.17 \\
& ELSE IF & \textfn{prior\_arrests$\geq$5} & AND & \textfn{prior\_arrests$\geq$1} & 0.41 & 0.60 \\
& ELSE IF & \textfn{age\_at\_release\_25\_to\_29} & AND & \textfn{age\_1st\_arrest\_18\_to\_24} & 0.41 & 0.16 \\
& ELSE IF & \textfn{age\_at\_release\_30\_to\_39} & AND & \textfn{prior\_arrests$\geq$1} & 0.37 & 0.35 \\
& ELSE &  &  & default & 0.21 &  \\
\bottomrule
\end{tabular}
\end{table}

\clearpage
\section{The Impact of Race}
\label{Sec::RaceImpact}
As discussed earlier, we chose not to include race as an input variable in our prediction problems. Some studies have shown that race is important for accurate recidivism prediction \citep{petersilia1987guideline, berk2009role}. 

We wanted to know the answers to two questions. First, whether including race as a feature would lead to more accurate predictions. Second, whether we could predict race from the features that we already had. If we could predict race well from our current set of features, this would show that race information could be implicitly included in any model we might construct. The results that follow show: (i) including race does not substantially increase prediction accuracy for our problems, and (ii) race can be predicted fairly well from the features we already have. These results indicate that most of the information necessary to predict recidivism is already included in the features we have, and these features also include relevant information for predicting race. 

To address whether race provided an increase in accuracy for predicting recidivism, we re-ran all methods other than SLIM on all new versions of each prediction problem that included three additional race-related input variables: $\textfn{white}$, $\textfn{black}$, $\textfn{hispanic}$. An overview of these variables can be seen in Table \ref{Table::RaceInputVariables}. Table \ref{Table::TestAUCwithRace} presents the models' test AUC when race-related indicator variables are included. Table \ref{Table::TestAUCChangewithRace} represent the percentage increase in AUC when compared to \ref{Table::testAUC}. As shown, the differences for most methods are negligible. In the cases of SVM RBF and Ridge, the accuracy increased slightly. In the case of RF, including race decreases accuracy (most likely because it exacerbates the overfitting problem).

\begin{figure}[]

\setlength{\tabcolsep}{0pt}
\centering
\tabulartitle{\textwidth}{\textds{white}}{\textds{black}} \\ 
\begin{tabular}{lr}
\rocinclude{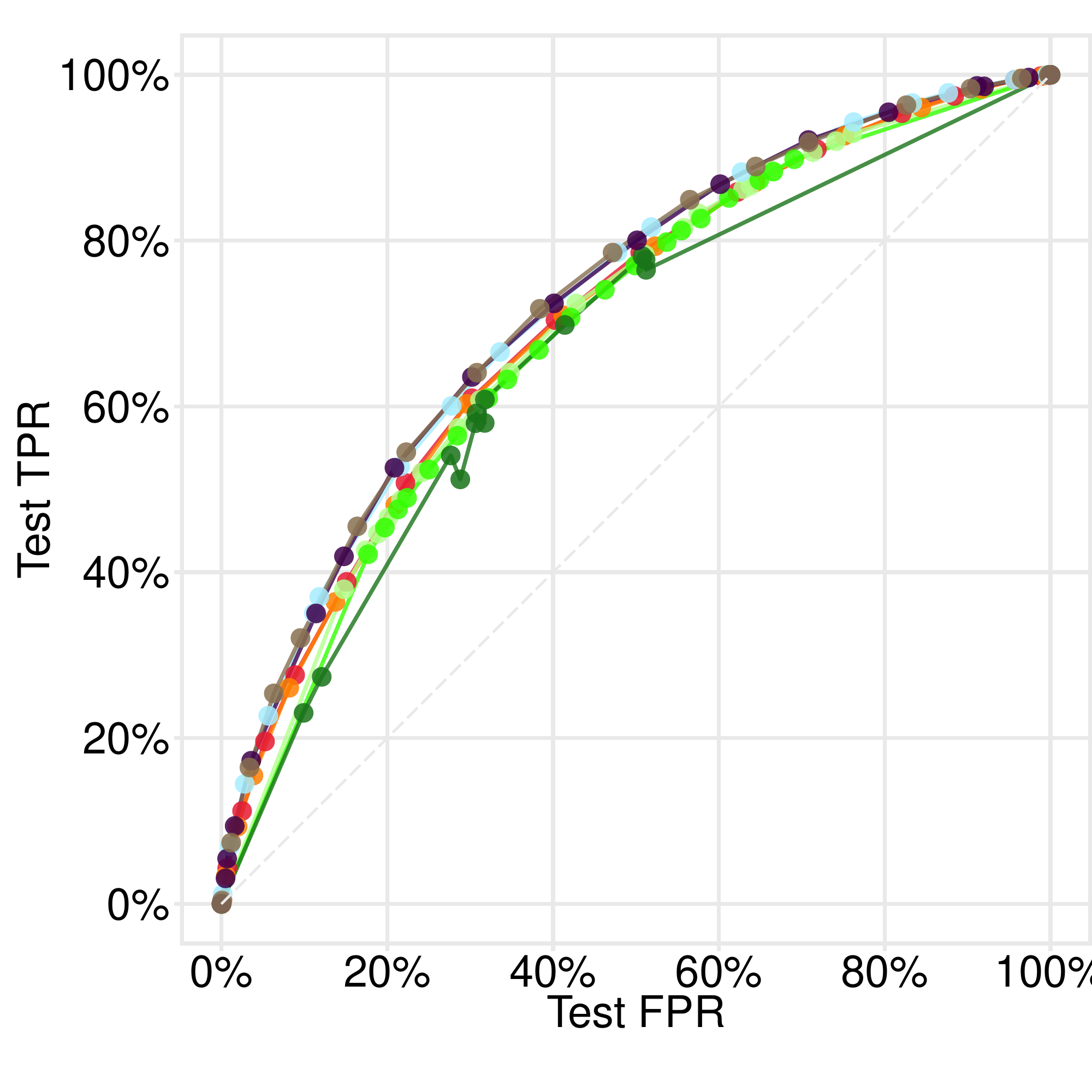} &
\rocinclude{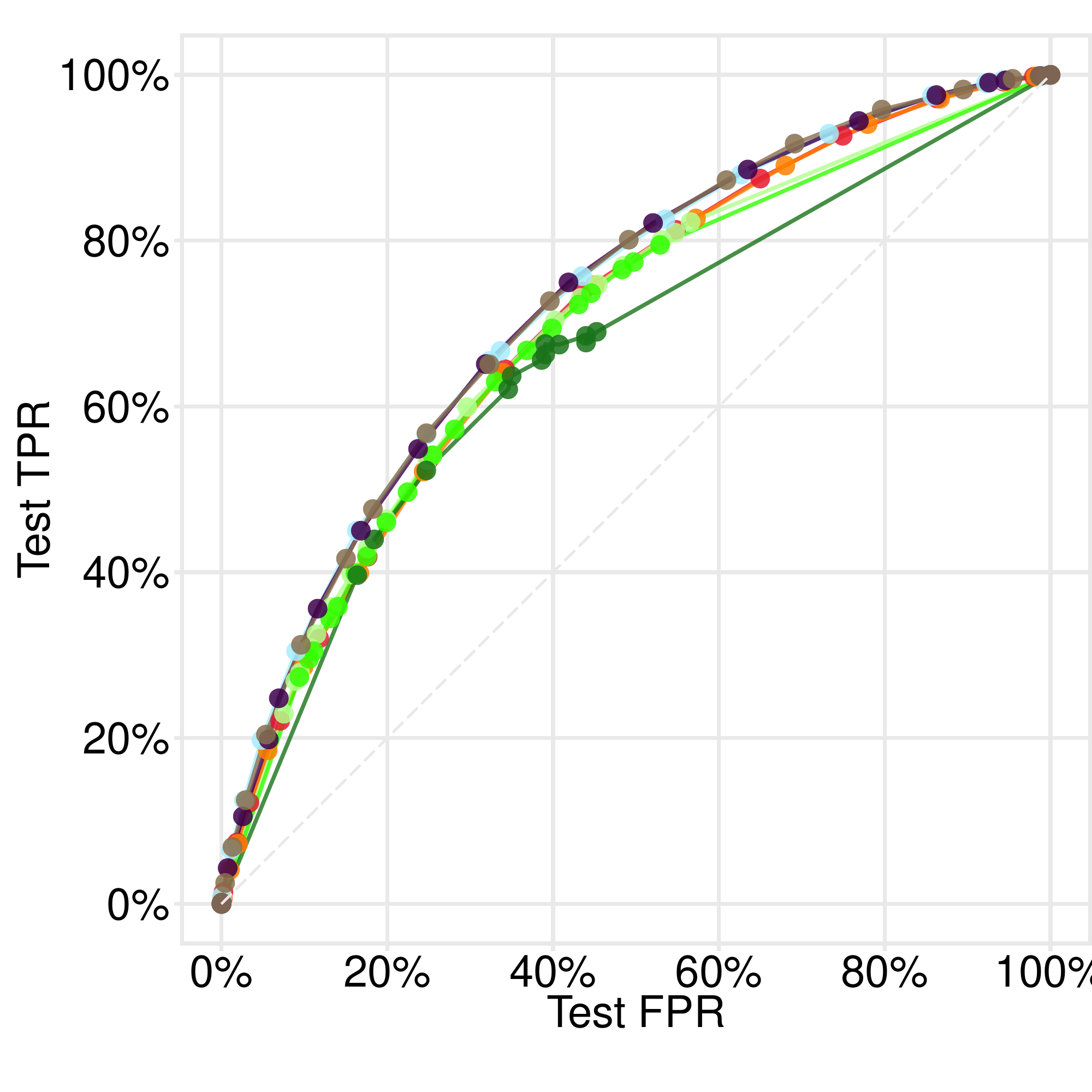} 
\end{tabular}

\vspace{1cm}
\begin{tabular}{c} 
\hspace{1 cm} \textds{hispanic} \\
\rocinclude{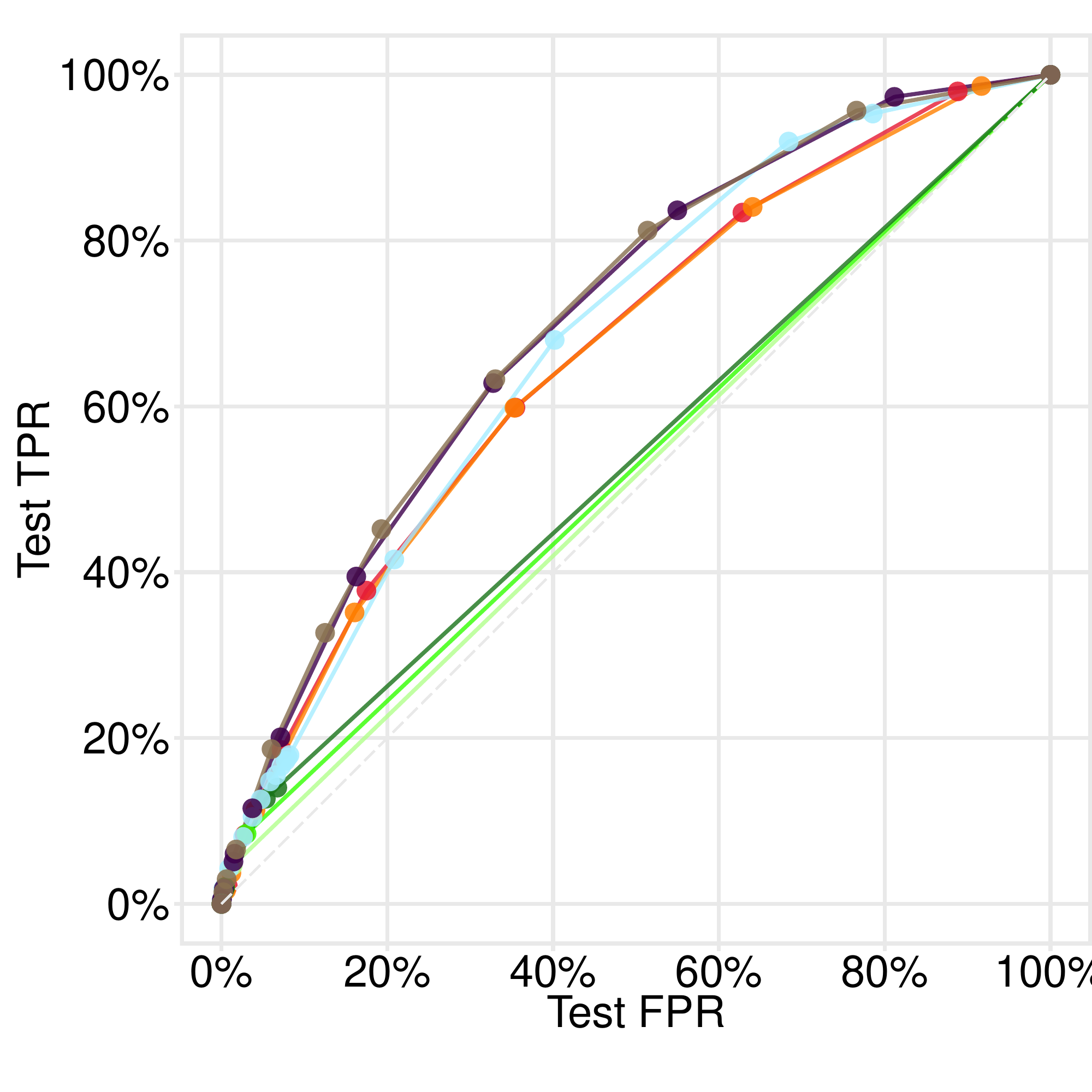} 
\end{tabular}

\vspace{1cm}

\includegraphics[trim=0.5in 0in 0.5in 0.1in,clip,width=0.8\textwidth]{figure/roc_plot_legend}
\caption{ROC curves for predicting \textfn{white}, \textfn{black} and \textfn{hispanic} using the standard set of input variables. }
\label{Figure::ROCPlotsRace}
\end{figure}

To determine whether race could be predicted from the current variables, we used three different race options (\textfn{white}, \textfn{black}, and \textfn{hispanic}) as outcomes and predicted each race as a function of our features. ROC plots are provided in Figure \ref{Figure::ROCPlotsRace}, showing that race can be predicted much better than random guessing. This is not a surprise, as we already know that blacks tend to have longer criminal histories than whites. On the other hand, we remark that we could not predict race perfectly with the features we have - in fact, our predictions (for all methods) were far from perfect. This means that not all of the information about race is contained in the features we have. 


\begin{table}
\caption{\label{Table::RaceInputVariables}Overview of race-related input variables, in addition to the variables in Table \ref{Table::InputVariables}. Each variable is a binary rule of the form $x_{ij} \in \{0,1\}$.}
\centering
\begin{tabular}{llp{2cm}p{5cm}}
\toprule
& {\textnormal{\textbf{Input Variable}}} & {\textbf{P}}{$(x_{ij}=1)$} & {\textbf{Definition}} \\
\toprule

& white & 0.53 & prisoner $i$ is white  \\

& black & 0.44 & prisoner $i$ is black  \\

& hispanic & 0.14 & prisoner $i$ is hispanic  \\

\bottomrule
\end{tabular}
\end{table}

\clearpage
\begin{table}
\caption{\label{Table::TestAUCwithRace}Test AUC for the baseline methods on all prediction problems using the standard set of input variables along with the race-related indicator variables \textfn{white}, \textfn{black} and \textfn{hispanic}.}
\scriptsize
\renewcommand{\arraystretch}{1.2}
\resizebox{\textwidth}{!} {\centering
\begin{tabular}{lcccccccc}
  \toprule
\bfcell{c}{Dataset} & \bfcell{c}{Lasso} & \bfcell{c}{Ridge} & \bfcell{c}{C5.0R} & \bfcell{c}{C5.0T} & \bfcell{c}{CART} & \bfcell{c}{RF} & \bfcell{c}{SVM RBF} & \bfcell{c}{Boosting} \\ 
  \toprule
\texttt{arrest} & \cell{c}{0.73} & \cell{c}{0.74} & \cell{c}{0.72} & \cell{c}{0.71} & \cell{c}{0.69} & \cell{c}{0.74} & \cell{c}{0.72} & \cell{c}{0.74} \\ 
   \midrule \texttt{drug} & \cell{c}{0.75} & \cell{c}{0.75} & \cell{c}{0.64} & \cell{c}{0.65} & \cell{c}{0.59} & \cell{c}{0.76} & \cell{c}{0.74} & \cell{c}{0.76} \\  
   \midrule \texttt{general\_violence} & \cell{c}{0.73} & \cell{c}{0.73} & \cell{c}{0.56} & \cell{c}{0.58} & \cell{c}{0.56} & \cell{c}{0.72} & \cell{c}{0.71} & \cell{c}{0.72} \\ 
   \midrule \texttt{domestic\_violence} & \cell{c}{0.77} & \cell{c}{0.77} & \cell{c}{0.50} & \cell{c}{0.50} & \cell{c}{0.52} & \cell{c}{0.65} & \cell{c}{0.77} & \cell{c}{0.78} \\
   \midrule \texttt{sexual\_violence} & \cell{c}{0.72} & \cell{c}{0.72} & \cell{c}{0.50} & \cell{c}{0.50} & \cell{c}{0.51} & \cell{c}{0.55} & \cell{c}{0.70} & \cell{c}{0.70} \\ 
   \midrule \texttt{fatal\_violence} & \cell{c}{0.68} & \cell{c}{0.69} & \cell{c}{0.50} & \cell{c}{0.50} & \cell{c}{0.50} & \cell{c}{0.50} & \cell{c}{0.69} & \cell{c}{0.70} \\
   \bottomrule \end{tabular}

}
\end{table}
\begin{table}
\caption{\label{Table::TestAUCChangewithRace}Percentage difference of test AUC for models with the inclusion of race-related indicator variables such as \textfn{white}, \textfn{black} and \textfn{hispanic} verses test AUC for models created without.} 
\renewcommand{\arraystretch}{1.2}
\resizebox{\textwidth}{!} {\centering
\begin{tabular}{lcccccccc}
  \toprule
\bfcell{c}{Dataset} & \bfcell{c}{Lasso} & \bfcell{c}{Ridge} & \bfcell{c}{C5.0R} & \bfcell{c}{C5.0T} & \bfcell{c}{CART} & \bfcell{c}{RF} & \bfcell{c}{SVM RBF} & \bfcell{c}{Boosting} \\ 
  \toprule
\texttt{arrest} & \cell{c}{1.2$\%$} & \cell{c}{0.9$\%$} & \cell{c}{-0.2$\%$} & \cell{c}{-0.1$\%$} & \cell{c}{1.4$\%$} & \cell{c}{0.6$\%$} & \cell{c}{-0.9$\%$} & \cell{c}{0.3$\%$} \\
   \midrule \texttt{drug} & \cell{c}{0.9$\%$} & \cell{c}{1.2$\%$} & \cell{c}{0.5$\%$} & \cell{c}{3.8$\%$} & \cell{c}{0.3$\%$} & \cell{c}{1.3$\%$} & \cell{c}{0.9$\%$} & \cell{c}{0.8$\%$} \\  
   \midrule \texttt{general\_violence} & \cell{c}{0.9$\%$} & \cell{c}{1.1$\%$} & \cell{c}{-0.7$\%$} & \cell{c}{1.5$\%$} & \cell{c}{1.1$\%$} & \cell{c}{0.6$\%$} & \cell{c}{1.7$\%$} & \cell{c}{0.6$\%$} \\ 
   \midrule \texttt{domestic\_violence} & \cell{c}{0.0$\%$} & \cell{c}{-0.1$\%$} & \cell{c}{0.0$\%$} & \cell{c}{0.0$\%$} & \cell{c}{-1.0$\%$} & \cell{c}{1.1$\%$} & \cell{c}{-0.1$\%$} & \cell{c}{-0.1$\%$} \\
   \midrule \texttt{sexual\_violence} & \cell{c}{0.2$\%$} & \cell{c}{0.2$\%$} & \cell{c}{0.0$\%$} & \cell{c}{0.0$\%$} & \cell{c}{-0.7$\%$} & \cell{c}{1.2$\%$} & \cell{c}{0.4$\%$} & \cell{c}{0.0$\%$} \\ 
   \midrule \texttt{fatal\_violence} & \cell{c}{1.4$\%$} & \cell{c}{1.4$\%$} & \cell{c}{0.0$\%$} & \cell{c}{0.0$\%$} & \cell{c}{-0.0$\%$} & \cell{c}{-0.3$\%$} & \cell{c}{-1.0$\%$} & \cell{c}{0.8$\%$} \\
   \bottomrule \end{tabular}

}
\end{table}
\clearpage
\setlength{\fwidth}{0.5\textwidth}

\end{document}